\documentclass[preprint,12pt]{elsarticle}
\usepackage[pdfborder={0 0 0}]{hyperref}
\usepackage[margin=1in]{geometry}
\usepackage{amssymb}
\usepackage{amsmath}
\usepackage{hyperref}
\usepackage[utf8]{inputenc}
\usepackage{graphicx}
\usepackage{amsmath}
\usepackage{amsfonts}
\usepackage{graphicx}
\usepackage{everypage}
\usepackage{xspace}
\usepackage{epsfig} 
\usepackage{tikz}
\usepackage{graphicx}
\usepackage{mathtools}
\usepackage{algorithm}
\usepackage{algorithmic}
\usepackage{amsmath}
\usepackage{mathrsfs}
\usepackage{bm}
\usepackage{amsfonts}
\usepackage{color}
\usepackage{xcolor}
\usepackage{soul}
\usepackage{verbatim}
\usepackage{commath}
\usepackage{changepage}
\usepackage{multicol}
\usepackage{lipsum}    
\usepackage{lastpage} 
\usepackage{url} 
\usepackage{booktabs}
\usepackage{rotating}
\usepackage{amsthm}
\usepackage{subcaption}
\usepackage{yhmath}
\usepackage{amsmath}
\usepackage{enumitem}
\usepackage{float}
\usepackage{amsmath,calc}
\usepackage{mathtools}
\usepackage{relsize}
\usepackage{wrapfig}
\usepackage{siunitx}
 \usepackage{rotating}
\usepackage[most]{tcolorbox}
\usepackage{fixltx2e} 
\usepackage{breqn}
\usepackage{amsmath,empheq}
\usepackage{mathtools}
\usepackage[thinc]{esdiff}
\usepackage{color,soul}
\usepackage{nomencl}
\usepackage{multirow}
\usepackage{arydshln}
\usepackage{makecell}
\usepackage{tabularx}
\usepackage[normalem]{ulem}

\newcommand{\x}{\mathbf{x}}

\usepackage{bm}

\newcommand{\ignore}[1]{}

\newcommand{\SKsays}[1]{\textcolor{black}{#1}}
\usepackage[margins]{trackchanges}

\begin{document}
\addeditor{SG}
\begin{frontmatter}

\title{\textcolor{black}{Physics-Informed Latent Neural Operator for Real-time Predictions of time-dependent parametric PDEs}}
\author[1]{Sharmila Karumuri}
\author[1]{Lori Graham-Brady}
\author[1]{Somdatta Goswami\corref{cor1}}

\affiliation[1]{organization={Johns Hopkins University}, 
            addressline={Department of Civil and Systems Engineering}, 
            city={Baltimore},
            postcode={21218}, 
            state={Maryland},
            country={USA}}
\cortext[cor1]{Corresponding author}

\begin{abstract}
Deep operator network (DeepONet) has shown significant promise as surrogate models for systems governed by partial differential equations (PDEs), enabling accurate mappings between infinite-dimensional function spaces. However, when applied to systems with high-dimensional input–output mappings arising from large numbers of spatial and temporal collocation points, these models often require heavily overparameterized networks, leading to long training times. Latent DeepONet addresses some of these challenges by introducing a two-step approach: first learning a reduced latent space using a separate model, followed by operator learning within this latent space. While efficient, this method is inherently data-driven and lacks mechanisms for incorporating physical laws, limiting its robustness and generalizability in data-scarce settings. In this work, we propose PI-Latent-NO, a physics-informed latent neural operator framework that integrates governing physics directly into the learning process. Our architecture features two coupled DeepONets trained end-to-end: a Latent-DeepONet that learns a low-dimensional representation of the solution, and a Reconstruction-DeepONet that maps this latent representation back to the physical space. By embedding PDE constraints into the training via automatic differentiation, our method eliminates the need for labeled training data and ensures physics-consistent predictions. The proposed framework is both memory and compute-efficient, exhibiting near-constant scaling with problem size and demonstrating significant speedups over traditional physics-informed operator models. We validate our approach on a range of  parametric PDEs, showcasing its accuracy, scalability, and suitability for real-time prediction in complex physical systems.
\end{abstract}


\end{frontmatter}

\section{Introduction}
\label{sec:introduction}
\noindent

Neural operators have emerged as a powerful class of deep learning models for building efficient surrogates for expensive parametric partial differential equations (PDEs). 
These operators can be categorized into meta-architectures, those based on the universal approximation theorem for operators \cite{chen1995universal} such as Deep Operator Networks (DeepONet) \cite{lu2021learning}, resolution independent neural operator (RINO) \cite{bahmani2024resolution}, and basis-to-basis operator learning \cite{ingebrand2024basis}; and those based on integral transforms, including the Graph Kernel Network (GKN) \cite{anandkumar2020neural}, Fourier Neural Operator (FNO) \cite{li2020fourier}, Wavelet Neural Operator (WNO) \cite{tripura2023wavelet}, and Laplace Neural Operator (LNO) \cite{cao2024laplace}.
These models learn mappings between infinite-dimensional function spaces, accelerating complex simulations such as material failure prediction \cite{goswami2022physics,fabiani2025enabling,wang2025accelerating} and climate modeling \cite{kurth2023fourcastnet}, while also addressing tasks such as uncertainty propagation~\cite{karumuri2020simulator, abdar2021review, zou2024neuraluq, zou2025uncertainty}, inverse problems (model calibration) \cite{kaltenbach2023semi, molinaro2023neural, karumuri2024learning, long2024invertible, cho2024physics, kag2024learning}, and design optimization \cite{shukla2024deep, ramezankhani2024advanced} in diverse fields. 
However, their practical deployment faces three critical challenges: degrading performance with increasing system dimensionality and complexity, the requirement for extensive training data, and the inability to guarantee physics compliance in their predictions.

While recent advances in latent space operator learning and physics-informed training have separately addressed some of these limitations, a unified framework that simultaneously tackles all these challenges has remained elusive. 
Existing latent deep neural operators \cite{oommen2022learning, zhang2022multiauto, kontolati2024learning}, though computationally efficient, rely on a two-step training process: first learning an efficient latent space through a reduced-order model, then learning the neural operator within this latent space. 
This separation makes physics compliance difficult to achieve, \textcolor{black}{as the decoupled training process hinders the incorporation of physics constraints.}

\textcolor{black}{To address this, the recently proposed WgLaSDI framework~\cite{he2025physics}, takes a more integrated route by combining physics-informed active learning, weak-form dynamics identification (WENDy), and nonlinear dimensionality reduction via autoencoders. Joint training with a weak-form loss enables WgLaSDI to model latent dynamics efficiently and robustly, outperforming full-order solvers in both speed and noise tolerance. The reliance on handcrafted weak-form losses demands significant domain expertise and tuning, potentially limiting its generalizability and ease of adoption.}

Conversely, physics-informed variants that operate directly on the full-order space \textcolor{black}{ (see Figure~\ref{fig:PIVanillaNO}) } often become computationally intractable for complex systems, primarily due to the prohibitively high computational cost associated with calculating PDE gradient terms. 
Recent advancements in separable techniques \cite{yu2024seponet, mandl2024separable} have made these frameworks more computationally efficient.

In this paper, we introduce the physics-informed latent neural operator (see {\color{black}Figure~\ref{fig:PILatentNO}}), representing a fundamental shift in approaching these challenges. Among the various neural operators developed, we employ DeepONet for its architectural flexibility. Our framework combines dimensionality reduction techniques with physics-informed training through two coupled DeepONets trained in a single shot \textcolor{black}{in an encoder-decoder configuration — where the first DeepONet functions as the encoder and the second as the decoder}. The first network learns compact latent representations of the system dynamics, while the second reconstructs solutions in the original space. As compared to the PI-Vanilla-NO illustrated in Figure~\ref{fig:PIVanillaNO}, this architecture introduces built-in separability that enables approximately linear scaling with problem dimensionality — a significant advancement over existing methods that typically scale exponentially. The key contributions of this work include: \vspace{-6pt}
\begin{itemize}
\item The first end-to-end neural operator framework that performs learning directly in latent space, enabling efficient handling of \SKsays{ PDE systems that involve high-dimensional input–output mappings} by leveraging the governing physics.
\item An architecture with inherent separability that drastically accelerates training and inference for \SKsays{ problems with large spatial and temporal resolution.} \vspace{-6pt}
\item Demonstration of approximately linear scaling with problem size, making this approach particularly valuable for complex \SKsays{ PDE-based systems}. \vspace{-6pt}
\end{itemize}
As a proof-of-concept, we demonstrate that our framework achieves accuracy comparable to the state-of-the-art, while requiring significantly fewer computational resources and training data compared to existing methods. These results suggest a promising direction for real-time prediction of complex physical systems, with potential applications ranging from climate modeling to engineering design optimization.

The remainder of this paper is structured as follows: {\color{black}Section~\ref{sec:relatedworks} reviews recent advances in neural operators, latent neural operators, and reduced-order models}. 
Section~\ref{sec:methodology} presents the proposed physics-informed latent neural operator architecture and provide its theoretical foundations. 
Section~\ref{sec:results} demonstrates the effectiveness of the proposed approach through four benchmark problems, comparing its performance against the traditional physics-informed DeepONet (PI-Vanilla-NO) in terms of prediction accuracy and computational efficiency.
\SKsays{Section~\ref{sec:computational cost} provides a detailed discussion on computational cost, including breakeven analysis and inference efficiency of the proposed approach.}
 \textcolor{black}{Section~\ref{sec:latent dynamics} presents a quantitative assessment of physical consistency through the analysis of latent dynamics, while Section~\ref{sec:pde squared residuals} evaluates physics consistency via PDE residuals.}
Finally, Section~\ref{sec:summary} summarizes our key findings and discusses future research directions.

\section{Related Works}
\label{sec:relatedworks}

\subsection{Neural Operators}
In recent years, several neural operator regression methods have been proposed to learn mappings between functional spaces using neural networks. 
In 2019, Sharmila et al. \cite{karumuri2020simulator, karumuri2022physics} introduced a method for learning input-to-output function mappings using residual neural networks. 
However, theoretical guarantees for the universal approximation of operators had not yet been established at that time. 
Later, in 2021, Lu et al. \cite{lu2021learning} introduced DeepONet, which is based on the universal approximation theorem for operators by Chen and Chen \cite{chen1995universal}, enabling the mapping between infinite-dimensional functions using deep neural networks.
In the following years, additional operator regression methods based on integral transforms \cite{li2020fourier, tripura2022wavelet, cao2024laplace} were proposed. These advances will be discussed in detail below.

The DeepONet architecture features two deep neural networks: a branch net, which encodes the input functions at fixed sensor points, and a trunk net, which encodes the spatiotemporal coordinates of the output function. 
The solution operator is expressed as the inner product of the branch and trunk network outputs. The branch and trunk network outputs represent the coefficients and basis functions of the target output function, respectively. 
While DeepONet offers significant flexibility and the ability to learn solution operators for parametric PDEs, it also faces challenges related to training complexity, data requirements, and long-time integration. 
Several modified DeepONet frameworks have been proposed to address these limitations \cite{jin2022mionet, he2023novel, kontolati2023influence, bahmani2024resolution, cao2024deep, kumar2024synergistic, haghighat2024deeponet, he2024sequential, karumuri2024efficient, michalowska2024neural, taccari2024developing}.

Fourier Neural Operators (FNOs) \cite{li2020fourier, qin2024toward} employ neural networks combined with Fourier transforms to map input functions to target functions in the frequency domain. 
The core innovation of FNOs lies in their Fourier layer, which transforms the input into the frequency domain via the Fast Fourier Transform (FFT), applies a linear transformation to the lower Fourier modes, and filters out the higher modes.
The inverse FFT then reconstructs the filtered representation back into the spatial domain. Despite their efficiency and flexibility for various parametric PDE problems, FNOs encounter challenges with non-periodic, heterogeneous, or high-frequency problems, as well as computational scalability, data requirements, and interpretability.

Wavelet Neural Operators (WNOs) \cite{tripura2023wavelet} integrate wavelet transforms with neural networks, decomposing functions into multiscale representations that capture local and global features more effectively. 
Their architecture involves applying a wavelet transform to input data, extracting multiscale features, and processing them through layers before applying an inverse wavelet transform to map the output back to the solution space. 
While WNOs enhance computational efficiency and flexibility, challenges remain in training and selecting suitable wavelet bases.

These operator learning methods have demonstrated promising results across various applications \cite{cai2021deepm, lin2021operator, mao2021deepm, liu2022learning, di2023neural, jiang2024fourier, tripura2023wavelet, chiu2024deeposets}.
However, their effectiveness in solving complex parametric PDEs is constrained by three key challenges: (1) performance deterioration with increasing system size and complexity, (2) the requirement for substantial paired input-output data, making large-scale dataset generation expensive and time-consuming, and (3) approximate solution operators that do not necessarily satisfy the governing PDEs.

To address the challenge of scaling to complex systems, recent advancements in latent space operator methods have shown promise.
These methods accelerate computations by learning operators in low-dimensional latent spaces. 
The approach typically involves dimensionality reduction to obtain a latent representation, followed by operator learning within this reduced space.
Several studies have explored operator learning in latent spaces using DeepONets \cite{oommen2022learning, zhang2022multiauto, kontolati2024learning}. 
Wang et al. \cite{wang2024latent, wang2024latenttime} employed cross-attention-based encoders to project inputs into latent space, followed by transformer layers for operator learning, and decoded the outputs back into the original space using inverse cross-attention. 
Meng et al. 
\cite{meng2024general} proposed a reduced-order neural operator on Riemannian manifolds. 

Despite their potential, these architectures rely heavily on data-driven training, necessitating large datasets.
Physics-informed training, which incorporates PDEs into the loss function, offers a pathway to addressing the second and third challenges. 
Several works have explored physics-informed variants of operator learning methods \cite{wang2021learning, goswami2022physics, mandl2024separable, goswami2023physics, li2024physics, navaneeth2024physics}. 
However, scaling physics-informed training to large and complex problems remains computationally challenging, and existing latent neural operator architectures often lack compatibility with such physics-informed training approaches.

\subsection{Reduced-Order Models (ROMs)}
Reduced-order models (ROMs) \cite{o2022learning} are indispensable tools for accelerating high-fidelity simulations of complex physical systems by projecting these systems onto lower-dimensional subspaces, thereby reducing computational costs while maintaining sufficient accuracy for real-time applications, uncertainty quantification, and optimization. 
One of the earliest techniques in dimensionality reduction is principal component analysis (PCA) \cite{bishop2006pattern}, which identifies the principal directions of variance in the data by finding orthogonal eigenvectors of the covariance matrix. 
PCA is widely used for its simplicity and effectiveness in handling linear systems, providing a foundation for many ROM techniques.
A more advanced approach derived from PCA is proper orthogonal decomposition (POD), which identifies orthogonal basis functions by decomposing datasets into principal components. 
Early work by Willcox et al. \cite{willcox2002balanced, benner2015survey} demonstrated its effectiveness in fields such as aerodynamics \cite{vetrano2015pod}, fluid dynamics \cite{rowley2005model}, and control systems \cite{noack2011reduced}.
More recently, autoencoders have emerged as powerful alternatives, leveraging neural networks to learn efficient, low-dimensional representations of complex systems. 
Unlike POD, autoencoders can capture non-linear relationships within the data, enabling accurate approximations for highly non-linear systems. 
Consisting of an encoder to project input data into a compressed latent space and a decoder to reconstruct the original data, autoencoder-based ROMs have been successfully applied in fluid dynamics to model turbulent flows \cite{mohan2019compressed}, in structural damage detection \cite{li2024mechanics}, and in climate data fusion \cite{johnson2024fusing} offering greater flexibility in handling non-linearity and variability for modern simulation, optimization, and control tasks.
\SKsays{Building on these advances, a growing body of work is focusing on developing physics-informed deep learning-based reduced-order models (PI-DL-ROMs), where the available physical knowledge (e.g., governing equations, conservation laws, or geometric constraints) is explicitly incorporated into the learning process. These approaches are leveraging deep neural networks not only to discover efficient latent representations but also to ensure that the reduced-order dynamics remain consistent with the underlying physics. For instance,}
\textcolor{black}{ a recent work~\cite{friedl2024riemannian} has proposed a geometric network architecture that incorporates physics and geometry as inductive biases, enabling the learning of physically consistent, reduced-order dynamics of high-dimensional Lagrangian systems with improved data efficiency and interpretability.
\SKsays{In addition, authors of~\cite{brivio2024ptpi}, have recently introduced pre-trained PI-DL-ROMs, where an autoencoder is coupled with a DeepONet parametrization to obtain a physics-informed decoder. 
Their approach requires a pre-training phase on data from high-fidelity solvers, followed by a fine-tuning stage carried out in a physics-informed manner to further improve predictive capability. 
While this strategy successfully integrates POD, deep learning, and physics-informed training into a unified reduced-order framework, it still depends on pre-training data availability. 
By contrast, our proposed approach does not need any pre-training data, and instead automatically learns a latent representation while enforcing the governing physical laws, reconstructing the solution in a fully physics-informed manner.}}

\section{Methodology}
\label{sec:methodology}


This work focuses on accelerating simulations of physical systems described by high-dimensional PDEs. We consider PDEs of the following general form, which encompasses time-dependent dynamics, initial conditions, and boundary constraints, expressed as:
\begin{equation}
\begin{cases} 
\frac{\partial u}{\partial t} +
\mathcal{N}\left(u, \frac{\partial u}{\partial t}, \frac{\partial u}{\partial \bm{x}}, \frac{\partial^2 u}{\partial \bm{x}^2}, \ldots, t, \bm{x}, \gamma(t, \bm{x}) \right) = 0, \quad \text{in } \Omega \times (0, T],\\
u(0, \bm{x}) = g(\bm{x}), \quad \text{for } \bm{x} \in \Omega,\\
\mathcal{B}\left(u, \frac{\partial u}{\partial \bm{x}}, t, \bm{x}, \gamma \right) = 0, \quad \text{on } \partial \Omega \times (0, T],
\end{cases}
\end{equation}
where \(\mathcal{N}\) is the nonlinear PDE operator, \(u\) is the solution field varying in space and time, and \(\gamma\) is the input field varying in space and time, which could represent fields such as conductivity, source, or velocity depending on the PDE considered. 
Here, \(\Omega\) denotes the spatial domain, \(T\) is the time duration, \(g(\bm{x})\) specifies the initial condition, and \(\mathcal{B}\) is the boundary condition operator defined on \(\partial \Omega\). 
Our research primarily addresses scenarios where the input field \(\gamma\) or the initial condition \(g(\bm{x})\) is a random stochastic field. 
We represent the discretized version of these random stochastic fields by \( \boldsymbol{\xi} \). 
The main objective of this work is to efficiently learn the mapping between these stochastic input configurations \( \boldsymbol{\xi}\) and the corresponding resultant solution fields \(u\) using our PI-Latent-NO model.

\begin{figure}[!h]
    \centering
    \subfloat[\SKsays{Architecture of the physics-informed vanilla neural operator (PI-Vanilla-NO). The branch network processes $n_i$ input functions sampled at $m$ sensor locations, while the trunk network handles $n_{\text{eval}}=(n_t+1) \times n_{\bm{x}}$ spatiotemporal coordinates. The network outputs solution responses for all input functions with dimensions $(n_i, n_{\text{eval}})$.}]{
        \includegraphics[height=1.5in, width=3.5in]{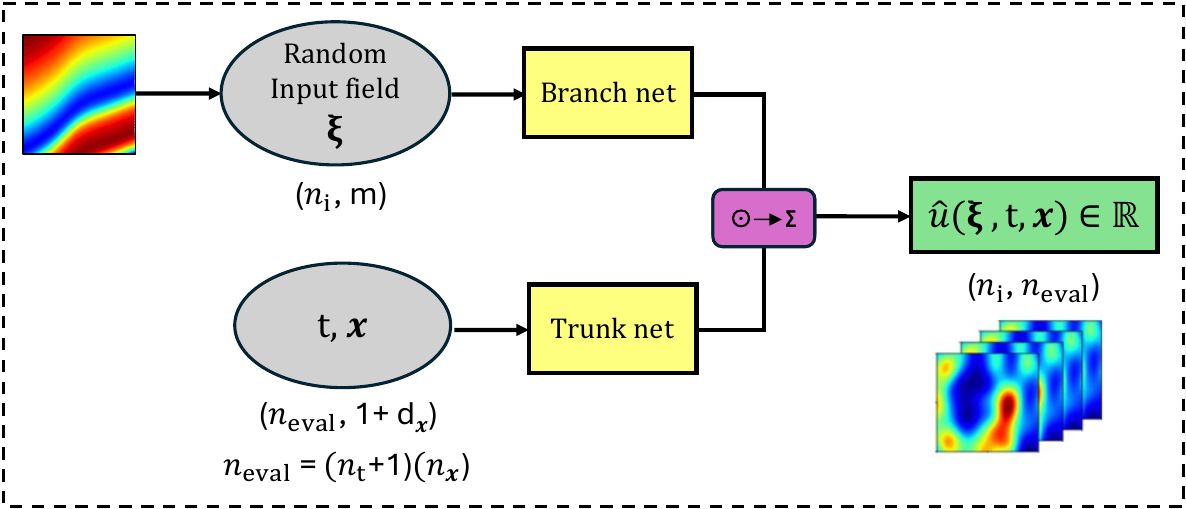}
        \label{fig:PIVanillaNO}
    }
    
    \vspace{0.5cm} 
    
    \subfloat[\SKsays{Schematic of the Physics-Informed Latent Neural Operator (PI-Latent-NO).  Our approach learns solution operators in a latent space using the PI-Latent-NO architecture, which enables physics-informed modeling through two coupled DeepONets: (1) a \textit{Latent DeepONet} that maps $n_i$ input functions to latent trajectories of dimension $n_{\bm{z}}$ over $n_t + 1$ time steps, and (2) a \textit{Reconstruction DeepONet} that decodes these latent trajectories into full-field solutions at $n_{\bm{x}}$ spatial locations, producing outputs of shape $(n_i, n_t + 1, n_{\bm{x}})$. When partial training data is available, ground-truth latent representations can optionally be obtained via a dimensionality reduction technique (e.g., PCA or autoencoders). These can be incorporated as soft constraints during training to further guide the predicted latent fields towards the data manifold. Key advantages: (i) Enables physics-informed training through automatic differentiation of temporal and spatial derivatives, and (ii) achieves linear scaling for large problems through inherent time-space separability, improving upon the quadratic scaling of PI-Vanilla-NO.}]{
        \includegraphics[height=3.2in, width=5.3in]{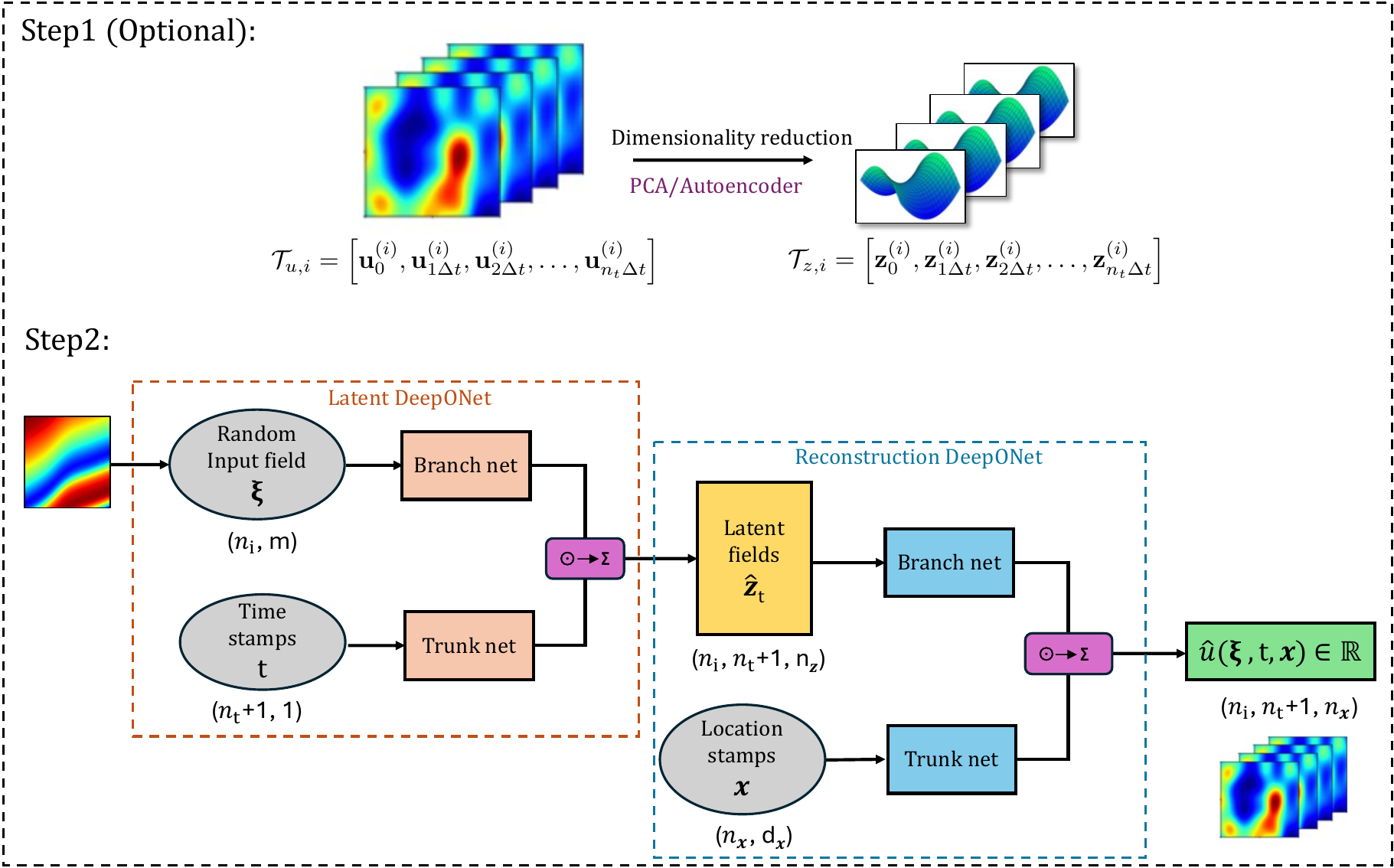}
        \label{fig:PILatentNO}
    }
    
    \caption{\SKsays{Schematics of (a) the baseline physics-informed vanilla neural operator (PI-Vanilla- NO) architecture and (b) our proposed physics-informed latent neural operator (PI-Latent-NO) architecture featuring coupled DeepONets for latent representation and solution reconstruction.}}
\end{figure}

\textcolor{black}{We learn the required mapping using the PI-Latent-NO architecture, shown in Figure~\ref{fig:PILatentNO}, comprised of 
two stacked DeepONets: (1) a Latent-DeepONet that acts as an encoder and learns a low-dimensional latent representation of the PDE solution trajectory at a given time, and (2) a Reconstruction-DeepONet that functions as a decoder and reconstructs the full-order solution in the original spatial domain.}\textcolor{black}{The output of this network is the predicted solution field $\hat{u}$, which approximates the true solution $u$.} \textcolor{black}{Both networks are trained concurrently (in a single shot) using a physics-informed loss, enabling the model to learn operators in the latent space without any labeled data.}

\textcolor{black}{The size of the low-dimensional latent representation, i.e., the output dimension of the Latent-DeepONet, is treated as a hyperparameter in our framework. Its value is problem-dependent: for complex solution fields with intricate spatiotemporal variations, a higher latent dimension is typically required to capture the underlying dynamics adequately; for simpler problems with more regular or smooth behavior, a lower latent dimension often suffices. This flexibility allows the model to balance expressiveness and computational efficiency, depending on the characteristics of the target PDE system.}

\textcolor{black}{
In scenarios where training data is available, the framework can be further extended to incorporate supervision on the latent fields. Specifically, ground-truth latent representations can be obtained by applying a dimensionality reduction method of choice - such as PCA, POD, or autoencoders — to the available solution trajectories. These ground-truth latent fields can then be introduced as an additional constraint in the loss function to guide the latent representations learned by the Latent-DeepONet to produce physically and statistically consistent latent representations.
The procedure for obtaining these ground-truth latent trajectories from data is as follows.  
We begin by sampling \(n_{\text{train}}\) random input configurations \( \boldsymbol{\xi} \) (note that \(n_{\text{train}}\) in this architecture is relatively small compared to purely data-driven training) and obtaining the corresponding full-field ground-truth trajectories of the PDEs, sampled at fixed discrete time intervals \(\Delta t\). 
Each ground-truth trajectory is denoted as the time-ordered set:
\begin{align}
\label{eqn:trajectory}
\mathcal{T}_{u,i} = \left[ \mathbf{u}_0^{(i)}, \mathbf{u}_{1\Delta t}^{(i)}, \mathbf{u}_{2\Delta t}^{(i)}, \ldots, \mathbf{u}_{n_t\Delta t}^{(i)} \right], \quad i = 1, \ldots, n_{\text{train}},
\end{align}
where \(n_t\) denotes the length of the training trajectory, \(\mathbf{u}_a \in \mathbb{R}^{n_{\bm{x}}}\) represents the solution field at time \(t=a\), and \(n_{\bm{x}}\) is the number of spatial grid points.
Using dimensionality reduction techniques, we extract the latent trajectories from these full-order solution trajectories \(\{\mathcal{T}_{u,i}\}_{i=1}^{n_{\text{train}}}\) as
\begin{align}
\mathcal{T}_{z,i} = \left[ \mathbf{z}_0^{(i)}, \mathbf{z}_{1\Delta t}^{(i)}, \mathbf{z}_{2\Delta t}^{(i)}, \ldots, \mathbf{z}_{n_t\Delta t}^{(i)} \right], \quad i = 1, \ldots, n_{\text{train}}.
\end{align}
where \(\mathbf{z}_a \in \mathbb{R}^{{\color{black} n_{\bm{z}}}}\) represents the latent field at time \(t = a\), and \( {\color{black} n_{\bm{z}}} \) represents the dimensionality of the latent field at a given time, with \(n_{\bm{z}} \ll n_{\bm{x}}\).
For example, when using an autoencoder, the latent vectors \(\mathbf{z}_a\) are obtained by minimizing the reconstruction loss:
\begin{equation}
\mathcal{L}(\bm{\theta}_{\text{AE}}) = 
\frac{1}{{n_{\text{train}}} (n_t + 1){\color{black} n_{\bm{x}}}} \sum_{i=1}^{n_{\text{train}}} \sum_{j=0}^{n_t}  \left\Vert \mathbf{u}^{(i)}_{j\Delta t} - \tilde{\mathbf{u}}^{(i)}_{j\Delta t}  \right\Vert_2^2,
\end{equation}
where \(\tilde{\mathbf{u}}^{(i)}_{j\Delta t}\) denotes the decoder’s output.}

\textcolor{black}{The key merits of this architecture are its ability to efficiently estimate low-dimensional latent spaces, which are crucial for managing the complexity of high-dimensional systems. 
Additionally, with our architecture we can obtain derivatives such as \(\left( \frac{\partial \hat{u}}{\partial t}, \frac{\partial \hat{u}}{\partial \bm{x}}, \frac{\partial^2 \hat{u}}{\partial \bm{x}^2}, \ldots \right)\) via automatic differentiation (AD), enabling us to train the model in a purely physics-informed manner for predicting complex responses 
to various PDEs by learning operators in low-dimensional latent spaces. 
We learn the network parameters, $\bm{\theta}$, of this architecture by minimizing the following loss function with physics-informed and data-driven loss components, defined as:
\begin{align}
\label{eqn:Loss function}
\mathcal{L}(\bm{\theta}) = \mathcal{L}_{\text{physics-informed}}(\bm{\theta}) + \mathcal{L}_{\text{data-driven}}(\bm{\theta}) ,
\end{align}
where, 
\begin{align}
\label{eqn:Loss function PI}
\mathcal{L}_{\text{physics-informed}} (\bm{\theta}) &= \SKsays{\lambda_{r}} \mathcal{L}_{r}(\bm{\theta}) + \SKsays{\lambda_{bc}} \mathcal{L}_{bc}(\bm{\theta}) + \SKsays{\lambda_{ic}} \mathcal{L}_{ic}(\bm{\theta})\nonumber \\
& = \frac{\SKsays{\lambda_{r}}}{{n_i} n_{t}^r n_{\bm{x}}^r} \sum_{i=1}^{n_i} \sum_{j=1}^{n_{t}^r} \sum_{k=1}^{n_{\bm{x}}^r} 
\left( \frac{\partial \hat{u}(\boldsymbol{\xi}^{(i)}, t^{(j)}, \bm{x}^{(k)})}{\partial t}+ 
\mathcal{N}[\hat{u}] (\boldsymbol{\xi}^{(i)}, t^{(j)}, \bm{x}^{(k)}) \right)^2  \nonumber \\
&+ \frac{\SKsays{\lambda_{bc}}}{{n_i} n_{t}^{bc} n_{\bm{x}}^{bc}} \sum_{i=1}^{n_i} \sum_{j=1}^{n_{t}^{bc}} \sum_{k=1}^{n_{\bm{x}}^{bc}} 
\left(  \mathcal{B}[\hat{u}] (\boldsymbol{\xi}^{(i)}, t^{(j)}, \bm{x}^{(k)}) \right)^2 \nonumber \\
&+ \frac{\SKsays{\lambda_{ic}}}{{n_i} n_{\bm{x}}^{ic}} \sum_{i=1}^{n_i} \sum_{k=1}^{n_{\bm{x}}^{ic}} 
\left(  \hat{u} (\boldsymbol{\xi}^{(i)}, 0, \bm{x}^{(k)})  - g(\bm{x}^{(k)}) \right)^2,
\end{align}
\begin{align}
\label{eqn:Loss function DD}
\mathcal{L}_{\text{data-driven}} (\bm{\theta})= &  \frac{\SKsays{\lambda_{u}}}{{n_{\text{train}}} (n_t + 1) n_{\bm{x}}} \sum_{i=1}^{n_{\text{train}}} \sum_{j=0}^{n_t} \sum_{k=1}^{n_{\bm{x}}}  
\left( u(\boldsymbol{\xi}^{(i)}, j\Delta t, \bm{x}^{(k)}) - \hat{u}(\boldsymbol{\xi}^{(i)}, j\Delta t, \bm{x}^{(k)}) \right)^2 \\ 
& + \frac{\SKsays{\lambda_{\bm{z}}}}{{n_{\text{train}}} (n_t + 1) {\color{black} n_{\bm{z}}} } \sum_{i=1}^{n_{\text{train}}} \sum_{j=0}^{n_t}  
\left\Vert \mathbf{z}(\boldsymbol{\xi}^{(i)}, j\Delta t) - \hat{\mathbf{z}}(\boldsymbol{\xi}^{(i)}, j\Delta t) \right\Vert_2^2 \nonumber 
 .
\end{align}}

\textcolor{black}{The physics-informed loss term has three components: residual loss, boundary condition loss, and initial condition loss, all based on the PDE for \(n_i\) input functions sampled in each iteration. These loss terms are evaluated at the collocation points \( \{t^{(j)}\}_{j=1}^{n_t^r} \{\bm{x}^{(k)}\}_{k=1}^{n_{\bm{x}}^r} \) within the domain, as well as at the collocation points \( \{t^{(j)}\}_{j=1}^{n_t^{bc}} \{\bm{x}^{(k)}\}_{k=1}^{n_{\bm{x}}^{bc}} \) on the boundary and \( \{\bm{x}^{(k)}\}_{k=1}^{n_{\bm{x}}^{ic}} \) on the initial condition.}

\textcolor{black}{The data-driven loss function consists of two terms: the first term minimizes the MSE between the ground truth solution fields \(u\),  and the predicted responses from our Reconstruction-DeepONet, \(\hat{u}\). The second term minimizes the mean squared error (MSE) between the ground truth latent field, \(\mathbf{z}(\boldsymbol{\xi}^{(i)}, j\Delta t)\), obtained from the dimensionality reduction method, and the predicted latent response from our Latent-DeepONet, \(\hat{\mathbf{z}}(\boldsymbol{\xi}^{(i)}, j\Delta t)\).} 

\SKsays{In this work we set all weighting coefficients 
($\lambda_{r}, \lambda_{bc}, \lambda_{ic}, \lambda_{u}, \lambda_{\bm{z}}$) 
for the respective loss components to $1$, unless otherwise stated, in order to maintain simplicity and reduce the number of hyperparameters. 
Nevertheless, depending on the specific application, the reader may adjust these coefficients 
to emphasize certain components of the loss over others. In practice, careful tuning or use of 
adaptive weighting strategies may further improve training performance, particularly in scenarios
where certain loss terms dominate or under-contribute during optimization.}

Now, moving to how we estimate the gradients in the physics-informed loss, we can see from Figure~\ref{fig:PILatentNO} that the batch-based forward passes of our model reveal a mismatch in the leading dimensions of $t$, $\bm{x}$ and $\hat{u}$. 
This dimension mismatch prevents the direct application of default reverse-mode AD in deep learning frameworks such as PyTorch and TensorFlow to compute the necessary gradients \(\left( \frac{\partial \hat{u}}{\partial t}, \frac{\partial \hat{u}}{\partial \bm{x}}, \frac{\partial^2 \hat{u}}{\partial \bm{x}^2}, \ldots \right)\) for the residual loss term.
To address this, one approach would involve reshaping these quantities to align their leading dimensions, enabling the use of reverse-mode AD, or alternatively, writing a custom reverse AD module with for loops to obtain the required gradients.
However, this would increase computational overhead. 
Therefore, we employ forward-mode AD to estimate the necessary gradients efficiently. 
For instance, forward-mode AD enables us to directly estimate the gradient \(\frac{\partial \hat{u}}{\partial t}\), which aligns with the shape of \(\hat{u}\), as forward AD computes gradients by traversing the computational graph from left to right, avoiding the issue of leading dimension mismatch.

\begin{figure}[H]
    \vspace{-10pt}
    \centering
    \includegraphics[width=0.7\textwidth]{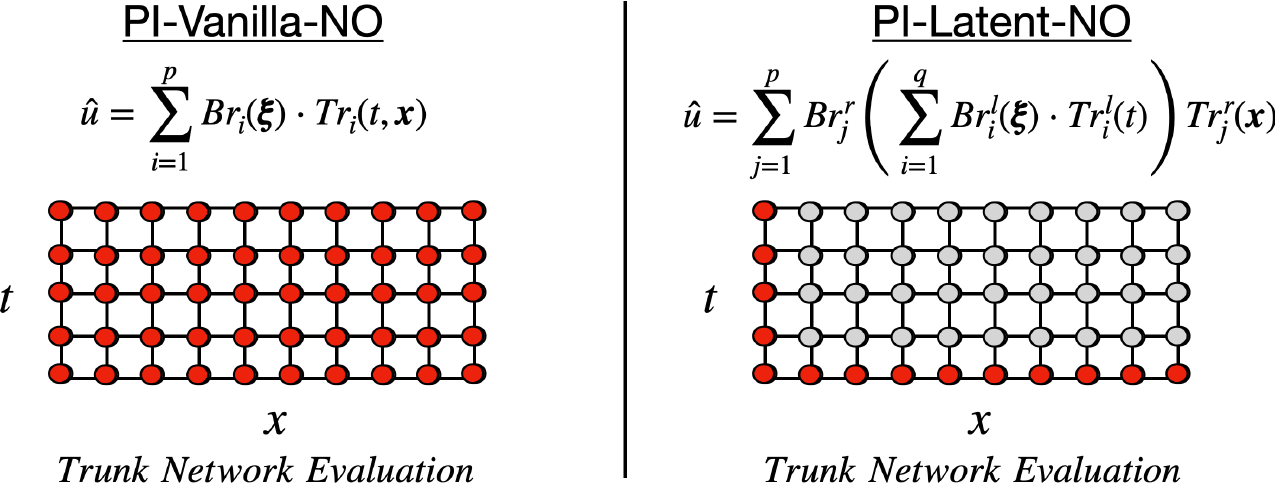}
    \caption{\textcolor{black}{Schematic comparing the number of trunk network evaluations between the baseline PI-Vanilla-NO and the proposed PI-Latent-NO.In a scenario requiring solution evaluation at $5$ time stamps and $10$ spatial grid points, PI-Vanilla-NO has to perform $50$ evaluations (shown in red) per input field — one for each spatiotemporal location. In contrast, PI-Latent-NO reduces the number of trunk evaluations to to just $15$ ($= 5+10$), because of its inherent separability.}}
    \label{fig:Trunk_net_evaluation}
\end{figure}

Comparing the proposed architecture with the PI-Vanilla-NO model in Figure~\ref{fig:PIVanillaNO}, we can clearly see that there is an inherent separability in time and space in our architecture, which provides substantial advantages. 
In large-scale problems where the solution has to be evaluated at numerous time stamps and spatial coordinates — such as for example  at $100$ time stamps (\( n_t = 100 \)) with a spatial grid of $512$ points (\( n_{\bm{x}} = 512 \)) — this separability becomes crucial.
In this case, PI-Vanilla-NO model trunk network would need to be evaluated \( n_{\text{eval}} = 100 \times 512 \) times. 
In contrast, in the proposed model, the trunk network for the latent DeepONet requires only $100$ (\textcolor{black}{=$n_t$}) evaluations, and the reconstruction DeepONet trunk network requires $512$ (\textcolor{black}{=$n_{\bm{x}}$})  evaluations, resulting in a total of $612$ evaluations. 
Thus, because of this separability, the proposed approach scales linearly, in contrast to the quadratic scaling of the vanilla model, making the method highly advantageous for solving \SKsays{ parametric PDE systems that involve high-dimensional input–output mappings (e.g., learning the mapping from an initial condition in $\mathbb{R}^{1024}$ to a solution field in $\mathbb{R}^{10240}$ of the PDE; see Figure~\ref{fig:Trunk_net_evaluation})}.

The complete training process of the PI-Latent-NO model is outlined in detail in Algorithm~\ref{alg:Constraining Latent fields} and Algorithm~\ref{alg:PI-Latent-NO_training}. 

\begin{algorithm}[H]
\scriptsize  
\caption{Latent Trajectory Estimation via Dimensionality Reduction (Optional)}
\label{alg:Constraining Latent fields}
\begin{algorithmic}[1]
\STATE \textbf{Input:}  Training data \(\mathcal{D}_{\text{train}} = \{(\bm{\xi}_i, \mathcal{T}_{u,i})\}_{i=1}^{n_{\text{train}}}\), where \(\bm{\xi}_i\) is the \(i\)-th input function, and \(\mathcal{T}_{u,i} = \left[ \mathbf{u}_0^{(i)}, \mathbf{u}_{\Delta t}^{(i)}, \mathbf{u}_{2\Delta t}^{(i)}, \ldots, \mathbf{u}_{n_t\Delta t}^{(i)} \right]\) is the corresponding output trajectory at different time steps;

\STATE \textbf{Latent Trajectory Estimation:} Obtain the latent trajectory of the output function at each time step for all training data, i.e., \(\{\mathcal{T}_{z,i}\}_{i=1}^{n_{\text{train}}}\), where \(\mathcal{T}_{z,i} = \left[ \mathbf{z}_0^{(i)}, \mathbf{z}_{\Delta t}^{(i)}, \mathbf{z}_{2\Delta t}^{(i)}, \ldots, \mathbf{z}_{n_t\Delta t}^{(i)} \right]\), using reduced-order modeling techniques:\\
 \hspace*{1.5em} a) \textbf{Using PCA:} Learn principal components \(\mathbf{W}_{d_{\bm{z}}}\) and define latent states \(\mathbf{z}_{j\Delta t}^{(i)} = \mathbf{W}_{d_{\bm{z}}}^T \mathbf{u}_{j\Delta t}^{(i)}\).\\
 \hspace*{2.5em} Procedure: 
 \hspace*{2.5em} \[
\mathcal{L}(\bm W) = 
\frac{1}{{n_{\text{train}}} (n_t + 1) n_{\bm{x}}} \sum_{i=1}^{n_{\text{train}}} \sum_{j=0}^{n_t}  \left\Vert \mathbf{u}^{(i)}_{j\Delta t} - \tilde{\mathbf{u}}^{(i)}_{j\Delta t}  \right\Vert_2^2,
\]
 \hspace*{2.5em} where \(\tilde{\mathbf{u}}_{j\Delta t}^{(i)} = \mathbf{W} \mathbf{W}^T \mathbf{u}_{j\Delta t}^{(i)}\) is the reconstruction of \(\mathbf{u}_{j\Delta t}^{(i)}\) using the principal components, and \(\mathbf{W}\) is the matrix of principal components obtained from the SVD. The latent representation of the solution at time step \(j\Delta t\) is given by:
\[
\mathbf{z}_{j\Delta t}^{(i)} = \mathbf{W}_{d_{\bm{z}}}^T \mathbf{u}_{j\Delta t}^{(i)},
\]
 \hspace*{2.5em} where \(\mathbf{W}_{d_{\bm{z}}}\) is the matrix of \(d_{\bm{z}}\) principal components.\\

 \hspace*{1.5em} b) \textbf{Using Autoencoder:} Train encoder-decoder networks to minimize reconstruction error and define latent states \(\mathbf{z}_{j\Delta t}^{(i)} = \text{Encoder}(\mathbf{u}_{j\Delta t}^{(i)})\).\\
 \hspace*{2.5em} Procedure: Train the autoencoder by minimizing the reconstruction loss:
\[
\mathcal{L}(\bm{\theta}_{\text{AE}}) = 
\frac{1}{{n_{\text{train}}} (n_t + 1)n_{\bm{x}}} \sum_{i=1}^{n_{\text{train}}} \sum_{j=0}^{n_t}  \left\Vert \mathbf{u}^{(i)}_{j\Delta t} - \tilde{\mathbf{u}}^{(i)}_{j\Delta t}  \right\Vert_2^2,
\]
 \hspace*{2.5em} where \(\tilde{\mathbf{u}}_{j\Delta t}^{(i)} = \text{Decoder}(\mathbf{z}_{j\Delta t}^{(i)})\) and \(\mathbf{z}_{j\Delta t}^{(i)} = \text{Encoder}(\mathbf{u}_{j\Delta t}^{(i)})\).

\STATE \textbf{Output:}   \(n_{\bm{z}}\) and $\mathcal{T}_{z,i}$.
\end{algorithmic}
\end{algorithm}


\begin{algorithm}[H]
\scriptsize  
\caption{Training Algorithm for PI-Latent-NO}
\label{alg:PI-Latent-NO_training}
\begin{algorithmic}[1]
\STATE \textbf{Input:} A set of $n$ input functions \(\{\bm{\xi}_i\}_{i=1}^{n}\); neural network architectures for the branch and trunk networks of the Latent and Reconstruction DeepONets; size of latent dimension \(n_{\bm{z}}\); number of iterations \(n_{\text{iter}}\); batch size \(\text{bs}\); learning rate \(\alpha\); \SKsays{ and loss weighting coefficients ($\lambda_{r}, \lambda_{bc}, \lambda_{ic}, \lambda_{u}, \lambda_{\bm{z}}$);} training data \(\mathcal{D}_{\text{train}} = \{(\bm{\xi}_i, \mathcal{T}_{u,i})\}_{i=1}^{n_{\text{train}}}\), where \(\bm{\xi}_i\) is the \(i\)-th input function, and \(\mathcal{T}_{u,i} = \left[ \mathbf{u}_0^{(i)}, \mathbf{u}_{\Delta t}^{(i)}, \mathbf{u}_{2\Delta t}^{(i)}, \ldots, \mathbf{u}_{n_t\Delta t}^{(i)} \right]\) is the corresponding output trajectory at different time steps;

\STATE \textbf{Step 1 (Optional):} Get $n_{\bm{z}}$ and latent trajectories $\mathcal{T}_{z,i}$ from Algorithm~\ref{alg:Constraining Latent fields}.
\STATE \textbf{Step 2:} Train the PI-Latent-NO model.

\FOR{iteration = 1 to \(n_{\text{iter}}\)}
   \STATE Compute the physics-informed loss, $\mathcal{L}_{\text{physics-informed}}(\bm{\theta})$: 
      \STATE \hspace*{1.5em} Randomly sample \(\text{bs}\) input functions from  \(\{\bm{\xi}_i\}_{i=1}^{n}\),
      \STATE \hspace*{1.5em} Randomly sample \((n_t^r, n_{\bm{x}}^r)\) spatiotemporal collocation points from the domain, \((n_t^{bc}, n_{\bm{x}}^{bc})\) points from the boundary, and \((n_{\bm{x}}^{ic})\) points at the initial time,
    \hspace*{1.5em}
    \begin{align*}
    \mathcal{L}_{\text{physics-informed}}(\bm{\theta}) &= \SKsays{\lambda_{r}} \mathcal{L}_{r}(\bm{\theta}) + \SKsays{\lambda_{bc}} \mathcal{L}_{bc}(\bm{\theta}) + \SKsays{\lambda_{ic}} \mathcal{L}_{ic}(\bm{\theta}) \nonumber \\
    &= \frac{\SKsays{\lambda_{r}}}{{\text{bs}} \ n_{t}^r n_{\bm{x}}^r} \sum_{i=1}^{\text{bs}} \sum_{j=1}^{n_{t}^r} \sum_{k=1}^{n_{\bm{x}}^r} 
    \left( \frac{\partial \hat{u}(\boldsymbol{\xi}^{(i)}, t^{(j)}, \bm{x}^{(k)})}{\partial t}+ 
    \mathcal{N}[\hat{u}] (\boldsymbol{\xi}^{(i)}, t^{(j)}, \bm{x}^{(k)}) \right)^2 \nonumber \\
    &\quad + \frac{\SKsays{\lambda_{bc}}}{{\text{bs}} \  n_{t}^{bc} n_{\bm{x}}^{bc}} \sum_{i=1}^{\text{bs}} \sum_{j=1}^{n_{t}^{bc}} \sum_{k=1}^{n_{\bm{x}}^{bc}} 
    \left(  \mathcal{B}[\hat{u}] (\boldsymbol{\xi}^{(i)}, t^{(j)}, \bm{x}^{(k)}) \right)^2 \nonumber \\
    &\quad + \frac{\SKsays{\lambda_{ic}}}{{\text{bs}} \  n_{\bm{x}}^{ic}} \sum_{i=1}^{\text{bs}} \sum_{k=1}^{n_{\bm{x}}^{ic}} 
    \left(  \hat{u} (\boldsymbol{\xi}^{(i)}, 0, \bm{x}^{(k)})  - g(\bm{x}^{(k)}) \right)^2. 
    \end{align*}
   
   \STATE Compute the data-driven loss, $\mathcal{L}_{\text{data-driven}}(\bm{\theta})$: 
    \[
    \mathcal{L}_{\text{data-driven}} (\bm{\theta})= \frac{\SKsays{\lambda_{u}}}{{n_{\text{train}}} (n_t + 1) n_{\bm{x}}} \sum_{i=1}^{n_{\text{train}}} \sum_{j=0}^{n_t} \sum_{k=1}^{n_{\bm{x}}}  
    \left( u(\boldsymbol{\xi}^{(i)}, j\Delta t, \bm{x}^{(k)}) - \hat{u}(\boldsymbol{\xi}^{(i)}, j\Delta t, \bm{x}^{(k)}) \right)^2
    \]
       \[
    + 
    \frac{\SKsays{\lambda_{\bm{z}}}}{{n_{\text{train}}} (n_t + 1)  n_{\bm{z}}} \sum_{i=1}^{n_{\text{train}}} \sum_{j=0}^{n_t}  
    \left\Vert \mathbf{z}(\boldsymbol{\xi}^{(i)}, j\Delta t) - \hat{\mathbf{z}}(\boldsymbol{\xi}^{(i)}, j\Delta t) \right\Vert_2^2, 
    \]
     where $\hat{\bm{z}}$ and $\hat{u}$ denote the outputs of the Latent DeepONet and the Reconstruction DeepONet, respectively. The loss term involving $\|\mathbf{z} - \hat{\mathbf{z}}\|^2$ should be omitted if the optional step described above (i.e., Step 1) is not applied. 
    \STATE Compute the total loss:
    \[
    \mathcal{L}(\bm{\theta}) = \mathcal{L}_{\text{data-driven}}(\bm{\theta}) + \mathcal{L}_{\text{physics-informed}}(\bm{\theta}),
    \]
    \STATE Backpropagate the loss through the networks and update the weights of the PI-Latent-NO model: 
    \[
    \bm{\theta} \leftarrow \bm{\theta} - \alpha \nabla_{\bm{\theta}} \mathcal{L}(\bm{\theta}).
    \]
\ENDFOR

\STATE \textbf{Output:} Trained PI-Latent-NO model.
\end{algorithmic}
\end{algorithm}
\section{Results}
\label{sec:results}
 
In this section, we demonstrate the effectiveness of our proposed framework in predicting solutions for various benchmark parametric PDE examples from the literature, comparing the performance of the proposed model against the PI-Vanilla-NO model. 
A summary of the examples considered in this work is presented in Table~\ref{tab:examples}. 
For all examples presented in this work, we do not enforce any constraints on the latent fields to align with representations obtained from standard dimensionality reduction methods. However, readers may choose to impose such constraints by requiring the latent representation at a given time to match that derived from a dimensionality reduction technique of their choice, and include the corresponding loss in the data-driven loss term.
\textcolor{black}{The details of the network architectures and hyperparameter configurations for both the baseline PI-Vanilla-NO and the proposed PI-Latent-NO models are summarized in Tables~\ref{tab:architectures-hyperparameters-vanilla-1D} and~\ref{tab:architectures-hyperparameters-vanilla-2D} for the baseline model, and Tables~\ref{tab:architectures-hyperparameters-ours-1D} and~\ref{tab:architectures-hyperparameters-ours-2D} for the proposed model.}
The code and data for all examples will be made publicly available on \url{https://github.com/Centrum-IntelliPhysics/Physics-Informed-Latent-DeepONet} upon publication. The training for all the examples shown was carried on a single Nvidia A$100$ GPU with 40GB memory. For the memory and runtime comparison studies in Examples $1$ and $3$, we have employed Nvidia A$100$ GPU with $80$ GB memory.

In this work, the performance of models is evaluated on the test samples based on two metrics:
\begin{enumerate}
\item The mean $R^2$ score  (coefficient of determination) of the test data, defined as:
\begin{equation}
\label{eqn:mean_$R^2$_score}
\overline{R^2}_{\text{test}} = \frac{1}{n_{\text{test}}} \sum_{i=1}^{n_{\text{test}}} \left(1 - \frac{\sum_{j=1}^{n_{t}} \sum_{k=1}^{n_{\bm{x}}} \left(u (\boldsymbol{\xi}^{(i)}, t^{(j)}, \bm{x}^{(k)})  - \hat{u}(\boldsymbol{\xi}^{(i)}, t^{(j)}, \bm{x}^{(k)}) \right)^2}{ \sum_{j=1}^{n_{t}} \sum_{k=1}^{n_{\bm{x}}}\left(u (\boldsymbol{\xi}^{(i)}, t^{(j)}, \bm{x}^{(k)})  - \bar{u}(\boldsymbol{\xi}^{(i)} ) \right)^2}\right),
\end{equation}
where $i$ indexes all test samples, and $j$ and $k$ indexes over all the output spatiotemporal locations at which the PDE solution is available.
Here $u$ and $\hat{u}$ are the ground-truth and the predicted values of the solution field, respectively. $\bar{u}(\boldsymbol{\xi}^{(i)})$ is the mean for the $i^{\text{th}}$ test sample. 
The mean $R^2$ score measures how well a predictive model captures the variance in the true data across multiple test samples. 
Ranging from $-\infty$ to $1$, an $R^2$ score of $1$ signifies a perfect prediction, $0$ indicates that the model is no better than predicting the mean of the true values, and negative values suggest worse performance than the mean-based model. 
Averaging this score across multiple test samples provides a comprehensive measure of the model’s ability to generalize to new data.
\item The mean relative $L_2$ error of test data is defined as:
\begin{equation}
\label{eqn:relative_l2}
\text{Mean Rel. $L_2$ Error}_{\text{test}} = \frac{1}{n_{\text{test}}} \sum_{i=1}^{n_{\text{test}}} \frac{\sqrt{ \sum_{j=1}^{n_{t}} \sum_{k=1}^{n_{\bm{x}}}
\left( u (\boldsymbol{\xi}^{(i)}, t^{(j)}, \bm{x}^{(k)}) - \hat{u} (\boldsymbol{\xi}^{(i)}, t^{(j)}, \bm{x}^{(k)}) \right)^2 }
}{
  \sqrt{\sum_{j=1}^{n_{t}} \sum_{k=1}^{n_{\bm{x}}}
\left( u (\boldsymbol{\xi}^{(i)}, t^{(j)}, \bm{x}^{(k)}) \right)^2}
},
\end{equation}
where $i$ indexes all test samples, $j$ and $k$ indexes over all the spatiotemporal locations at which the PDE solution is available.
Here $u$ and $\hat{u}$ are the ground-truth and the predicted values of the solution field, respectively. This metric measures the discrepancy between true and predicted solutions relative to the norm of the true solution for all test samples.
\end{enumerate}

\begin{table}[H]
\centering
\caption{Schematic of 1D and 2D operator learning  benchmarks under consideration in this work.}
\tiny 
\renewcommand{\arraystretch}{1.2} 
\setlength{\tabcolsep}{6pt}      
\begin{tabular}{|c|p{1.85in}|p{2.0in}|p{2.5in}|} 
\hline
\textbf{Case} & \textbf{PDE} & \textbf{Input Function} & \textbf{Samples Visualization}\\ 
\hline
\rotatebox[origin=c]{90}{1D Diffusion - reaction dynamics } &
$\begin{aligned}
 &\frac{\partial u}{\partial t}  = D \frac{\partial^2 u}{\partial x^2} + k u^2 + s(x), \\
&D=0.01, \ k=0.01, \\
&(t, x) \in (0, 1] \times (0, 1],\\
&u(0, x) = 0, \ x \in (0,1)\\
&u(t, 0) = 0, \ t \in (0,1)\\
&u(t, 1) = 0, \ t \in (0,1)\\
&\mathcal G_{\boldsymbol{\theta}}: s(x) \to u(t,  x).
\end{aligned}$ &
$\begin{aligned}
&s(x) \sim \mathrm{GP}(0, k(x, x')), \\
&\ell_{x} = 0.2, \ \sigma^2 = 1.0,\\
&k(x, x') = \sigma^2 \exp\left\{- \frac{\|x - x'\|^2}{2\ell_x^2}\right\}.
\end{aligned}$ &
\begin{minipage}{6.5cm}
    \vspace{3pt}
    \centering
    \includegraphics[width=0.90\linewidth]{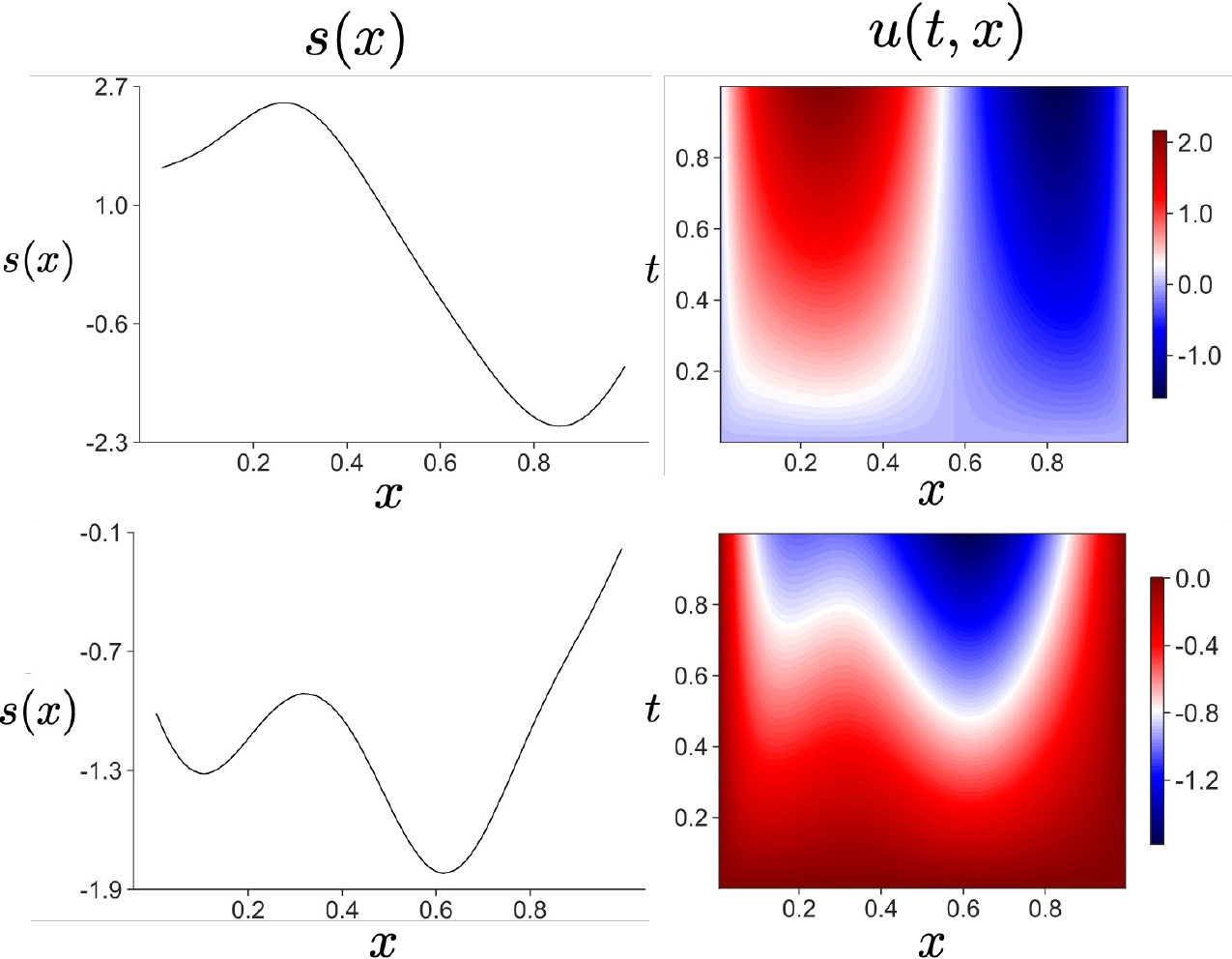}
    \vspace{3pt}
    \vspace{3pt}
\end{minipage} \\
\hline
\rotatebox[origin=c]{90}{1D Burgers’ \SKsays{equation} } &
$\begin{aligned}
&\frac{\partial u}{\partial t} + u \frac{\partial u}{\partial x} - \nu \frac{\partial^2 u}{\partial x^2} = 0, \\
&\nu=0.01, \\
&(t, x) \in (0, 1] \times (0, 1],\\
&u(0, x) = g(x), \ x \in (0,1)\\
&u(t, 0) = u(t, 1)\\
&\frac{\partial u}{\partial x} (t, 0) = \frac{\partial u}{\partial x} (t, 1)\\
&\mathcal G_{\boldsymbol{\theta}}: g(x) \to u(t,  x).
\end{aligned}$ &
$\begin{aligned}
&g(x) \sim \mathcal{N}\left(0, 25^2 \left( -\Delta + 5^2I \right)^{-4} \right) \\
\end{aligned}$ &
\begin{minipage}{6.5cm}
    \vspace{3pt}
    \centering
    \includegraphics[width=0.90\linewidth]{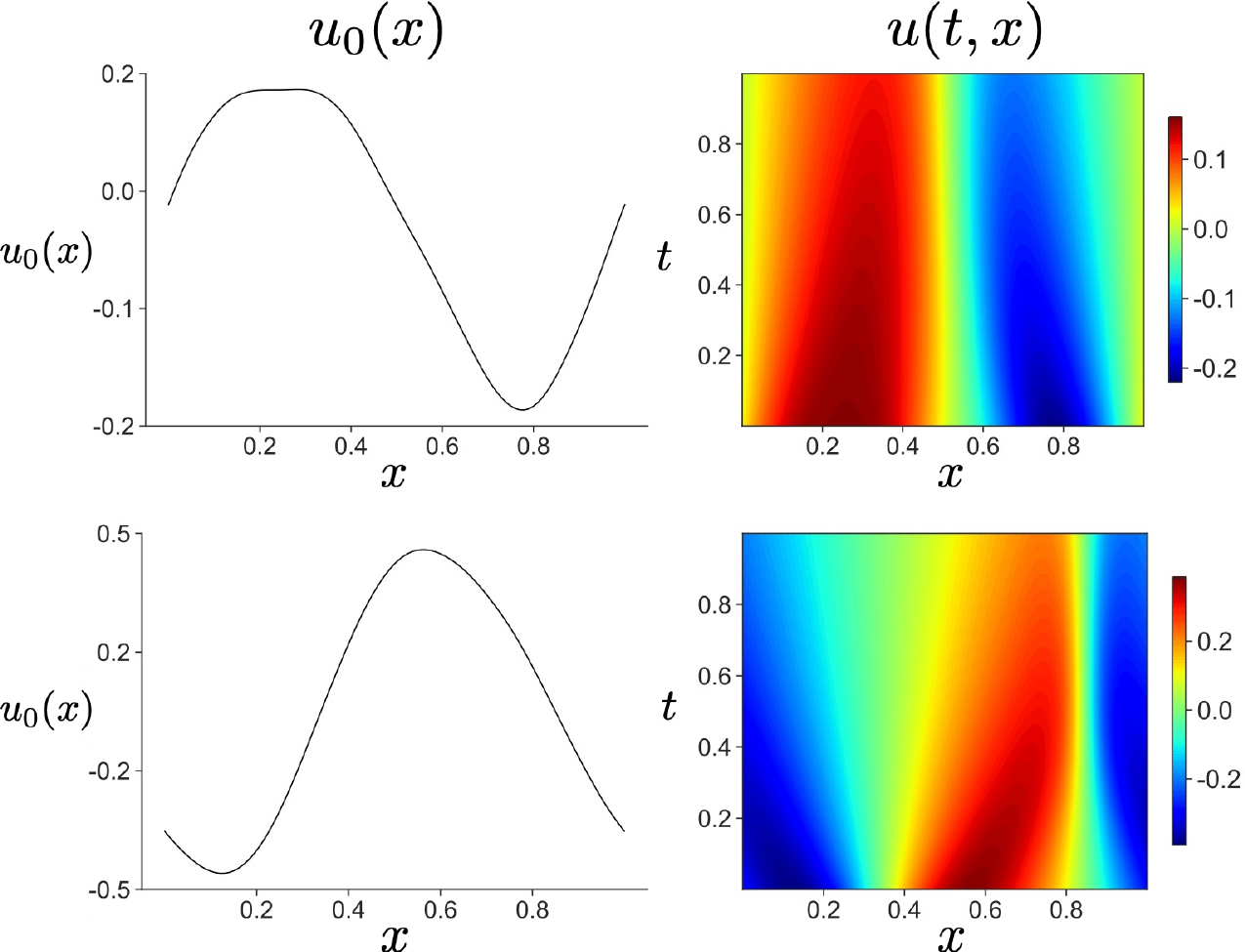} 
    \vspace{3pt}
    \vspace{3pt}
\end{minipage}\\
\hline
\rotatebox[origin=c]{90}{2D Stove burner simulation} &
$\begin{aligned}
&\frac{\partial u}{\partial t} = D \left( \frac{\partial^2 u}{\partial x_1^2} + \frac{\partial^2 u}{\partial x_2^2} \right) \\
&  \quad \quad \quad  + s(x_1, x_2, \text{shape}, r, a), \\
&D = 1, \quad t \in (0, 1], \\
&(x_1, x_2) \in [-2, 2]^2, \\
&u(0, x_1, x_2) = 0, \\
&u(t, x_1, x_2) = 0 \quad \text{on } \partial \Omega, \\
&\mathcal{G}_{\boldsymbol{\theta}}: s(x_1, x_2, \text{shape}, r, a) \to u(t, x_1, x_2).
\end{aligned}$&
$\begin{aligned}
&\text{shape} \sim \mathcal{U}(\{\text{circle}, \text{half-circle}, \text{rhombus}, \dots\}),\\
&r \sim \mathcal{U}(0.75, 1.25) \quad \text{(burner size)}, \\
&a \sim \mathcal{U}(5, 15) \quad \text{(burner intensity)}.
\end{aligned}$ &
\begin{minipage}{6.5cm}
    \vspace{3pt}
    \centering
    \includegraphics[width=0.90\linewidth]{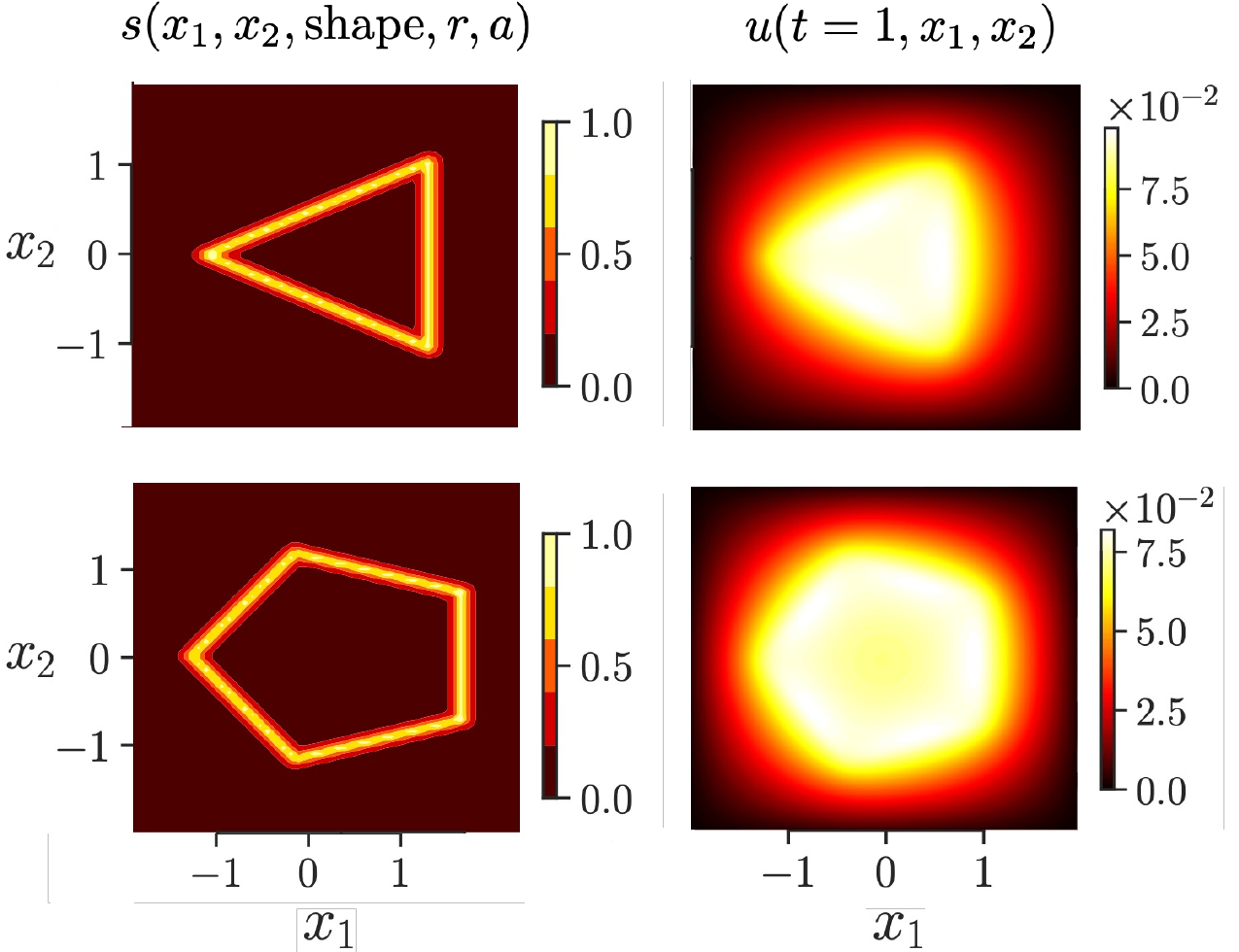} 
    \vspace{3pt}
\end{minipage}\\
\hline
\rotatebox[origin=c]{90}{2D Burgers’ \SKsays{equation}} &
$\begin{aligned}
&\frac{\partial u}{\partial t} + u \frac{\partial u}{\partial x_1} + u \frac{\partial u}{\partial x_2} = \nu \left( \frac{\partial^2 u}{\partial x_1^2} + \frac{\partial^2 u}{\partial x_2^2} \right), \\
&\nu = 0.01, \\
&t \in (0, 1], \quad (x_1, x_2) \in [0, 1]^2, \\
&u(0, x_1, x_2) = u_0(x_1, x_2), \\
&u(t, 0, x_2) = u(t, 1, x_2), \\
& \frac{\partial u}{\partial x_1}\bigg|_{x_1=0} = \frac{\partial u}{\partial x_1}\bigg|_{x_1=1}, \\
&u(t, x_1, 0) = u(t, x_1, 1), \\
& \frac{\partial u}{\partial x_2}\bigg|_{x_2=0} = \frac{\partial u}{\partial x_2}\bigg|_{x_2=1}, \\
&\mathcal{G}_{\boldsymbol{\theta}}: u_0(x_1, x_2) \to u(t, x_1, x_2).
\end{aligned}$&
$\begin{aligned}
&u_0(x_1, x_2) \sim \text{GRF}(\text{Mattern}, l=0.125, \\
&  \quad \quad \quad \quad \quad \quad \quad\ \sigma=0.15, \text{periodic BCs})
\end{aligned}$&
\begin{minipage}{6.5cm}
    \vspace{3pt}
    \centering
    \includegraphics[width=0.90\linewidth]{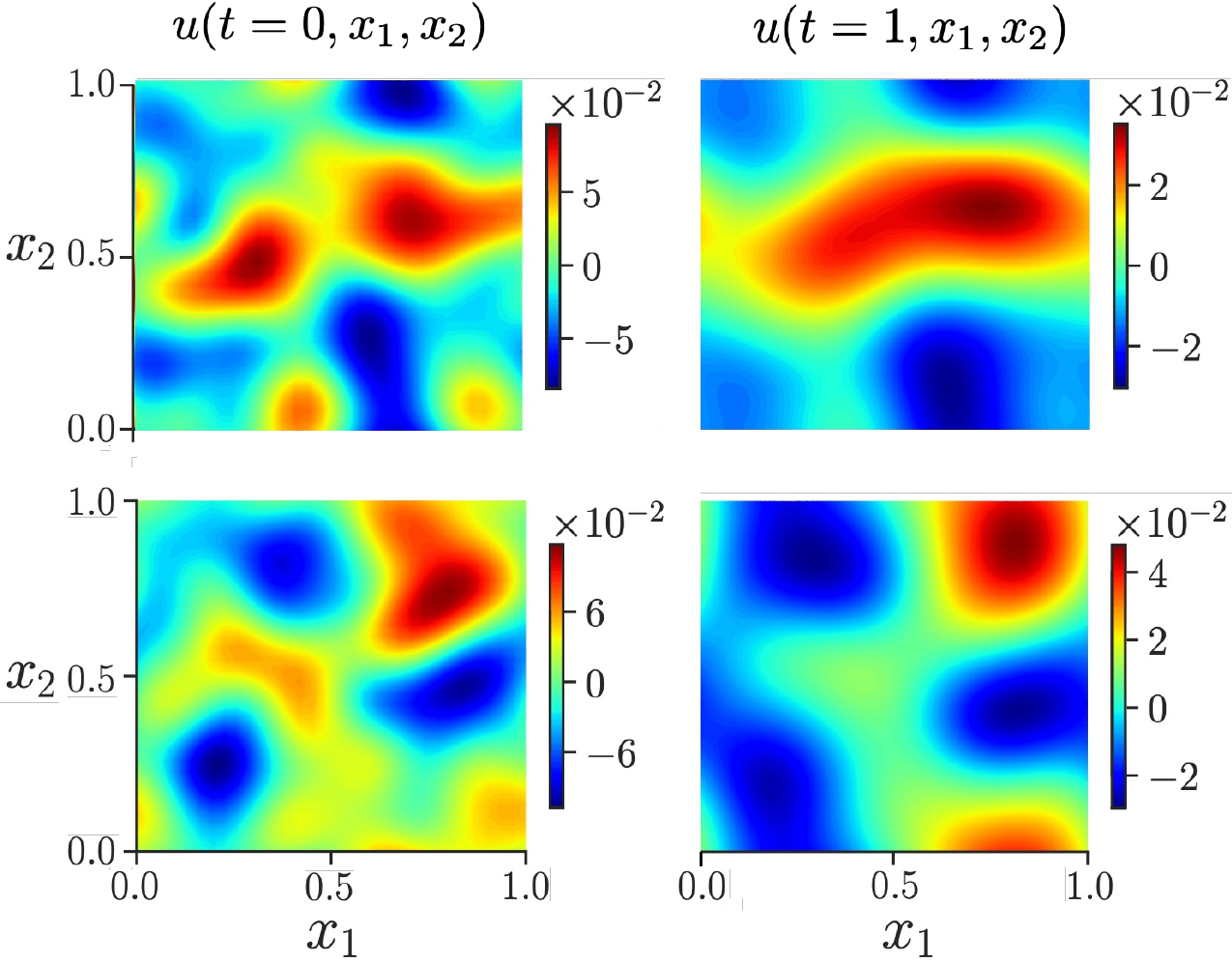} 
    \vspace{3pt}
\end{minipage}\\
\hline
\end{tabular}
\label{tab:examples}
\end{table}

\subsection{Example - 1D Diffusion-Reaction Dynamics}
In this example, we consider a diffusion-reaction system governed by the following equation:
\begin{equation}
\label{eqn:diffusion-reaction}
\begin{aligned}[b]
&\frac{\partial u}{\partial t} = D \frac{\partial^2 u}{\partial x^2} + k u^2 + s(x), \quad (t, x) \in (0, 1] \times (0, 1], \\
&u(0, x) = 0 \ \forall \ x \in (0,1), \\
&u(t, 0) = 0 \ \forall \ t \in (0,1), \\
&u(t, 1) = 0 \ \forall \ t \in (0,1),
\end{aligned}
\end{equation}
where, $D = 0.01$ is the diffusion coefficient and $k = 0.01$ is the reaction coefficient. The source term $s(x)$ is modeled as a random field generated from a Gaussian random process.
The goal is to learn the solution operator that maps these random source terms $s(x)$ to their corresponding solutions $u(t, x)$, i.e., $\mathcal{G}_{{\theta}}: s(x) \to u(t, x)$.

A total of $1{,}500$ input source field functions were generated for this study. Of these, $1{,}000$ $(=n)$ were allocated for training, with ground-truth solutions computed for a randomly selected subset of $n_{\text{train}} \in \{0, 100, 200\}$ input functions. The remaining $n_{\text{test}} = 500$ input functions were reserved for testing, with their corresponding ground-truth solutions also evaluated. Each source function was discretized over $100$ equally spaced spatial points, while the solution fields were discretized over $n_t + 1 = 101$ time points and $n_{\bm{x}} = 100$ spatial points, yielding a spatiotemporal grid of $101 \times 100$ points.

Based on empirical evaluations, we found that setting the output dimension of the Latent-DeepONet to $n_{\bm{z}} = 9$ was sufficient to effectively capture the spatiotemporal dynamics of the solution field. Ablation studies were conducted for different values of $n_{\text{train}}$, as described above. For each setting, the corresponding data-driven loss term was computed using the available ground truth solution fields and the physics-informed loss is evaluated at $(n_{t}^r \times n_{\bm{x}}^r) = 256^2$ collocation points within the solution space in each iteration. Both the PI-Vanilla-NO model and the proposed PI-Latent-NO model were trained for $50{,}000$ iterations. To ensure robustness, each training configuration was repeated across five independent trials with different random seeds.

\begin{figure}[H]
\centering
\includegraphics[width=4.25in, height=2.75in]{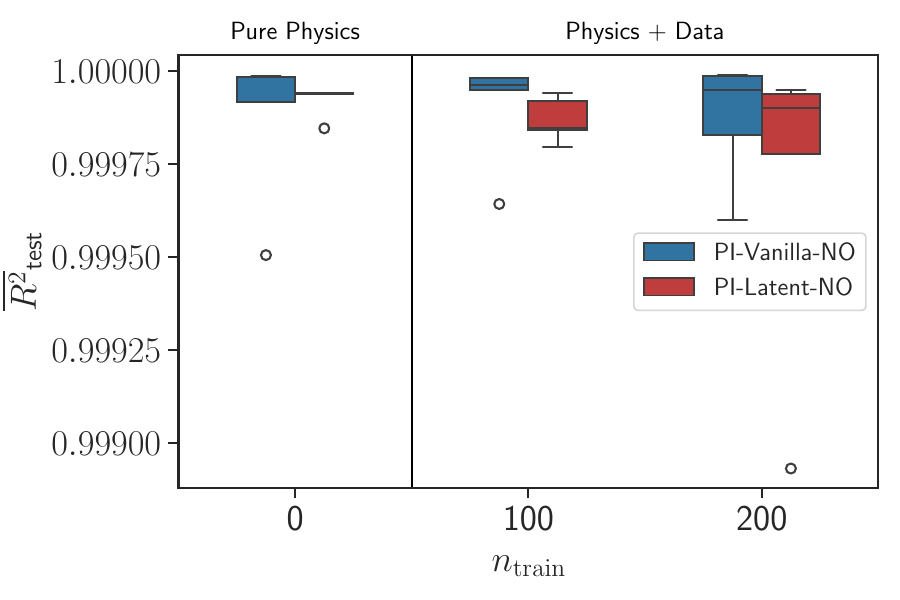}
\vspace{-0.85em}
\\ \hspace{4em}(a)

\includegraphics[width=4.25in, height=2.75in]{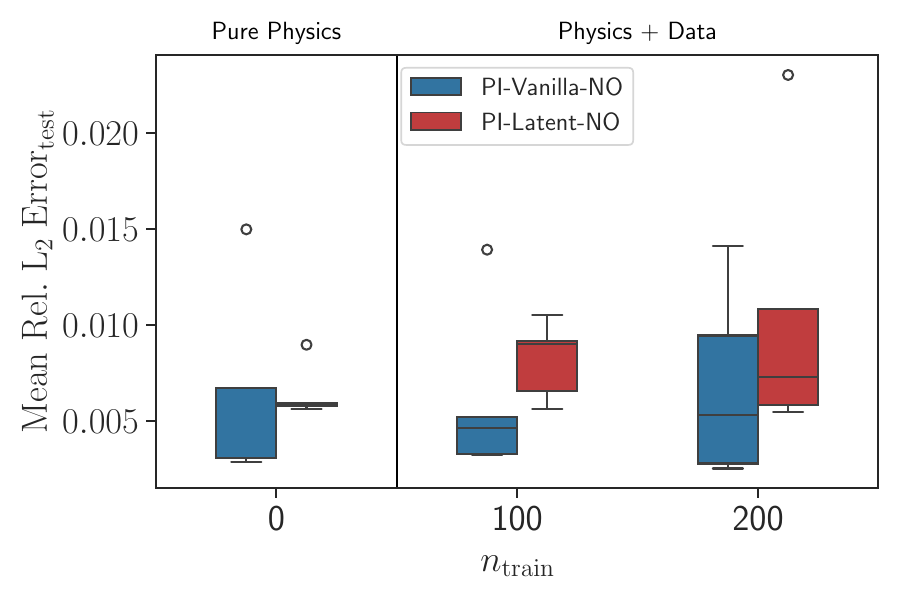}
\vspace{-0.85em}
\\ \hspace{4em}(b)

\includegraphics[width=4.25in, height=2.75in]{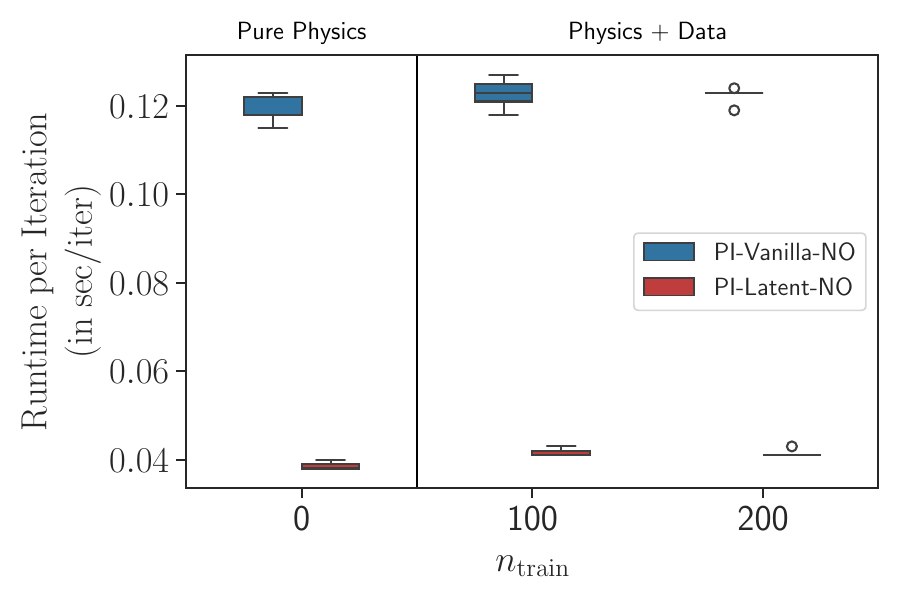}
\vspace{-0.85em}
\\ \hspace{4em}(c)

\caption{1D Diffusion-reaction dynamics: Comparison between the PI-Vanilla-NO and the PI-Latent-NO  (a) mean $R^2$ score of the test data, (b) mean relative $L_2$ error of test data, and (c) training per iteration. The results are based on $5$ independent runs with different seeds, varying the number of training samples $n_{\text{train}}$.}
\label{fig:Example1_boxplots}
\end{figure}

\begin{figure}[H]
    \centering
    \includegraphics[width=0.8\textwidth]{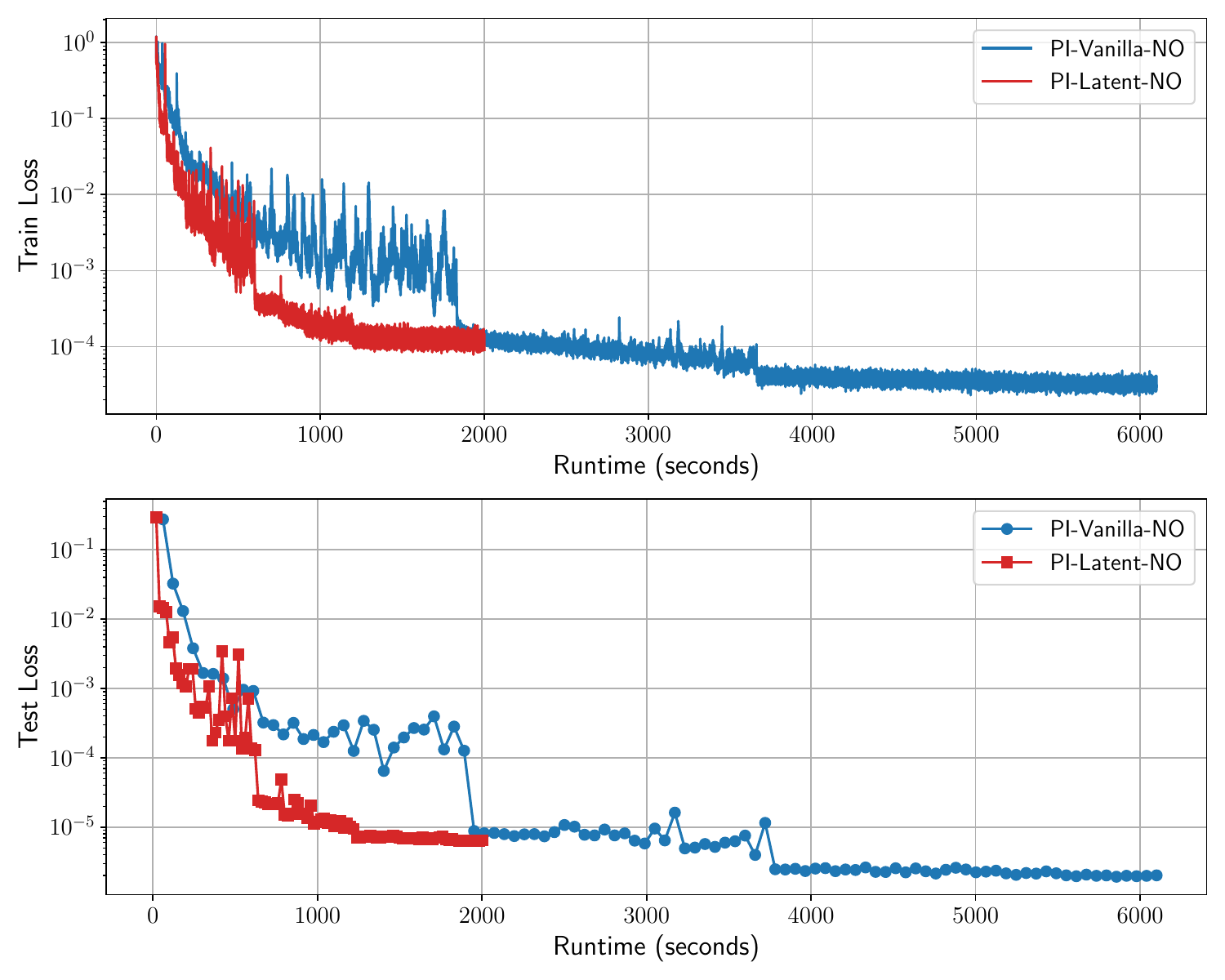}
    \caption{1D Diffusion-reaction dynamics: Comparison of the train and test losses with respect to runtime for models trained in a purely physics-informed manner (i.e., $n_{\text{train}} = 0$).} 
    \label{fig:Example1_train_test_loss_vs_runtime}
\end{figure}

\begin{figure}[H]
    \centering
    \includegraphics[width=\textwidth]{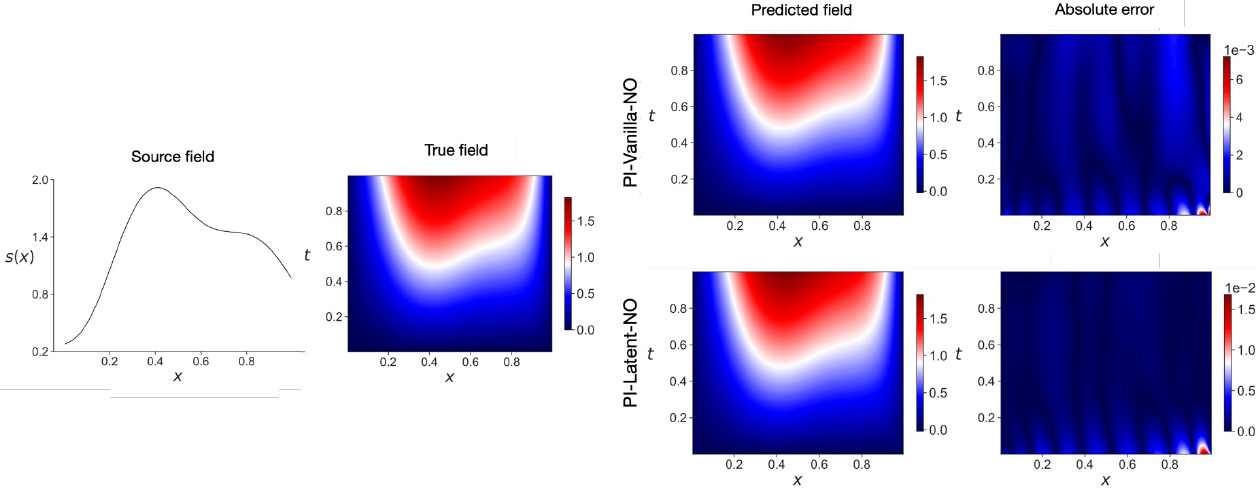}
    \caption{1D Diffusion-reaction dynamics: Comparison of all models on a representative test sample, trained in a purely physics-informed manner (i.e., $n_{\text{train}} = 0$).}
    \label{fig:Example1_sample_realization}
\end{figure}

Table~\ref{tab:Example1_Performance metrics} summarizes the quantitative performance metrics, while Figure~\ref{fig:Example1_boxplots} shows the corresponding box plots for both accuracy and runtime.
These results clearly demonstrate that our model achieves comparable accuracy with significantly reduced runtimes - training times are reduced by almost one-third using our method.
Additionally, Figure~\ref{fig:Example1_train_test_loss_vs_runtime} illustrates that the train and test losses associated with our model decrease at a significantly faster rate than those of the baseline vanilla model.
Our model not only converges more rapidly but also achieves lower loss values within a shorter runtime, underscoring its efficiency and effectiveness in enhancing convergence.
Furthermore, Figure~\ref{fig:Example1_sample_realization} presents a comparison of the models for a representative test sample.

\begin{figure}[H]
\centering
\begin{minipage}{0.48\linewidth}
\centering
\includegraphics[width=3.35in, height=2.2in]{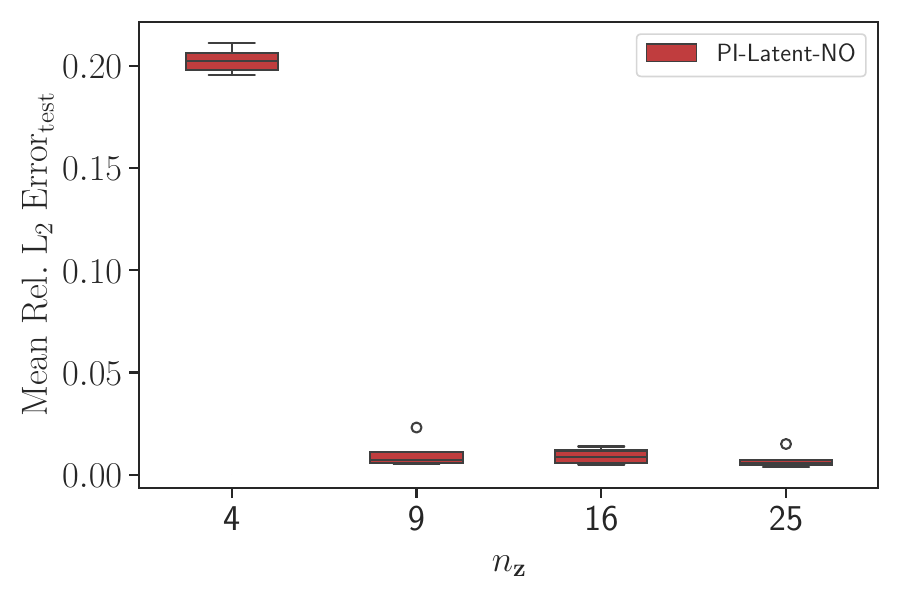}
\vspace{-1.75em}
\\ \hspace{4em} (a)
\end{minipage}
\hfill
\begin{minipage}{0.48\linewidth}
\centering
\includegraphics[width=3.35in, height=2.2in]{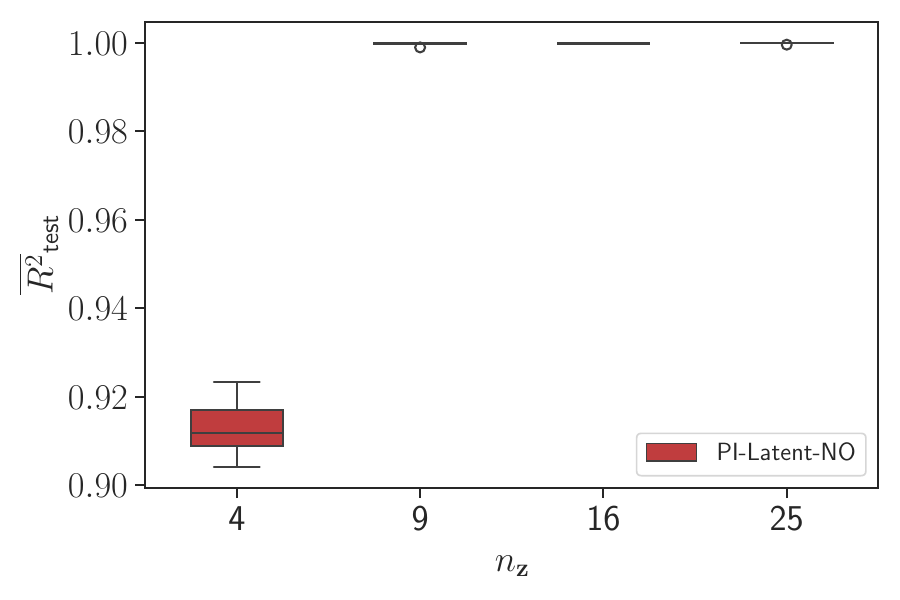}
\vspace{-1.75em}
\\ \hspace{4em}(b)
\end{minipage}
\caption{\SKsays{1D Diffusion-reaction dynamics: Effect of output dimension of the
Latent-DeepONet on the predictive accuracy.}}
\label{fig:Example1_study-latent_dimensionality}
\end{figure}

\SKsays{Following this, we conducted a study to investigate the effect of the Latent DeepONet output dimensionality ($n_{\bm{z}}$) on the predictive performance (see Figure~\ref{fig:Example1_study-latent_dimensionality}). As expected, increasing the latent dimension consistently improves predictive accuracy, since larger latent spaces offer greater representational capacity to capture the underlying solution behavior. However, beyond a certain point, the predictive accuracy plateaus.}

To further understand the runtime and memory usage differences between the two models under varying numbers of collocation points, we carried out two ablation studies.

\begin{figure}[H]
\centering
\begin{minipage}{0.48\linewidth}
\centering
\includegraphics[width=3.35in, height=2.2in]{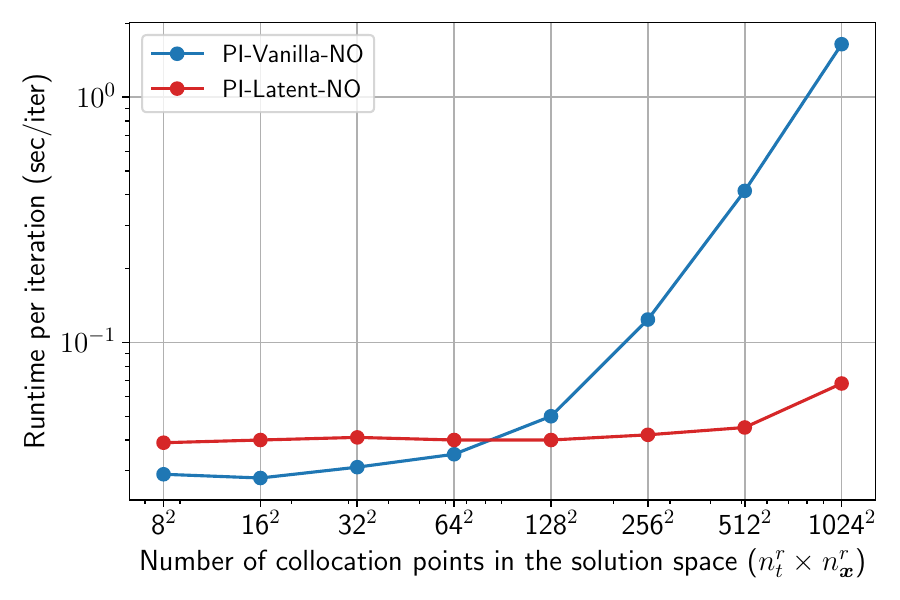}
\vspace{-1.75em}
\\ \hspace{4em} (a)
\end{minipage}
\hfill
\begin{minipage}{0.48\linewidth}
\centering
\includegraphics[width=3.35in, height=2.2in]{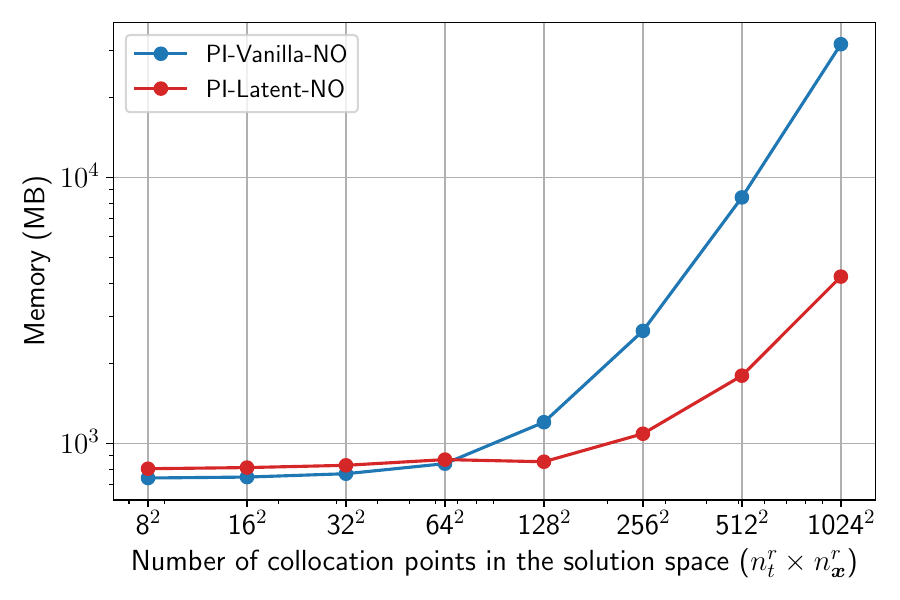}
\vspace{-1.75em}
\\ \hspace{4em}(b)
\end{minipage}
\caption{1D Diffusion-reaction dynamics: Comparison of the PI-Vanilla-NO and the PI-Latent-NO results - (a) runtime per iteration (seconds/iteration), and (b) memory (MB). The results are based on varying the number of collocation points in the solution space and by keeping $n_{\text{train}} = 0$.}
\label{fig:Example1_study1}
\end{figure}

In the first study, we varied the number of collocation points within the solution space $(n_{t}^r \times n_{\bm{x}}^r)$ from $8^2$ to $1024^2$, keeping $n_{\text{train}} = 0$ and $n_{t}^r = n_{\bm{x}}^r = n_{t}^{bc} = n_{\bm{x}}^{ic}$, with $n_{\bm{x}}^{bc} = 1$ for each boundary.
As shown in Figure~\ref{fig:Example1_study1}, the runtime per iteration and memory consumption are nearly independent of the solution space discretization with the PI-Latent-NO model.

\begin{figure}[H]
\centering
\begin{minipage}{0.48\linewidth}
\centering
\includegraphics[width=3.35in, height=2.2in]{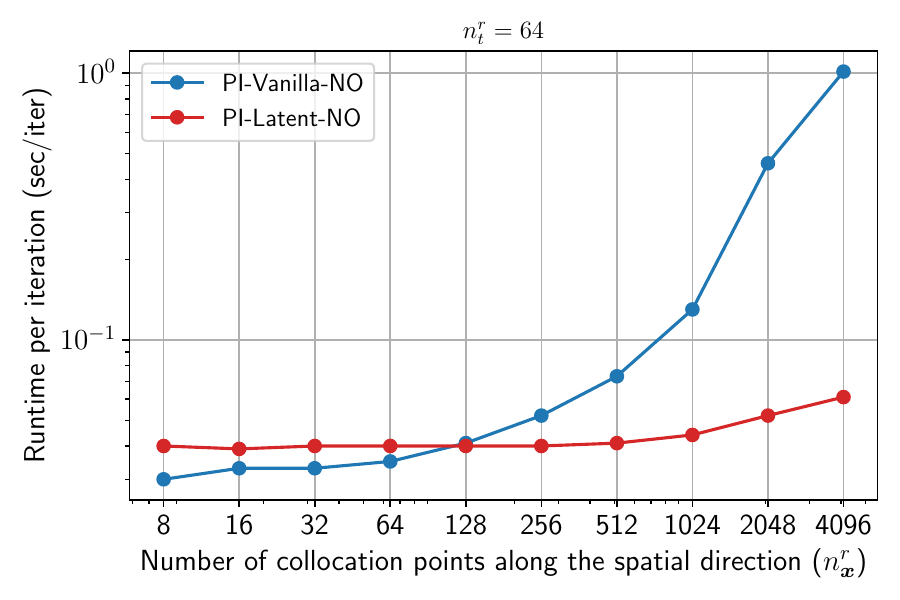}
\vspace{-1.75em}
\\ \hspace{4em} (a)
\end{minipage}
\hfill
\begin{minipage}{0.48\linewidth}
\centering
\includegraphics[width=3.35in, height=2.2in]{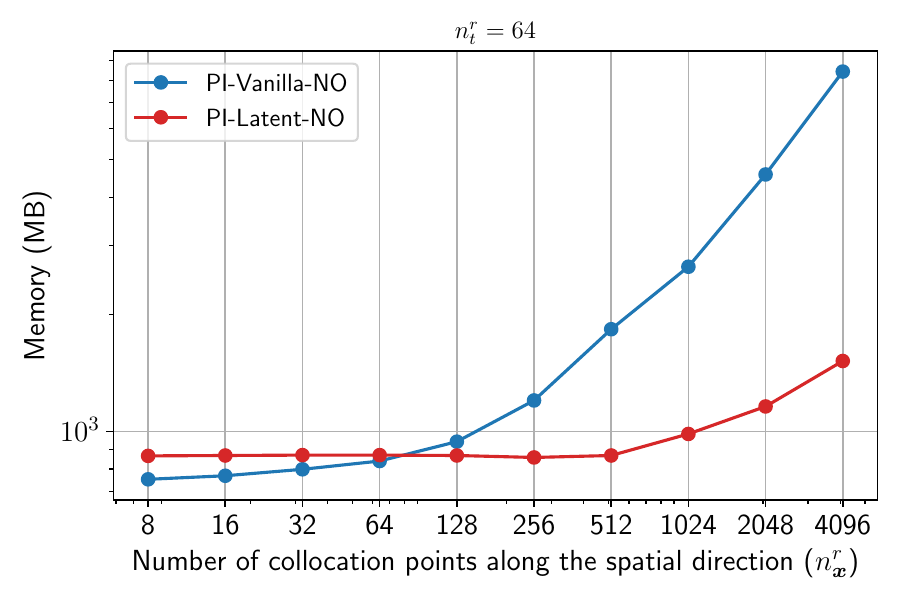}
\vspace{-1.75em}
\\ \hspace{4em}(b)
\end{minipage}
\caption{1D Diffusion-reaction dynamics: Comparison of the PI-Vanilla-NO and the PI-Latent-NO results - (a) runtime per iteration (seconds per iteration), and (b) memory usage (MB). The results are obtained by varying the number of collocation points along the spatial direction while keeping the number of collocation points along the temporal axis fixed at $n_t^r = 64$ and by keeping $n_{\text{train}} = 0$.}
\label{fig:Example1_study2}
\end{figure}

In the second study, the number of collocation points along the temporal axis was fixed at $n_t^r = 64$, while the number along the spatial direction $n_{\bm{x}}^r$ was varied from $8$ to $4096$.
We kept $n_{\text{train}} = 0$, with $n_{t}^{bc} = 64$, $n_{\bm{x}}^{bc} = 1$ for each boundary, and $n_{\bm{x}}^{ic} = n_{\bm{x}}^r$.
Figure~\ref{fig:Example1_study2} shows that the runtime per iteration and memory consumption again remains relatively constant with the proposed method, providing a significant advantage when solving large-scale physical problems with requiring large number of collocation points.

These studies clearly demonstrate that, as the number of spatiotemporal locations used for physics-informed loss evaluation increases, the training time and memory usage for the PI-Vanilla-NO model increase significantly.
In contrast, the training time and memory consumed by the proposed model remain nearly constant.
This consistency in computational efficiency arises from the separability of time and space in this model’s architecture, as previously discussed.

\begin{table}[ht]
\centering
\small
\caption{1D Diffusion-reaction dynamics: Performance metrics}
\label{tab:Example1_Performance metrics}
\begin{tabular}{lcccccc}
\hline
\addlinespace
$\mathrm{Model}$ & $\mathrm{n}_{\mathrm{train}}$ & $\overline{R^2}_{\text{test}}$ & \makecell[c]{Mean Rel. \\ $L_2$ Error$_{\text{test}}$} & \makecell[c]{Training\\Time (sec)} & \makecell[c]{Runtime\\per Iter. (sec/iter)} \\
\addlinespace
\hline
 PI-Vanilla-NO &0 & 0.9999 ± 0.0002 & 0.006 ± 0.005 & 6009 ± 169 & 0.120 ± 0.003 \\ 
 PI-Latent-NO (Ours) &0 & 0.9999 ± 0.0000 & 0.006 ± 0.001 & \textbf{1945 ± 37} & \textbf{0.039 ± 0.001} \\ 
 \hdashline
 PI-Vanilla-NO &100 & 0.9999 ± 0.0001 & 0.006 ± 0.004 & 6142 ± 173 & 0.123 ± 0.003 \\ 
 PI-Latent-NO (Ours) &100 & 0.9999 ± 0.0001 & 0.008 ± 0.002 & \textbf{2080 ± 55} & \textbf{0.042 ± 0.001} \\ 
 \hdashline
 PI-Vanilla-NO &200 & 0.9999 ± 0.0002 & 0.007 ± 0.005 & 6111 ± 92 & 0.122 ± 0.002 \\ 
 PI-Latent-NO (Ours) &200 & 0.9997 ± 0.0004 & 0.010 ± 0.007 & \textbf{2063 ± 46} & \textbf{0.041 ± 0.001} \\ 
\hline
\end{tabular}
\end{table}

\subsection{1D Burgers’ \SKsays{ Equation}}
To highlight the proposed framework's capability to handle non-linearity in governing PDEs, we consider the one-dimensional (1D) Burgers’ equation with periodic boundary conditions:
\begin{equation}
\label{eqn:burger}
\begin{aligned}[b]
\frac{\partial u}{\partial t} + u \frac{\partial u}{\partial x} - \nu \frac{\partial^2 u}{\partial x^2} &= 0, \quad (t, x) \in (0,1) \times (0,1] \\
u(0, x) &= g(x), \quad x \in (0,1) \\
u(t, 0) &= u(t, 1), \\
\frac{\partial u}{\partial x} (t, 0) &= \frac{\partial u}{\partial x} (t, 1),
\end{aligned}
\end{equation}
where, \( t \in (0,1) \), the viscosity \( \nu \) is set to \( \nu = 0.01 \), and the initial condition \( g(x) \) is generated from a Gaussian Random Field (GRF) satisfying the periodic boundary conditions.
The objective is to learn the mapping between the initial condition \( g(x) \) and the solution field \( u(t, x) \), i.e., \( \mathcal{G}_{\boldsymbol{\theta}}: g(x) \to u(t, x) \).

In a manner similar to the previous example, a total of $1{,}500$ initial condition functions were generated for this case. 
Out of these, $1{,}000 (=n)$ functions were designated for training, with ground-truth solutions estimated for a randomly selected subset of $n_{\text{train}} \in \{0, 100, 200\}$ inputs. 
The remaining \( n_{\text{test}} = 500 \) functions were reserved for testing purposes, with ground-truth solutions calculated for them as well.
The initial condition functions are discretized at $101$ equally spaced spatial points and the solution field was resolved across \( n_t + 1 = 101 \) time steps and \( n_{\bm{x}} = 101 \) spatial locations, resulting in a \( 101 \times 101 \) grid.

\begin{figure}[H]
\centering
\includegraphics[width=\textwidth]{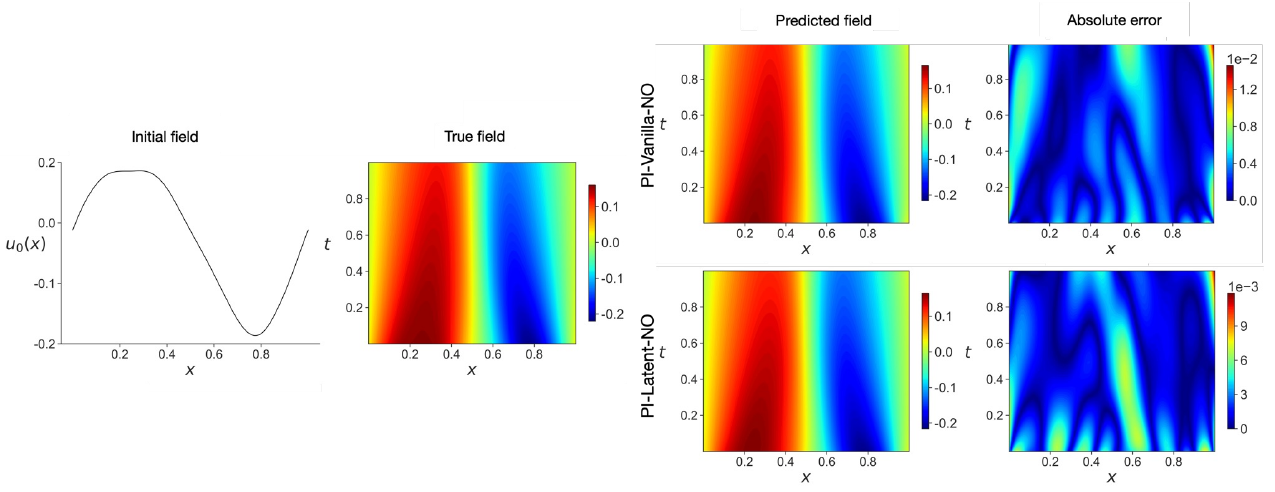}
\caption{\SKsays{1D Burgers’ equation: Model comparison for a representative test sample trained in a purely physics-informed manner (i.e., $n_{\text{train}} = 0$).}}  
\label{fig:Example2_sample_realization}
\end{figure}

Similar to the previous example, empirical results indicate that setting the output dimension of the Latent-DeepONet to $n_{\bm{z}} = 9$ is sufficient to capture the essential spatiotemporal features of the solution field in this case as well.
Both the baseline model and our proposed model were trained using five different random seeds, with the physics-informed loss evaluated at $(n_{t}^r \times n_{\bm{x}}^r) = 512^2$ collocation points in each iteration.
Table~\ref{tab:Example2_Performance metrics} shows the performance metrics, where similar behavior in accuracy is observed, consistent with the previous example. 
Our method  achieving approximately a \SKsays{ $69\%$} reduction in runtime compared to the baseline model.
Figure~\ref{fig:Example2_sample_realization} provides a comparison of model predictions for a representative test sample.

\begin{table}[ht]
\centering
\small
\caption{\SKsays{1D Burgers equation: Performance metrics}}
\label{tab:Example2_Performance metrics}
\begin{tabular}{ccccccc}
\hline
\addlinespace
$\mathrm{Model}$ & $\mathrm{n}_{\mathrm{train}}$ & $\overline{R^2}_{\text{test}}$ & \makecell[c]{Mean Rel. \\ $L_2$ Error$_{\text{test}}$} & \makecell[c]{Training\\Time (sec)} & \makecell[c]{Runtime\\per Iter. (sec/iter)} \\
\hline
 PI-Vanilla-NO &0 & 0.995 ± 0.001 & 0.065 ± 0.007 & 24240 ± 155 & 0.303 ± 0.002 \\ 
 PI-Latent-NO (Ours) &0 & 0.994 ± 0.001 & 0.069 ± 0.005 & \textbf{7519 ± 268} & \textbf{0.094 ± 0.003} \\ 
 \hdashline
 PI-Vanilla-NO &100 & 0.996 ± 0.000 & 0.058 ± 0.003 & 24521 ± 282 & 0.307 ± 0.003 \\ 
 PI-Latent-NO (Ours) &100 & 0.995 ± 0.001 & 0.060 ± 0.005 & \textbf{7734 ± 138} & \textbf{0.097 ± 0.002} \\ 
 \hdashline
 PI-Vanilla-NO &200 & 0.996 ± 0.001 & 0.057 ± 0.004 & 24437 ± 242 & 0.306 ± 0.003 \\ 
 PI-Latent-NO (Ours) &200 & 0.996 ± 0.001 & 0.059 ± 0.004 & \textbf{7718 ± 106} & \textbf{0.097 ± 0.001} \\ 
\hline
\end{tabular}
\end{table}


\subsection{2D Stove-Burner Simulation: Transient Diffusion with Variable Source Geometries}

Next, we consider a transient diffusion system modeling a Stove-Burner scenario, where the objective is to learn the mapping between a spatially distributed heat source of varying geometry and intensity, and the resulting spatiotemporal temperature field. The governing partial differential equation for this system is:
\begin{equation}
\frac{\partial u}{\partial t} = D \left( \frac{\partial^2 u}{\partial x_1^2} + \frac{\partial^2 u}{\partial x_2^2} \right) + s(x_1, x_2, \text{shape}, r, a),
\end{equation}
where \((x_1, x_2) \in \Omega = [-2, 2] \times [-2, 2]\), \(t \in (0, 1]\), and the diffusion coefficient \(D\) is set to 1. Homogeneous Dirichlet boundary conditions \(u(t, x_1, x_2) = 0\) are imposed along the domain boundary \(\partial \Omega\), and the initial condition is fixed as \(u(0, x_1, x_2) = 0\). The source term \(s(x_1, x_2, \text{shape}, r, a)\) is parameterized by a shape type, a burner size parameter \(r \sim \mathcal{U}(0.75, 1.25)\), and an intensity factor \(a \sim \mathcal{U}(5, 15)\). The goal is to learn the operator:
$\mathcal{G}_{\boldsymbol{\theta}}: s(x_1, x_2, \text{shape}, r, a) \mapsto u(t, x_1, x_2)$
that maps input source functions to the corresponding solution fields.

\begin{figure}[H]
    \centering
    \includegraphics[width=0.7\textwidth]{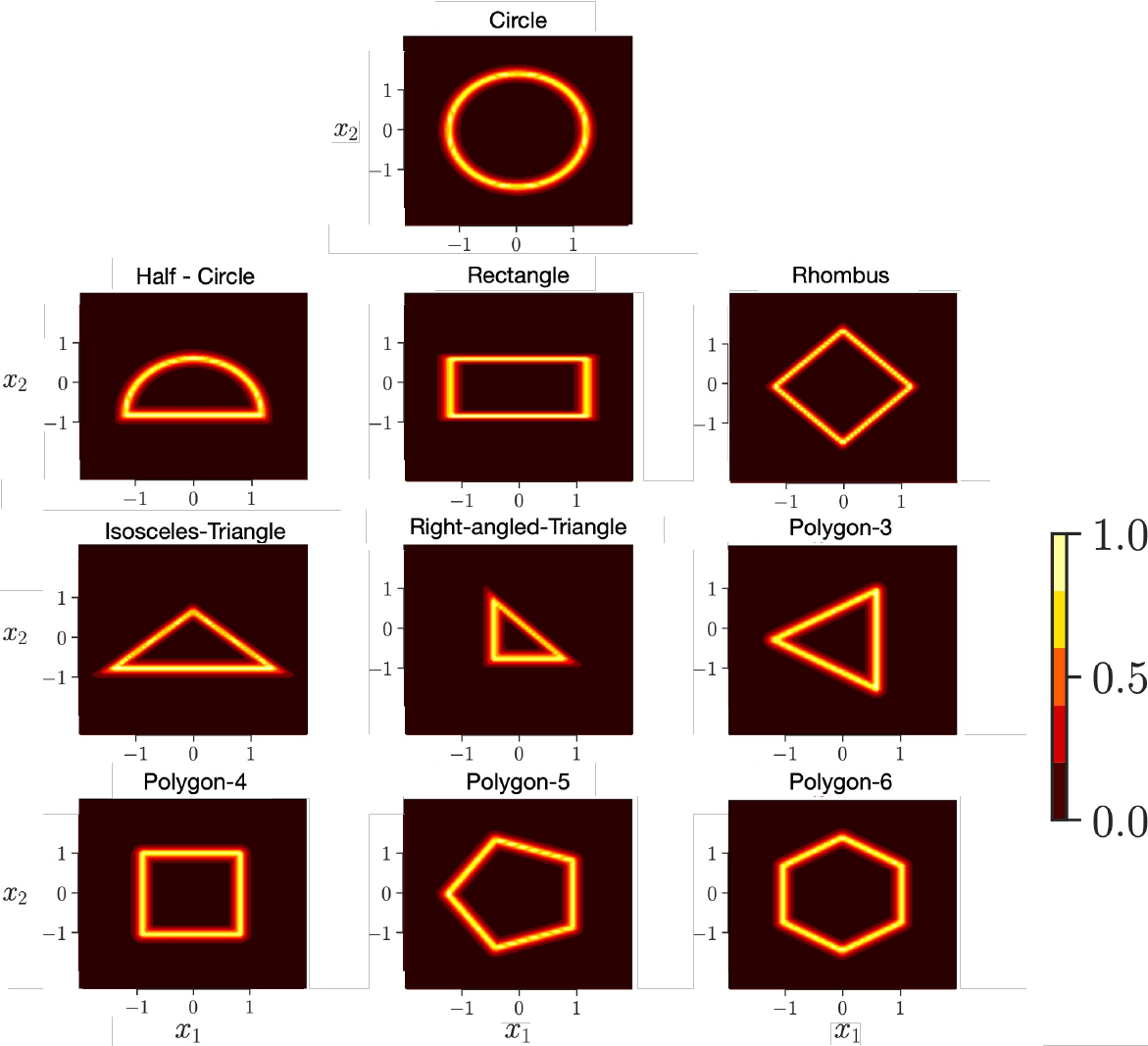}
    \caption{2D Stove-Burner Simulation: Sample representative heat source geometries used in this work.}
    \label{fig:Example3_source-geometries}
\end{figure}

\begin{figure}[H]
    \centering
    \includegraphics[width=0.35\textwidth]{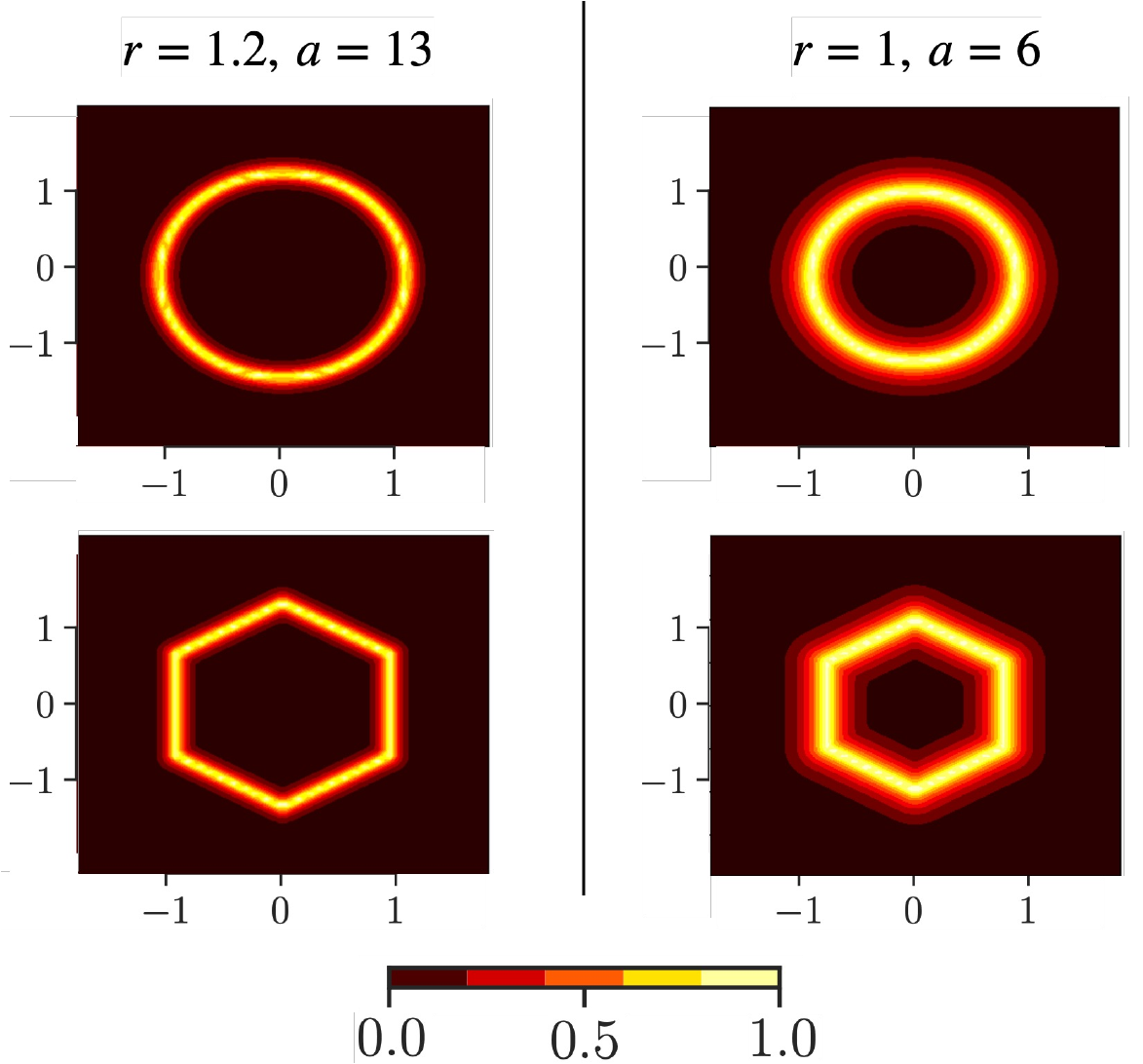}
    \caption{2D Stove-Burner Simulation: Sources of varying shape, size, and intensity. The top row shows circular sources, and the bottom row shows hexagonal sources. The left column corresponds to $(r, a) = (1.2,\ 13)$, and the right column to $(r, a) = (1,\ 6)$.}
    \label{fig:Example3_source-varying_sizes_intensities}
\end{figure}

To explore a wide range of physical configurations, we consider heat sources defined over several distinct geometries (see Figure~\ref{fig:Example3_source-geometries}), including circles, half-circles, squares, rectangles, rhombus, isosceles and right-angled triangles, as well as arbitrary polygons constructed via signed distance functions. Each shape is embedded with smooth exponential decay away from the core region, modulated by the intensity factor \(a\).
Larger values of $a$ result in the intensity being concentrated in a narrower region, while smaller values lead to a broader distribution of intensity—highlighting the influence of the source intensity on the spatial extent of heat deposition (see Figure~\ref{fig:Example3_source-varying_sizes_intensities}).
A comprehensive description of the source function definitions across all geometric configurations is provided in Table~\ref{tab:source_terms}.

A dataset comprising thousands of such source-solution pairs was generated, each defined over a spatial grid of \(64 \times 64\) points and temporal evolution over $20$ uniform time steps. This configuration captures the rich dynamics of heat diffusion under varied spatial forcing profiles. The diversity of source shapes ensures that the learned operator generalizes across distinct topologies and localized heating patterns, providing a robust benchmark for evaluating operator learning models in realistic physical settings.

In modeling the solution field, we did not explicitly enforce the initial and boundary conditions through the physics informed loss. Instead, these conditions are satisfied \emph{a priori} by reparameterizing the solution in a manner that guarantees their fulfillment by construction. Specifically, the predicted solution \(\widehat{u}(s, t, x_1, x_2)\) is expressed as:
\begin{equation}
\widehat{u}(s, t, x_1, x_2) = \mathrm{NN}_{\boldsymbol{\theta}}(s, t, x_1, x_2) \cdot \frac{t(x_1 + L)(L - x_1)(x_2 + L)(L - x_2)}{T (2L)^4},
\end{equation}
where \(L = 2\), \(T = 1\), and \(\mathrm{NN}_{\boldsymbol{\theta}}\) denotes the output of the neural operator parameterized by \(\boldsymbol{\theta}\), which takes as input the source function \(s\), time \(t\), and spatial coordinates \((x_1, x_2)\). The multiplicative factor vanishes at \(t = 0\) and along the spatial boundary \(\partial \Omega\), thereby ensuring that the predicted solution \(\widehat{u}\) exactly satisfies the homogeneous initial and Dirichlet boundary conditions.

As in previous examples, we evaluated the performance of the PI-Latent-NO model on the Stove-Burner simulation. A latent space dimension of $n_{\bm{z}} = 16$ was empirically determined to sufficiently capture the solution’s spatiotemporal dynamics. Experiments were conducted with varying numbers of training samples, $n_{\text{train}} \in \{0, 150, 300\}$, using the same setup as before. For each case, the data-driven loss was computed using ground-truth solutions, and the physics-informed residual loss was evaluated at $20$ random time stamps, with $64^2$ spatial collocation points per time stamp—amounting to a total of $20 \times 64^2 $collocation points per iteration. Both PI-Vanilla-NO and PI-Latent-NO models were trained for $50{,}000$ iterations, with each setting repeated five times using different random seeds to ensure statistical robustness.

\begin{figure}[H]
\centering
\includegraphics[width=4.25in, height=2.75in]{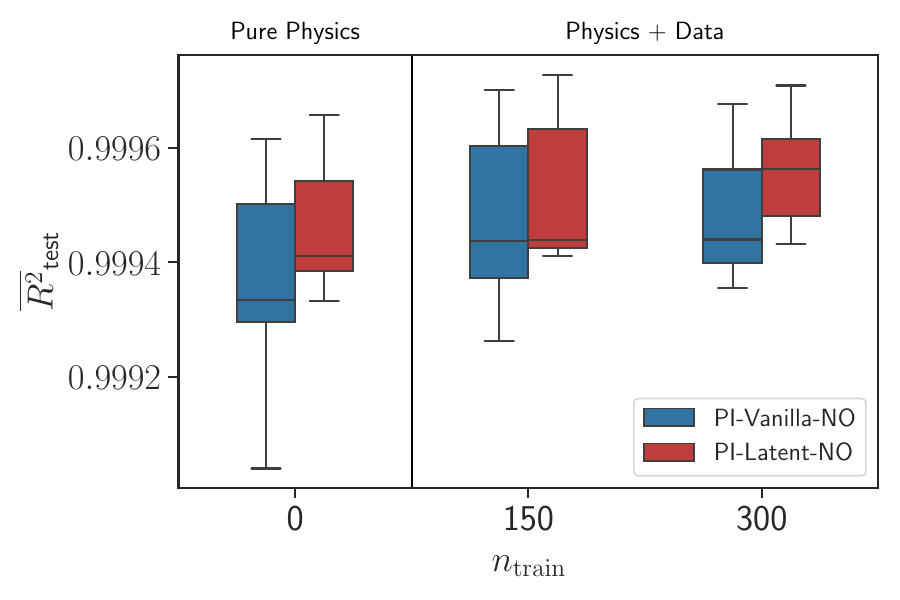}
\vspace{-0.85em}
\\ \hspace{4em}(a)

\includegraphics[width=4.25in, height=2.75in]{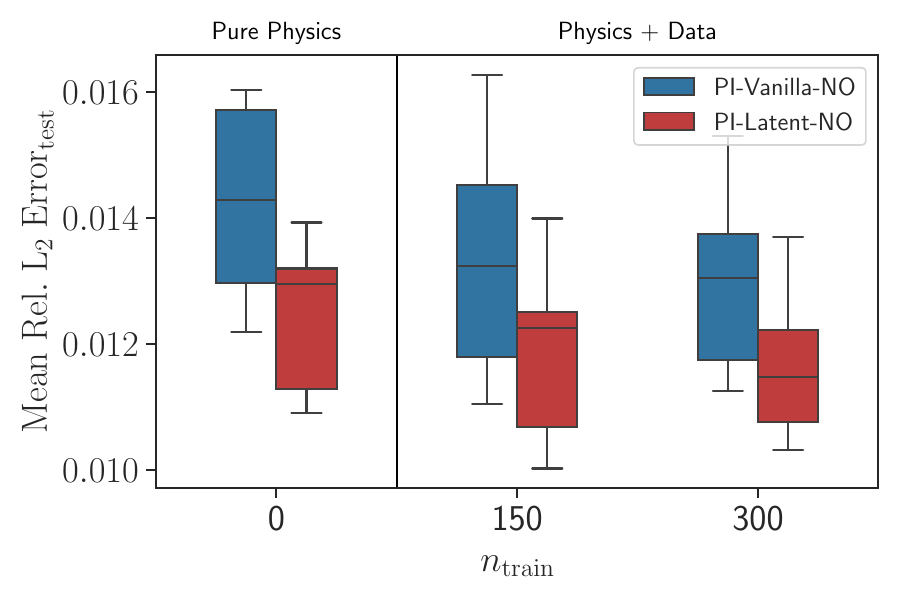}
\vspace{-0.85em}
\\ \hspace{4em}(b)

\includegraphics[width=4.25in, height=2.75in]{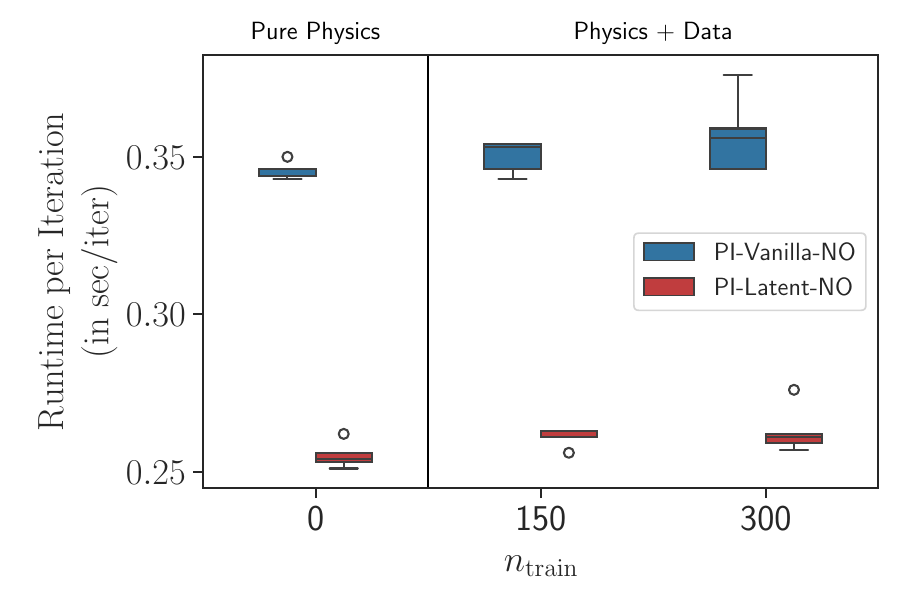}
\vspace{-0.85em}
\\ \hspace{4em}(c)

\caption{2D Stove-Burner Simulation: Comparison between the PI-Vanilla-NO and the PI-Latent-NO  (a) mean $R^2$ score of the test data, (b) mean relative $L_2$ error of test data, and (c) training time per iteration. The results are based on $5$ independent runs with different seeds, varying the number of training samples $n_{\text{train}}$.}
\label{fig:Example3_boxplots}
\end{figure}

\begin{figure}[H]
    \centering
    \includegraphics[width=0.8\textwidth]{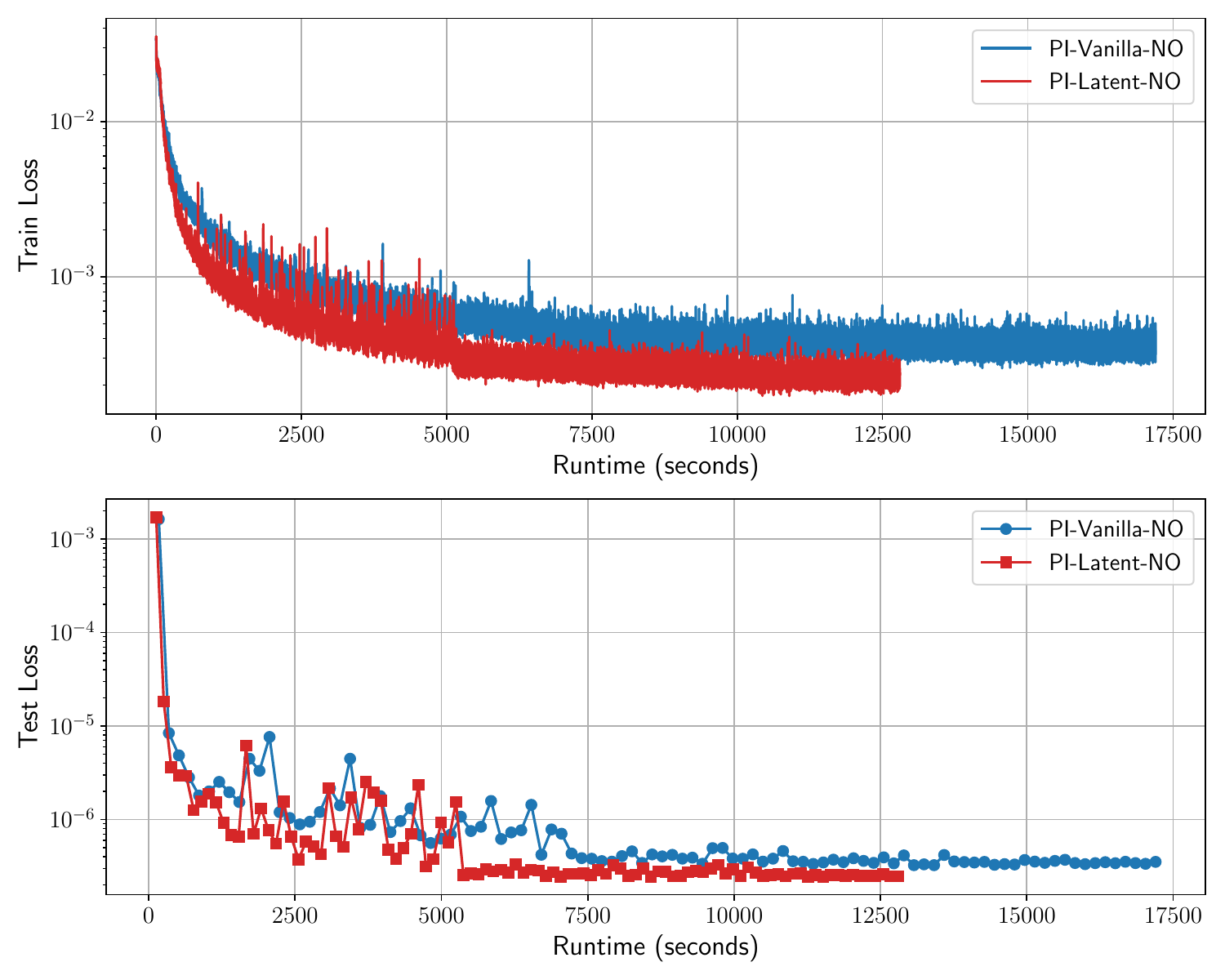}
    \caption{2D Stove-Burner Simulation: Comparison of the train and test losses with respect to runtime for models trained in a purely physics-informed manner (i.e., $n_{\text{train}} = 0$).} 
    \label{fig:Example3_train_test_loss_vs_runtime}
\end{figure}

\begin{figure}[H]
    \centering
    \includegraphics[width=0.95\textwidth]{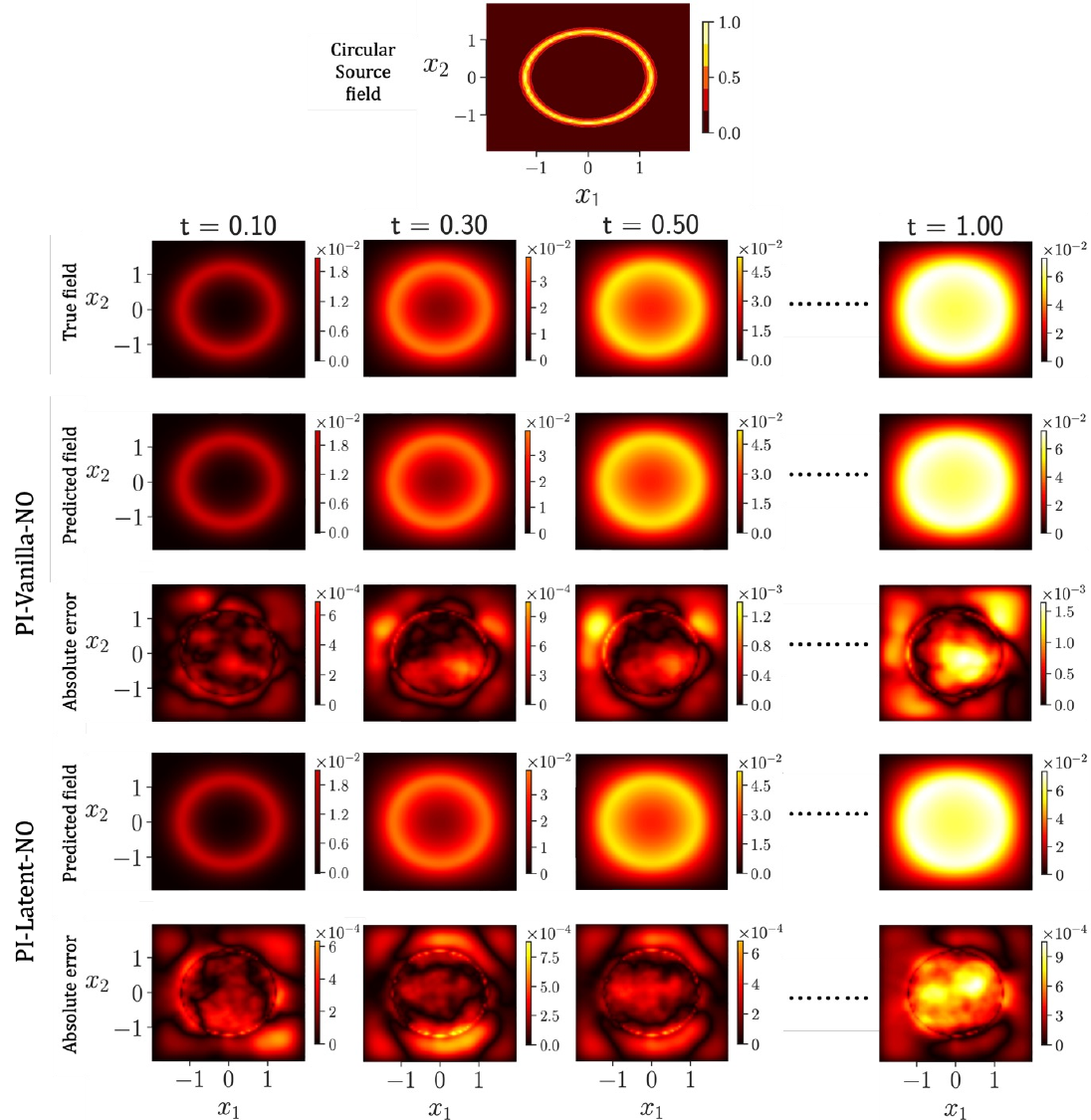}
    \caption{2D Stove-Burner Simulation: Comparison of all models on a representative test sample, trained in a purely physics-informed manner (i.e., $n_{\text{train}} = 0$).}
    \label{fig:Example3_sample_realization_1}
\end{figure}

\begin{figure}[H]
    \centering
    \includegraphics[width= 0.95\textwidth]{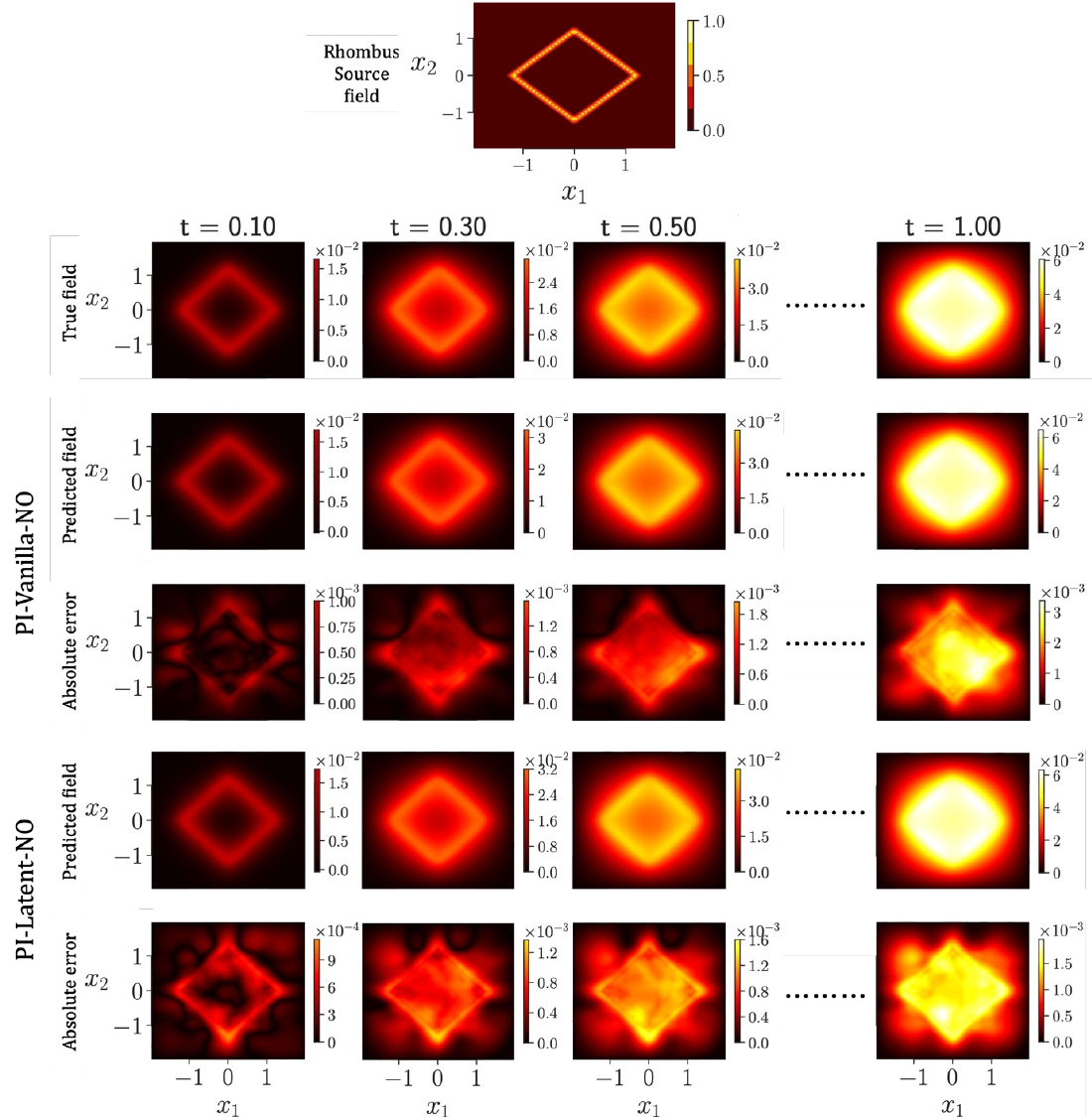}
    \caption{2D Stove-Burner Simulation: Comparison of all models on a representative test sample, trained in a purely physics-informed manner (i.e., $n_{\text{train}} = 0$).}
    \label{fig:Example3_sample_realization_2}
\end{figure}

The performance metrics for these studies are reported in Table~\ref{tab:Example3_Performance metrics} and Figure~\ref{fig:Example3_boxplots}. These results demonstrate that our model achieves comparable predictive performance while substantially reducing training times. In particular, training times are on average $30 \%$ lower with our method compared to the baseline.
Furthermore, Figure~\ref{fig:Example3_train_test_loss_vs_runtime} shows the evolution of train and test losses as a function of runtime. Similar to the 1D - diffusion-reaction example, the PI-Latent-NO model exhibits significantly faster convergence and reaches lower loss values within a shorter runtime. Figures~\ref{fig:Example3_sample_realization_1} and~\ref{fig:Example3_sample_realization_2} illustrate the qualitative performance of both models on a representative test sample.

\begin{figure}[H]
\centering
\begin{minipage}{0.48\linewidth}
\centering
\includegraphics[width=3.35in, height=2.2in]{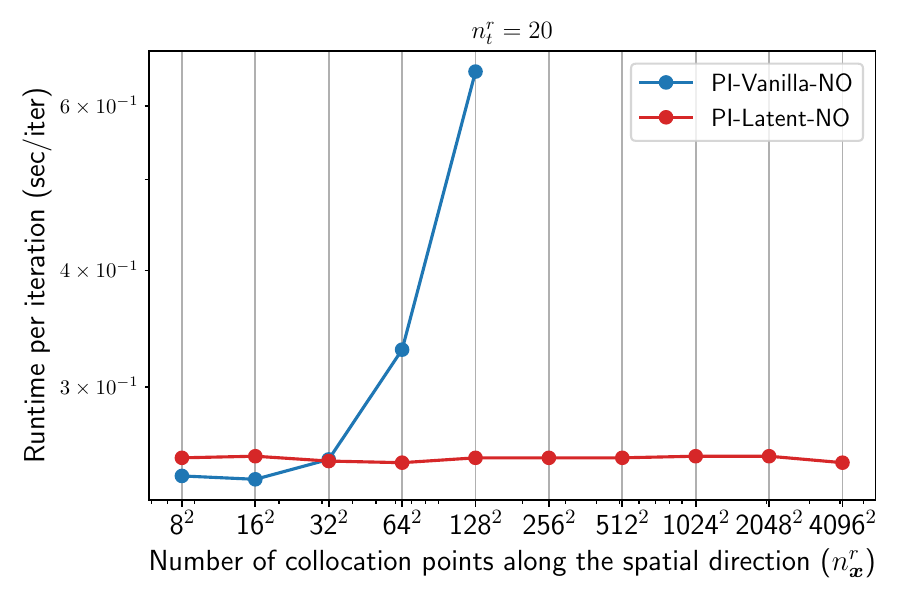}
\vspace{-1.75em}
\\ \hspace{4em} (a)
\end{minipage}
\hfill
\begin{minipage}{0.48\linewidth}
\centering
\includegraphics[width=3.35in, height=2.2in]{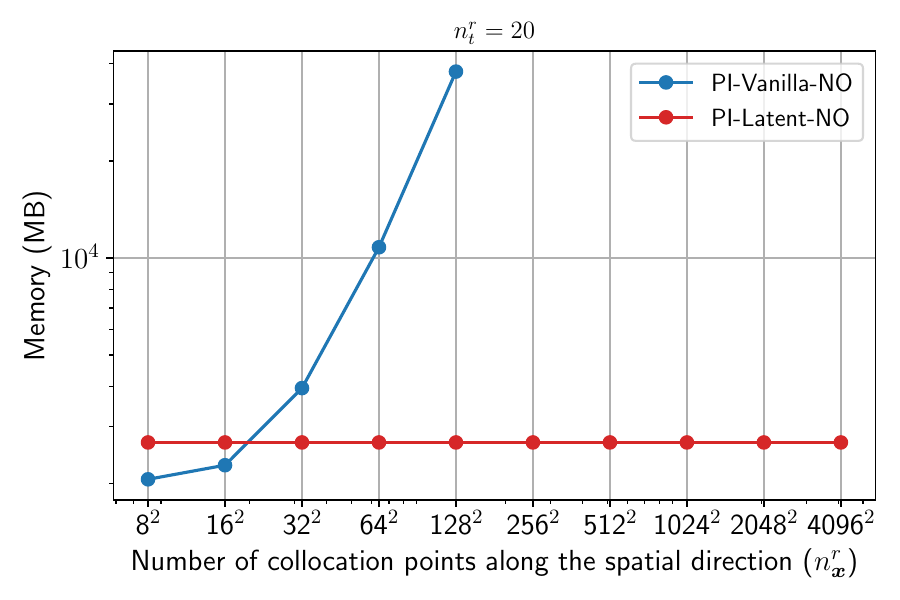}
\vspace{-1.75em}
\\ \hspace{4em}(b)
\end{minipage}
\caption{2D Stove-Burner Simulation: (a) runtime per iteration (seconds per iteration), and (b) memory usage (MB). The results are obtained by varying the number of collocation points along the spatial direction while keeping the number of time evaluations fixed at $n_t^r = 20$ and $n_{\text{train}} = 0$.}
\label{fig:Example3_study2}
\end{figure}

To evaluate the scalability of the proposed model with respect to the number of collocation points, we performed a final ablation study where we fixed the number of collocation points along the temporal direction to be $n_{t}^r = 20$, while varying $n_{x_1}^r \times n_{x_2}^r$ from $8^2$ to $4096^2$,  with $n_{\text{train}} = 0$. 
The results of this study are presented in Figure~\ref{fig:Example3_study2}. The plots illustrate runtime per iteration and memory usage across varying spatial resolutions. In particular, the PI-Vanilla-NO model does not scale to higher spatial resolutions and encounters out-of-memory (OOM) errors as $n_{x_1}^r \times n_{x_2}^r$ increases beyond a moderate threshold. In stark contrast, the PI-Latent-NO model remains robust across all tested resolutions, demonstrating constant memory consumption and stable runtime. This resilience stems from the model’s separable architecture, which decouples the spatial and temporal domains and significantly reduces computational overhead.

These findings reinforce the advantages of the proposed method in handling large-scale PDE problems. The ability to maintain low memory usage and runtime, even with extremely fine spatial discretizations, positions the PI-Latent-NO model as a practical and scalable solution for learning high-resolution scientific simulations in a physics-informed way.

\begin{table}[ht]
\centering
\small
\caption{2D Stove-Burner Simulation: Performance metrics}
\label{tab:Example3_Performance metrics}
\begin{tabular}{ccccccc}
\hline
\addlinespace
$\mathrm{Model}$ & $\mathrm{n}_{\mathrm{train}}$ & $\overline{R^2}_{\text{test}}$ & \makecell[c]{Mean Rel. \\ $L_2$ Error$_{\text{test}}$} & \makecell[c]{Training\\Time (sec)} & \makecell[c]{Runtime\\per Iter. (sec/iter)} \\
\addlinespace
\hline
  PI-Vanilla-NO &0 & 0.9994 ± 0.0002 & 0.014 ± 0.002 & 17276 ± 141 & 0.35 ± 0.00 \\ 
 PI-Latent-NO (Ours) &0 & 0.9995 ± 0.0001 & 0.012 ± 0.001 & \textbf{12759 ± 203} & \textbf{0.26 ± 0.00} \\ 
 \hdashline
 PI-Vanilla-NO &150 & 0.9995 ± 0.0002 & 0.013 ± 0.002 &  17505 ± 258  &  0.35 ± 0.01  \\ 
 PI-Latent-NO (Ours) &150 & 0.9995 ± 0.0001 & 0.012 ± 0.002 & \textbf{13065 ± 155} & \textbf{0.26 ± 0.00} \\ 
  \hdashline
 PI-Vanilla-NO &300 & 0.9995 ± 0.0001 & 0.013 ± 0.002 & 17819 ± 623 & 0.36 ± 0.01 \\ 
 PI-Latent-NO (Ours) &300 & 0.9996 ± 0.0001 & 0.012 ± 0.001 & \textbf{13147 ± 387} & \textbf{0.26 ± 0.01} \\ 
\hline
\end{tabular}
\end{table}

\subsection{2D Burgers’ \SKsays{ Equation}}

Finally, we demonstrate our method’s performance by solving a 2D Burgers’ transport dynamics problem with periodic boundary conditions, where the solution field $u(t, x_1, x_2)$ is a scalar quantity and the initial condition $u_0(x_1, x_2)$ is sampled from a Gaussian Random Field (GRF). The governing equation and the corresponding initial and boundary conditions are given by:
\begin{equation} 
\label{eqn:2dburgers}
\begin{aligned}[b]
\frac{\partial u}{\partial t} + u \frac{\partial u}{\partial x_1} + u \frac{\partial u}{\partial x_2} &= \nu \left( \frac{\partial^2 u}{\partial x_1^2} + \frac{\partial^2 u}{\partial x_2^2} \right), \\
u(0, x_1, x_2) &= u_0(x_1, x_2), \\
u(t, 0, x_2) &= u(t, 1, x_2),\\   
\frac{\partial u}{\partial x_1}\bigg|_{x_1=0} &= \frac{\partial u}{\partial x_1}\bigg|_{x_1=1}, \\
u(t, x_1, 0) &= u(t, x_1, 1),\\ 
\frac{\partial u}{\partial x_2}\bigg|_{x_2=0} &= \frac{\partial u}{\partial x_2}\bigg|_{x_2=1},
\end{aligned}
\end{equation}
where, $(x_1, x_2)\in [0,1]^2, t \in (0,1]$, and the viscosity is set to $\nu = 0.01$. The initial conditions are sampled from a periodic Matérn-type GRF with length scale $l = 0.125$ and standard deviation $\sigma = 0.15$. The objective is to learn the solution operator $\mathcal{G}_{\boldsymbol{\theta}}: u_0(x_1, x_2) \mapsto u(t, x_1, x_2)$ mapping the initial condition to the entire spatio-temporal solution field.

A total of \( n = 1{,}000 \) initial condition functions were generated. Ground truth solutions were computed for 350 of these using a numerical solver. From this set, $50$ solution trajectories were reserved exclusively for testing and performance evaluation, while the remaining $300$ were used to perform ablation studies by varying the number of training samples \( n_{\text{train}} \in \{0, 150, 300\} \). 

Each initial condition was discretized on a $32 \times 32$ spatial grid, and the corresponding solution field was evolved over $21$ uniformly spaced time steps, yielding a spatiotemporal output field of size $21 \times 32 \times 32$. Empirical investigations indicated that a latent dimensionality of $n_{\bm{z}} = 60$ was sufficient to capture the essential features of the spatiotemporal dynamics. During each training iteration, the physics-informed residual loss was evaluated using $21 \times 64 \times 64$ collocation points sampled over the time-space domain. As in the previous experiments, both the baseline and proposed models were trained using five different random seeds to account for variability and assess  statistical robustness.

\textcolor{black}{Further, to effectively capture the spatio-temporal dynamics of the solution field, we employ a Fourier feature expansion of the spatial coordinates $(x_1, x_2)$ in both the trunk networks of the PI-Vanilla-NO model and the reconstruction-deeponet of the proposed PI-Latent-NO model as in \cite{hao2023instability}. This expansion is particularly beneficial for learning smooth yet highly oscillatory solution fields that arise due to the periodic Matérn-type Gaussian random field used to sample the initial conditions. Specifically, the spatial coordinates $(x_1, x_2)$ are transformed as follows:}
\begin{align*}
(x_1, x_2) \mapsto \Big(&x_1, x_2, \cos(\pi x_1), \sin(\pi x_1), \cos(\pi x_2), \sin(\pi x_2), \cos(2\pi x_1), \sin(2\pi x_1), 
\cos(2\pi x_2), \sin(2\pi x_2),  \\ &\dots, \cos(10\pi x_1), \sin(10\pi x_1), \cos(10\pi x_2), \sin(10\pi x_2)\Big)
\end{align*}
\textcolor{black}{resulting in a total of $2 + 4n_f = 42$ features for $n_f = 10$. These Fourier features enhance the expressive capacity of the trunk networks, allowing them to more effectively capture complex spatial patterns, particularly those with high-frequency content.}

\begin{figure}[H]
\centering
\includegraphics[width=\textwidth]{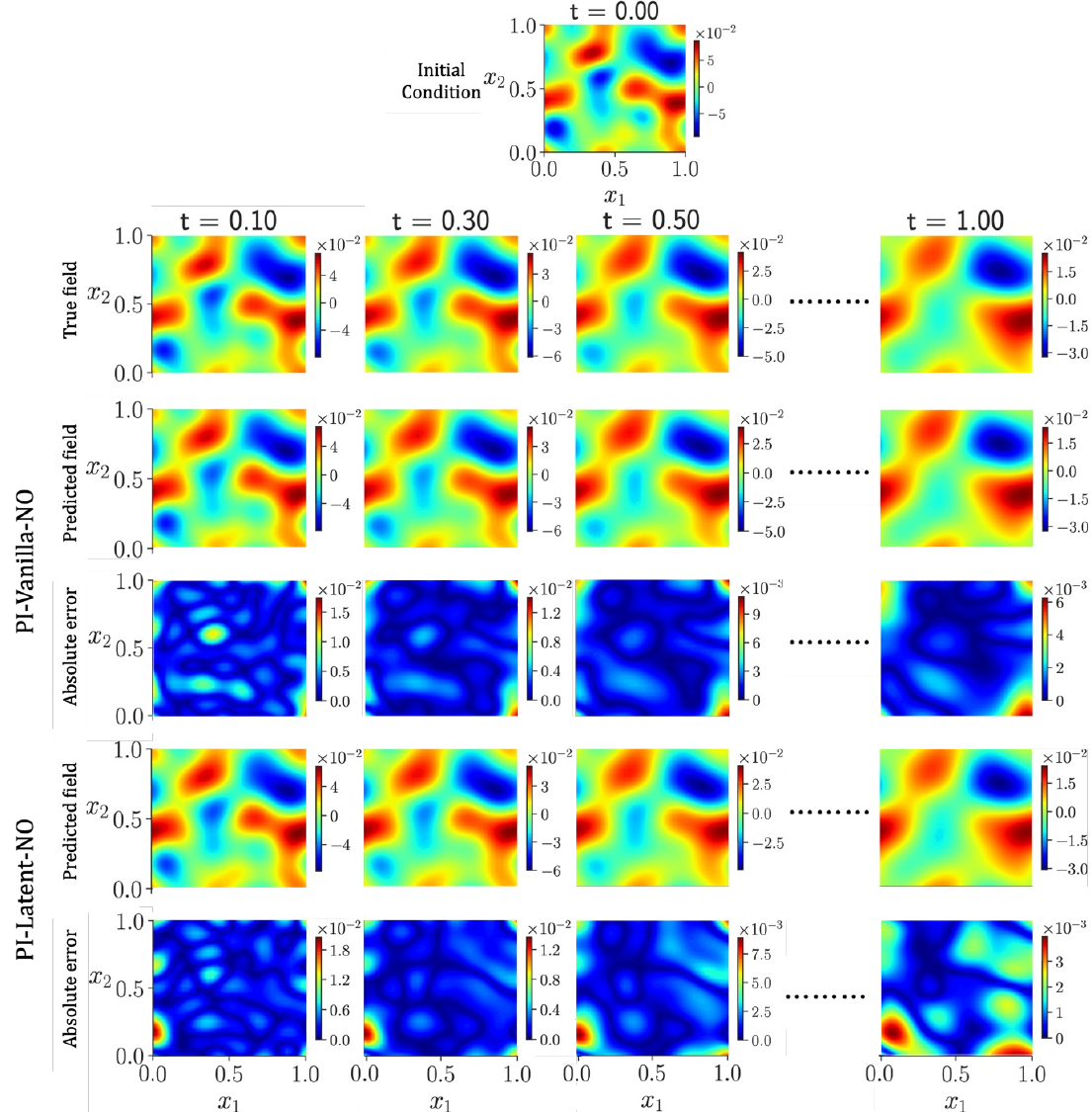}
\caption{2D Burgers’ \SKsays{ equation}: Model predictions for a representative test sample using the physics-informed training with $n_{\text{train}} = 0.$}
\label{fig:Example4_sample_realization_1}
\end{figure}

\begin{figure}[H]
\centering
\includegraphics[width=\textwidth]{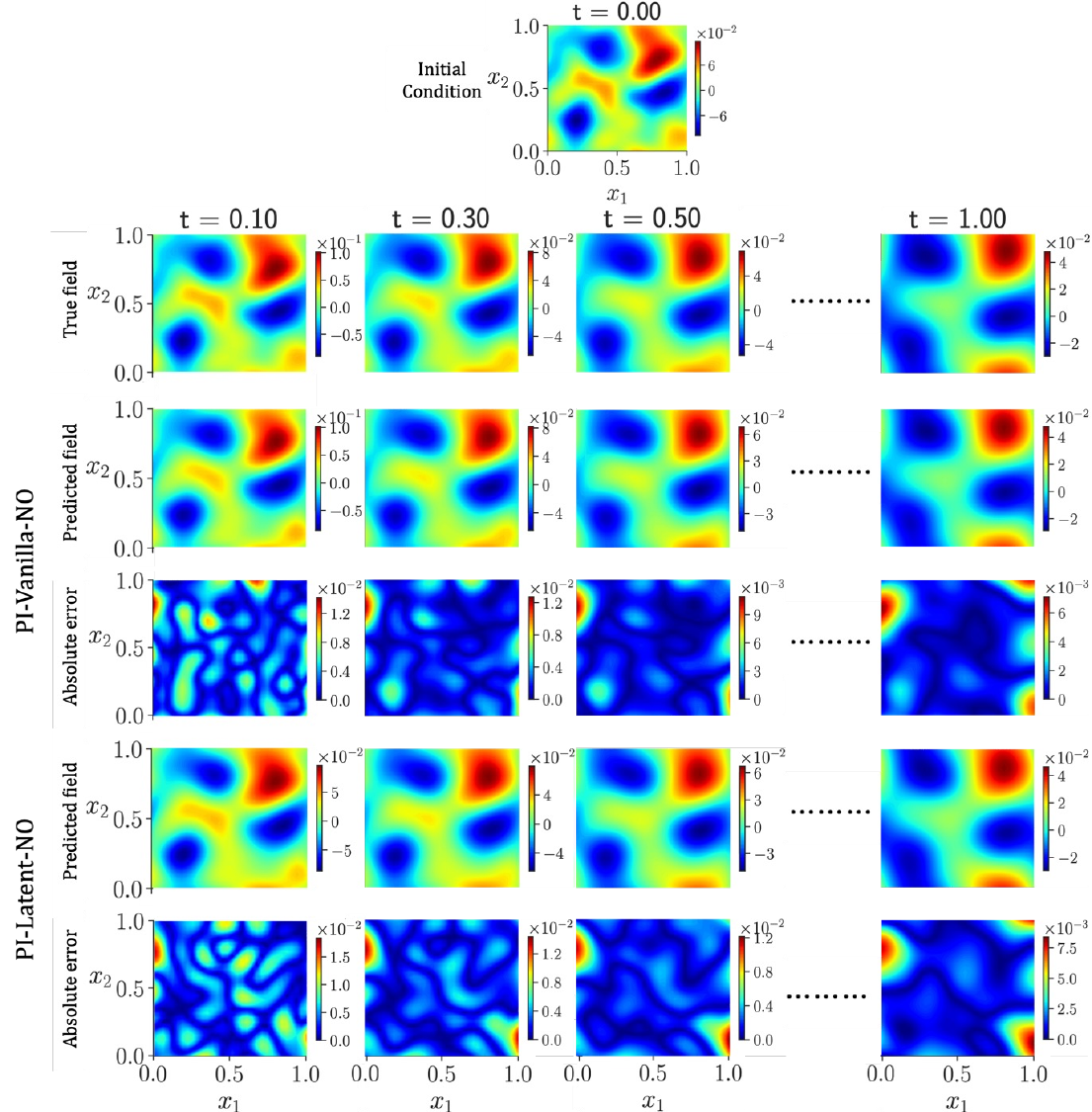}
\caption{2D Burgers’ \SKsays{ equation}: Model predictions for a representative test sample using the physics-informed training with $n_{\text{train}} = 0.$}
\label{fig:Example4_sample_realization_2}
\end{figure}

Results in Table~\ref{tab:Example4_Performance metrics} indicate that our method achieves comparable predictive accuracy while offering approximately on average a $15\%$ reduction in runtime relative to the baseline model. Representative test cases comparing the predictions from the baseline and the proposed models is illustrated in Figure~\ref{fig:Example4_sample_realization_1} and ~\ref{fig:Example4_sample_realization_2}, illustrating the qualitative agreement between predicted and ground-truth fields. While this runtime improvement is encouraging, the gain is relatively modest due to the significant number of derivatives that need to be evaluated as part of the physics-informed residual and boundary condition losses in this case — particularly those involving Neumann boundaries, as evident in the governing equations — which contribute substantially to the overall computational load. Nevertheless, there is room for further improvement. One promising avenue is to incorporate separability in the trunk network of the Reconstruction-DeepONet. This can be achieved by designing dedicated sub-networks for each spatial dimension, thereby leveraging the tensor-product structure of the input space and enhancing computational efficiency. Another possibility, similar to the earlier 2D Stove Burner simulation problem, is to transform the solution field in a way that automatically enforces the initial and boundary conditions, thereby alleviating the need to explicitly compute their associated losses during training.

\begin{table}[ht]
\centering
\small
\caption{2D Burgers \SKsays{ equation}: Performance metrics}
\label{tab:Example4_Performance metrics}
\begin{tabular}{ccccccc}
\hline
\addlinespace
$\mathrm{Model}$ & $\mathrm{n}_{\mathrm{train}}$ & $\overline{R^2}_{\text{test}}$ & \makecell[c]{Mean Rel. \\ $L_2$ Error$_{\text{test}}$} & \makecell[c]{Training\\Time (sec)} & \makecell[c]{Runtime\\per Iter. (sec/iter)} \\
\addlinespace
\hline
 PI-Vanilla-NO &0 & 0.987 ± 0.001 & 0.115 ± 0.003 & 60194 ± 404 & 0.75 ± 0.01 \\ 
 PI-Latent-NO (Ours) &0 & 0.987 ± 0.001 & 0.114 ± 0.004 & \textbf{51543 ± 745} & \textbf{0.64 ± 0.01} \\ 
   \hdashline
 PI-Vanilla-NO &150 & 0.989 ± 0.001 & 0.101 ± 0.003 & 60584 ± 557 & 0.76 ± 0.01 \\ 
 PI-Latent-NO (Ours) &150 & 0.988 ± 0.000 & 0.107 ± 0.002 & \textbf{52125 ± 922} & \textbf{0.65 ± 0.01} \\ 
   \hdashline
 PI-Vanilla-NO &300 & 0.990 ± 0.000 & 0.099 ± 0.002 & 60219 ± 150 & 0.75 ± 0.00 \\ 
 PI-Latent-NO (Ours) &300 & 0.989 ± 0.000 & 0.103 ± 0.001 & \textbf{52504 ± 750} & \textbf{0.66 ± 0.01} \\ 
\hline
\end{tabular}
\end{table}

\SKsays{\section{Discussion on Computational Cost}
\label{sec:computational cost}}

\textcolor{black}{In order to compare computational cost, we first consider the 1D diffusion–reaction dynamics example introduced in Section 4.1. In particular, we examine the case trained solely using the physics-informed loss ($n_{\text{train}} = 0$), which incurs no cost for data generation. Training the PI-Latent-NO model required approximately $1{,}996$ seconds. Once trained, the inference cost for each new case is only about $0.01$ seconds, whereas a conventional numerical solver requires roughly $0.4047$ seconds to generate a single high-fidelity simulation.} 
\textcolor{black}{The total computational cost for PI-Latent-NO method can thus be expressed as: 
\[
C_{\text{PI-Latent-NO}} = 1{,}996+ 0.01N,
\]  
where $N$ is the number of testing cases evaluated beyond the training data. The traditional solver cost is expressed as:
\[
C_{\text{solver}} = 0.4047N.
\]}  
\textcolor{black}{Equating the two gives a breakeven point of  
\[
1{,}996+ 0.01N = 0.4047N \quad \Rightarrow \quad N \approx 5{,}056.
\]  
This implies that when more than roughly $5{,}056$ simulations are required, PI-Latent-NO becomes computationally more efficient than the conventional solver. Such a threshold is commonly exceeded in uncertainty quantification, optimization, and real-time decision-making workflows.}

\begin{figure}[H]
\centering
\begin{minipage}{0.49\linewidth}
\centering
\includegraphics[width=3.4in, height=2.25in]{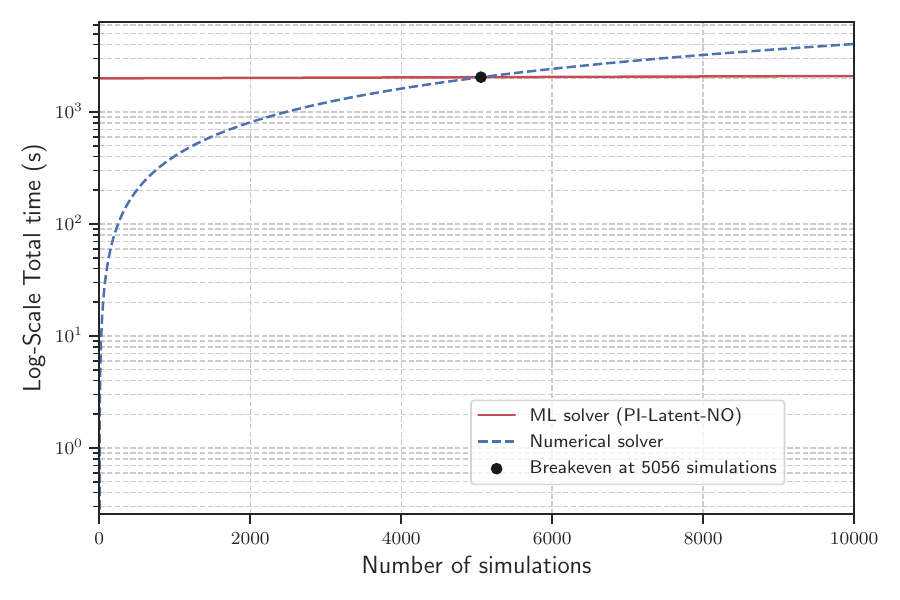}
\vspace{-1.75em}
\\ \hspace{4em} (a)
\end{minipage}
\hfill
\begin{minipage}{0.49\linewidth}
\centering
\includegraphics[width=3.4in, height=2.25in]{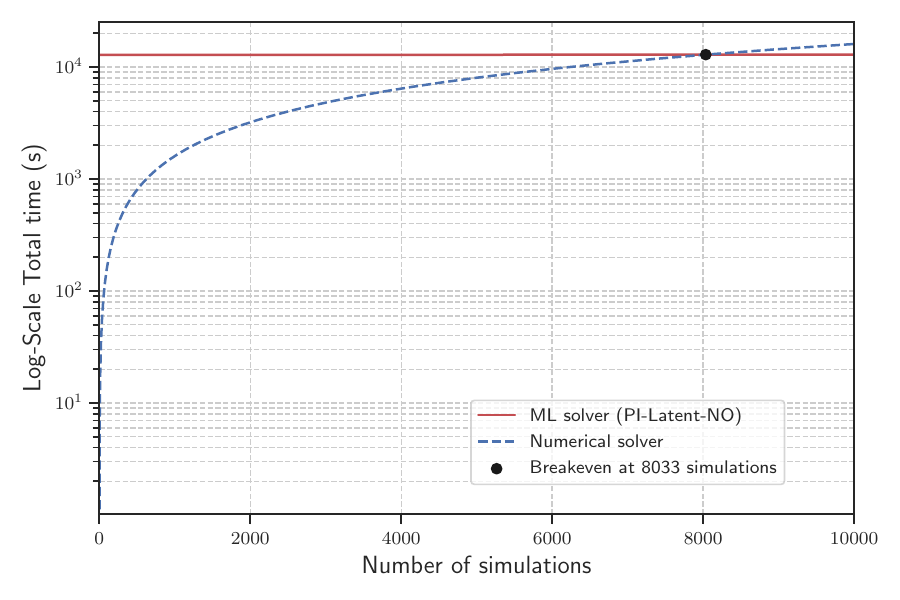}
\vspace{-1.75em}
\\ \hspace{4em}(b)
\end{minipage}
\caption{\SKsays{Breakeven analysis of total computational time comparing the PI-Latent-NO solver with the traditional numerical solver for (a) 1D diffusion–reaction dynamics and (b) 2D stove–burner simulation.}}
\label{fig:breakeven_analysis}
\end{figure}

\textcolor{black}{A similar analysis was carried out for the 2D stove–burner simulation discussed in Section 4.3. The results show that after approximately $8{,}033$ numerical simulations, the ML-based approach becomes the cheaper alternative. Figure~\ref{fig:breakeven_analysis} presents the comparative breakeven analysis for both problems. The computational cost of the numerical solver scales linearly with the number of simulations, while the PI-Latent-NO approach incurs an upfront offline cost (including both training data generation and model training), appearing as a fixed offset. Once trained, however, the inference cost of PI-Latent-NO is nearly negligible (approximately $0.01$ seconds per simulation).}

\textcolor{black}{Note that the numerical solver used in this breakeven analysis is the same as the one employed during the data generation process. For the 1D case, the model achieves excellent agreement with the numerical reference, indicating its ability to capture the underlying dynamics accurately. However, in higher-dimensional problems such as the 2D Burgers’ equation small discrepancies remain, which we plan to address in future work.}
\newline
\section{Quantitative Assessment of Physical Consistency via Plotting Latent dynamics}
\label{sec:latent dynamics}

\textcolor{black}{To substantiate the claim that the learned latent representations capture key physical behaviors and evolution patterns, we conducted a quantitative analysis of the latent dynamics across all four benchmark problems. For each case, we visualized the temporal evolution of the latent features and their corresponding trajectories in the first two principal component (PC) dimensions.}

\begin{figure}[H]
    \centering
    \includegraphics[trim = 0 0 0 20, clip, width=0.7\textwidth]{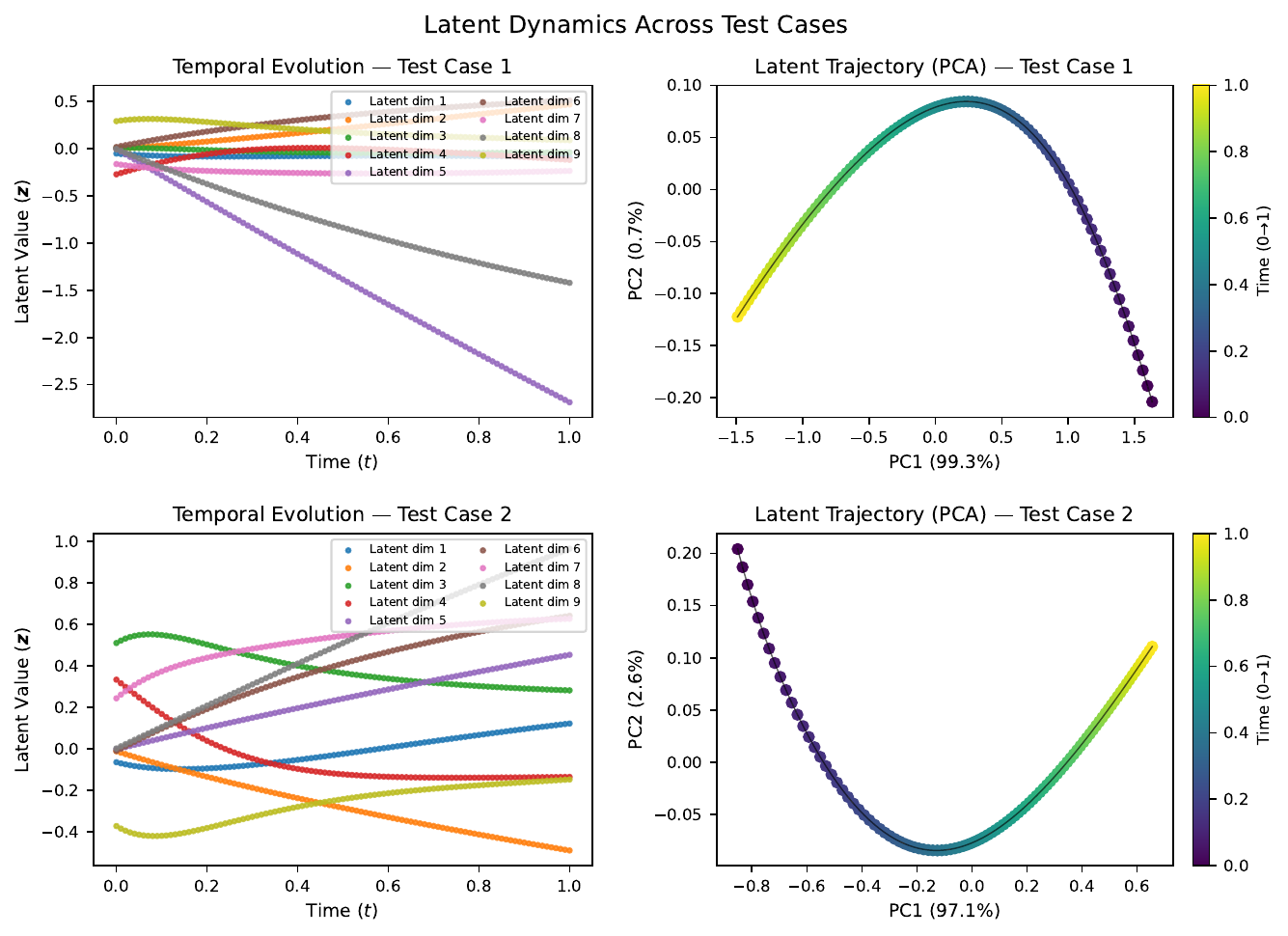}
    \caption{\textcolor{black}{Latent dynamics for the 1D Diffusion–reaction problem. The left panel shows the temporal evolution of individual latent dimensions over the time domain ($t \in [0,1]$), revealing smooth, continuous variations consistent with the diffusive behavior. The right panel displays the corresponding latent trajectories projected onto the first two principal components (PCs), which capture over 95\% of the latent variance. The low-dimensional parabolic manifold indicates that the latent space effectively encodes the dominant temporal modes of diffusion-reaction.}}
    \label{fig:Latent Dynamics - 1D Diffusion}
\end{figure}

\begin{figure}[H]
    \centering
    \includegraphics[trim = 0 0 0 20, clip, width=0.7\textwidth]{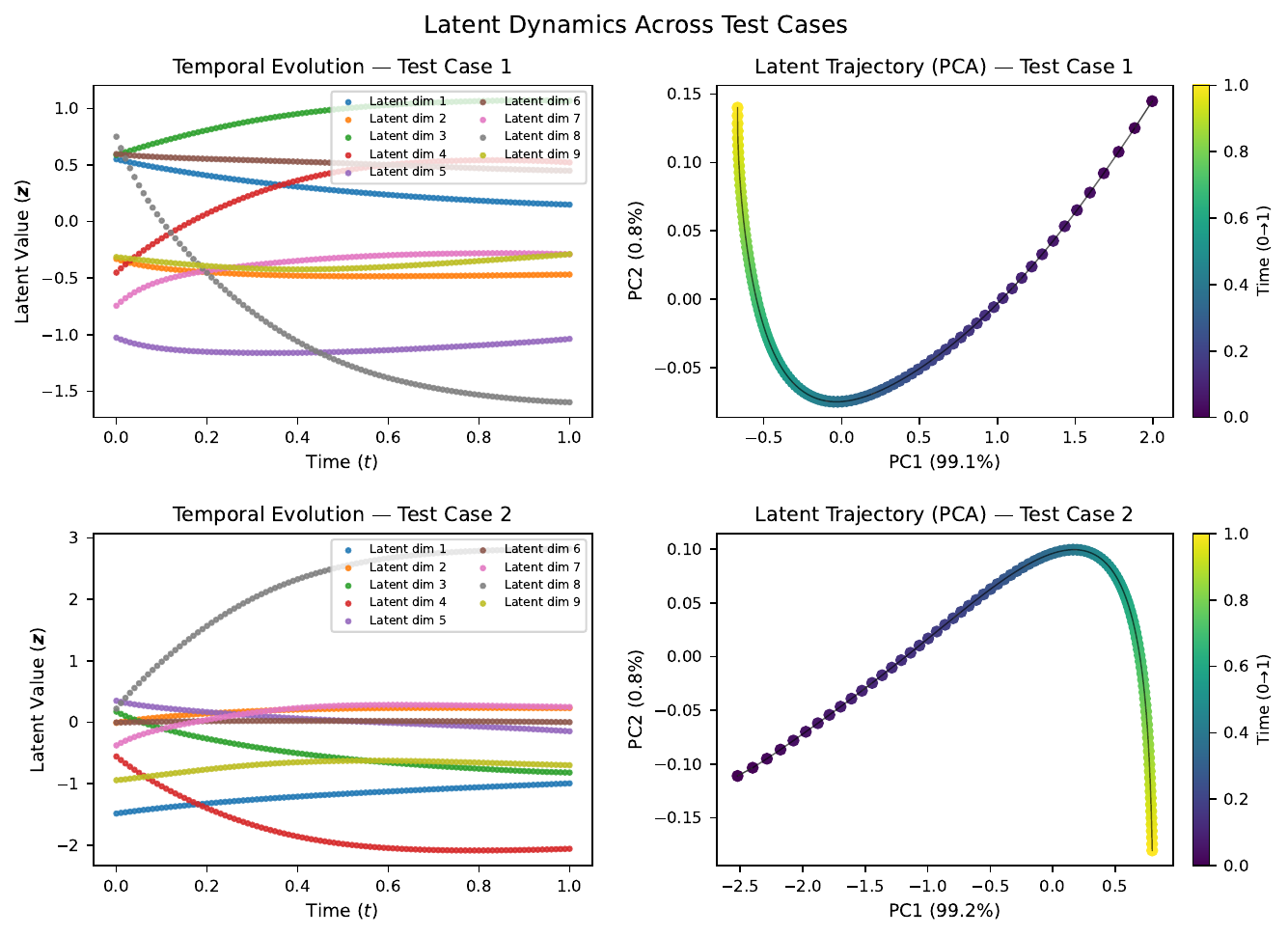}
    \caption{\textcolor{black}{Latent dynamics for the 1D Burgers’ equation benchmark. Temporal evolution plots (left) exhibit smooth latent transitions reflecting advective–diffusive dynamics without abrupt fluctuations. The 2D PCA projection (right) shows a compact, continuous trajectory, indicating that the learned latent manifold captures the nonlinear evolution of Burgers’ flow within a low-dimensional subspace.}}
    \label{fig:Latent Dynamics - 1D Burgers'}
\end{figure}

\begin{figure}[H]
    \centering
    \includegraphics[trim = 0 0 0 20, clip, width=0.7\textwidth]{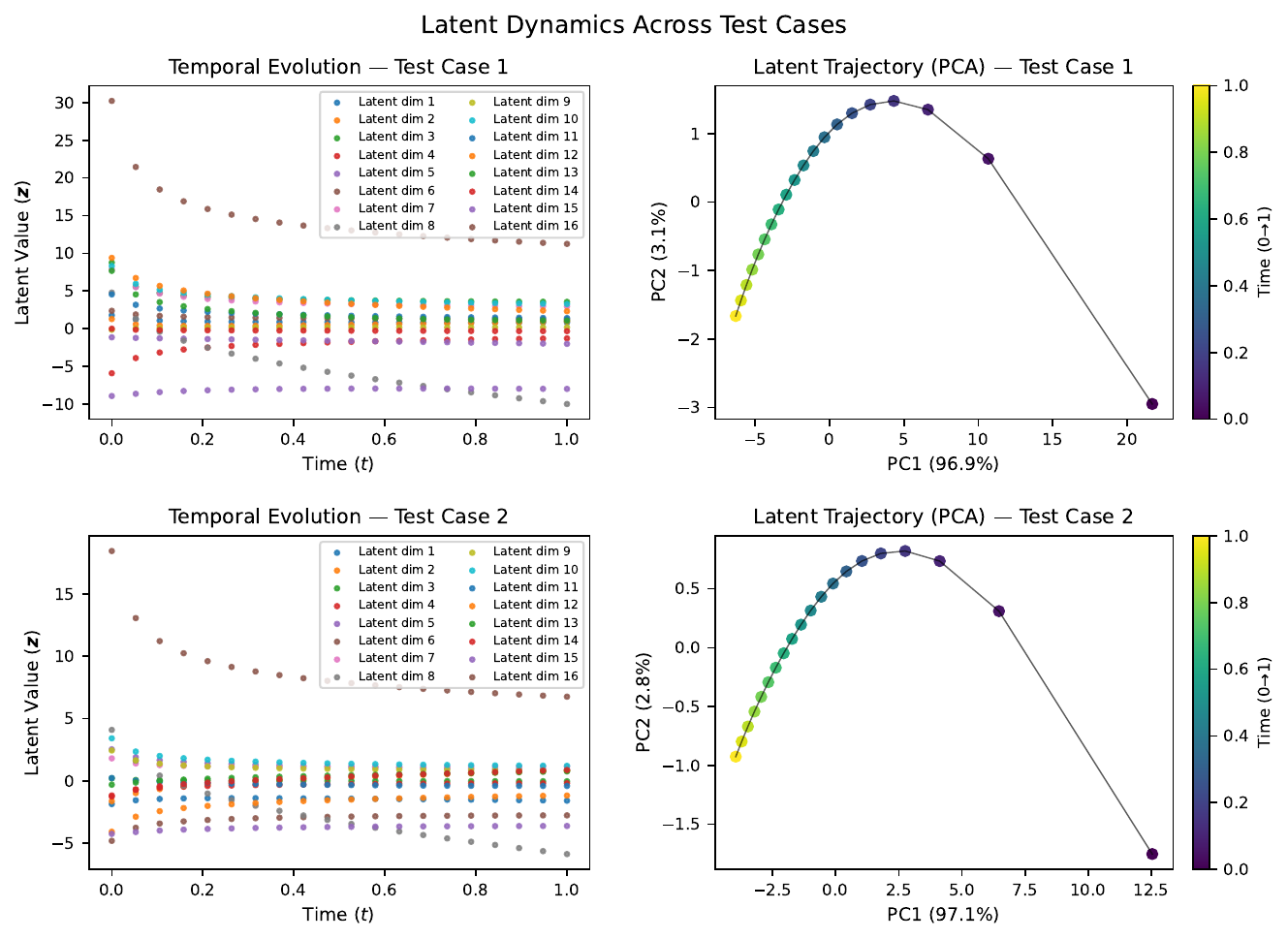}
    \caption{\textcolor{black}{Latent dynamics for the 2D Stove–burner simulation. Despite the higher spatial and temporal complexity, the latent variables evolve coherently over time (left), while the PCA-projected trajectories (right) remain organized and continuous. This demonstrates that the learned latent representation generalizes to complex thermal systems while maintaining physical consistency.}}
    \label{fig:Latent Dynamics - 2D Stove Burner Simulation}
\end{figure}

\begin{figure}[H]
    \centering
    \includegraphics[trim = 0 0 0 20, clip, width=0.7\textwidth]{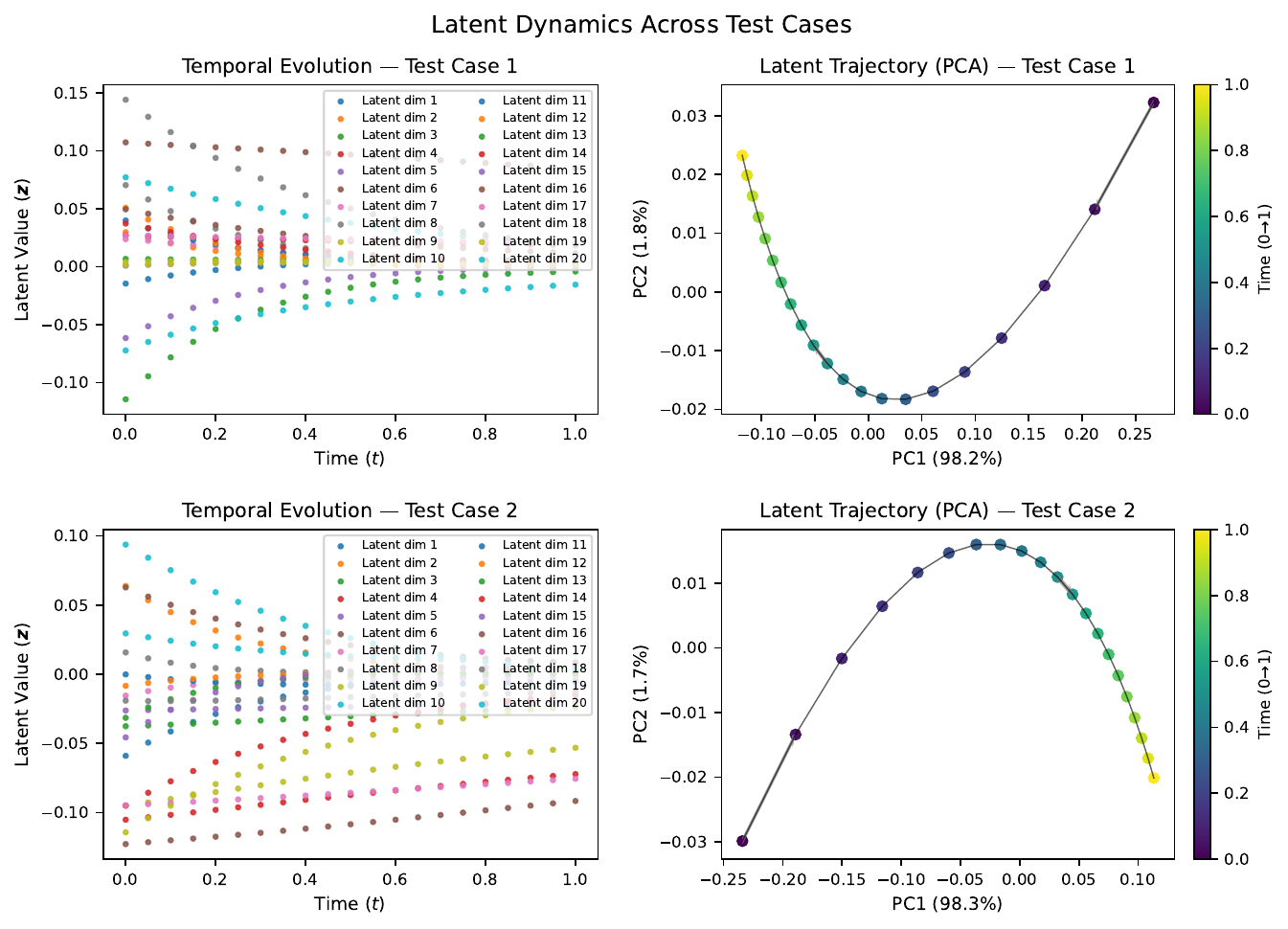}
    \caption{\textcolor{black}{Latent dynamics for the 2D Burgers’ equation problem. The temporal evolution of latent features (left) shows smooth, structured variations aligned with the underlying convective–diffusive flow physics. The PCA trajectories (right) form a well-defined low-dimensional manifold, confirming that the proposed PI-Latent-NO framework captures coherent spatiotemporal evolution in the latent space. The latent dimensionality for this problem is $n_{\bm{z}} = 60$; however, for visualization purposes, only the first $20$ latent dimensions are shown in the right panel.}}
    \label{fig:Latent Dynamics - 2D Burgers'}
\end{figure}

\textcolor{black}{In Figures~\ref{fig:Latent Dynamics - 1D Diffusion}–\ref{fig:Latent Dynamics - 2D Burgers'}, the left panels illustrate the evolution of individual latent dimensions over time ($t \in [0,1]$), while the right panels depict the latent trajectories in the reduced 2D PCA space. The temporal evolution plots exhibit smooth and continuous variations in the latent features, with no noise spikes or abrupt fluctuations, suggesting that the latent representations evolve coherently with the underlying physical dynamics. Furthermore, the PCA trajectories follow well-defined, low-dimensional manifolds - often resembling parabolic shapes - which indicate that the latent space effectively captures the dominant temporal modes of the system.}

\textcolor{black}{For instance, in the 1D Diffusion–reaction and 1D Burgers’ examples, the latent variables evolve smoothly over time with no abrupt fluctuations, reflecting the diffusive and advective nature of the respective processes. Similarly, for the 2D Stove–burner and 2D Burgers’ problems, despite their increased spatial and temporal complexity, the latent trajectories remain organized and continuous, confirming that the model successfully encodes consistent evolution patterns in the latent space. For all cases, the first two PCs of $\mathbf{z}$ account for more than 95\% of the total variance, indicating that almost all of the information regarding dynamic variation in the latent space can be represented by those two dimensions.}

\textcolor{black}{Overall, this latent dynamics analysis provides empirical evidence that the proposed PI-Latent-NO framework learns physically meaningful latent representations that align with the system’s inherent temporal evolution, thereby reinforcing the claim of physics-consistent learning in the latent space without requiring labeled data.}

\textcolor{black}{All the latent dynamics visualizations presented in this section correspond to the physics-informed case with $n_{\text{train}} = 0$, i.e., without using any labeled data. However, similar coherent and structured evolution patterns were consistently observed for cases with $n_{\text{train}} > 0$, demonstrating that the latent space evolution remains robust even when supervised data are incorporated during training.}

\section{Quantitative Assessment of Physical Consistency via PDE Squared Residuals}
\label{sec:pde squared residuals}

{\color{black}
To substantiate the claim of \textit{physics-consistent predictions without labeled data}, we report the discrete mean of the PDE squared residuals evaluated at the collocation points of the test dataset. This metric directly quantifies how well the predicted solutions satisfy the governing physical laws by measuring the magnitude of the PDE residuals over the spatiotemporal domain. This is computed as:

\begin{equation}
R_{\text{test}}^{\text{sq}} =
\frac{1}{n_{\text{test}}\, n_{t}\, n_{\bm{x}}}
\sum_{i=1}^{n_{\text{test}}} 
\sum_{j=1}^{n_{t}} 
\sum_{k=1}^{n_{\bm{x}}} 
\left( 
\frac{\partial \hat{u}(\boldsymbol{\xi}^{(i)}, t^{(j)}, \bm{x}^{(k)})}{\partial t} + 
\mathcal{N}[\hat{u}](\boldsymbol{\xi}^{(i)}, t^{(j)}, \bm{x}^{(k)})
\right)^2,
\end{equation}
where the term in parentheses represents the PDE residual at each collocation point, and the indices \( j \) and \( k \) span all spatiotemporal locations where the PDE solution is evaluated.
Lower residual values indicate stronger compliance with the governing PDE. The values reported in Tables~\ref{tab:1D Test PDE Squared Residuals}-\ref{tab:2D Test PDE Squared Residuals} correspond to the 1D and 2D benchmark problems, respectively, and reflect a single representative seed, as model weights were retained only for this run, whereas earlier results in Tables~\ref{tab:Example1_Performance metrics}–\ref{tab:Example4_Performance metrics} represent  results over five seeds. Nonetheless, these results provide meaningful insight into the physical consistency achieved by both the baseline vanilla model and our model.
}

\begin{table}[ht]
\centering
\caption{Test PDE Squared Residuals ($R_{\text{test}}^{\text{sq}}$) for 1D Problems (Single-Seed)}
\label{tab:1D Test PDE Squared Residuals}
\begin{tabular}{rlll}
\hline
$\mathrm{Model}$ & $\mathrm{n}_{\mathrm{train}}$ & 1D Diffusion-reaction & 1D Burgers' \\
\hline
PI-Vanilla-NO & 0 & 2.2e-05 & 1.8e-03 \\
PI-Latent-NO (Ours) & 0 & 8.2e-05 & 2.1e-03 \\
\hdashline
PI-Vanilla-NO & 100 & 6.8e-04 & 2.4e-03 \\
PI-Latent-NO (Ours) & 100 & 7.8e-05 & 1.8e-03 \\
\hdashline
PI-Vanilla-NO & 200 & 1.7e-05 & 2.7e-03 \\
PI-Latent-NO (Ours) & 200 & 9.1e-05 & 1.8e-03 \\
\hline
\end{tabular}
\end{table}

\begin{table}[ht]
\centering
\caption{Test PDE Squared Residuals ($R_{\text{test}}^{\text{sq}}$) for 2D Problems (Single-Seed)}
\label{tab:2D Test PDE Squared Residuals}
\begin{tabular}{rlll}
\hline
$\mathrm{Model}$ & $\mathrm{n}_{\mathrm{train}}$ &  2D Stove burner &  2D Burgers' \\
\hline
PI-Vanilla-NO & 0   & 4.1e-04 & 2.2e-05 \\
PI-Latent-NO (Ours) & 0   & 2.7e-04 & 2.4e-05 \\
\hdashline
PI-Vanilla-NO & 150 & 4.0e-04 & 2.2e-05 \\
PI-Latent-NO (Ours)  & 150 & 2.6e-04 & 2.6e-05 \\
\hdashline
PI-Vanilla-NO & 300 & 3.9e-04 & 2.2e-05 \\
PI-Latent-NO (Ours) & 300 & 2.7e-04 & 2.3e-05 \\
\hline
\end{tabular}
\end{table}

{\color{black}
\noindent
As observed, our PI-Latent-NO method consistently yields comparable or lower PDE squared residuals relative to the vanilla implementation across different training set sizes. These smaller residual magnitudes demonstrate that the latent-space representations not only reproduce accurate reconstructions but also remain consistent with the governing physical laws. The residual analysis thus provides direct empirical support for the physical consistency of the proposed framework.
}
\section{Summary}
\label{sec:summary}

In this work, we addressed the key limitations of existing Latent DeepONet \cite{kontolati2024learning,oommen2022learning} architectures, which rely on large datasets for data-driven training and cannot incorporate governing physics due to the two-step training process. To overcome these challenges, we introduced PI-Latent-NO, a physics-informed latent operator learning framework. This end-to-end architecture employs two coupled DeepONets: the \textit{Latent-DeepONet}, which identifies and learns a low-dimensional latent space, and the \textit{Reconstruction-DeepONet}, which maps the latent representations back to the original physical space.
This architecture offers two key advantages:
\begin{itemize}
\item It facilitates physics-informed training by enabling the computation of temporal and spatial derivatives through automatic differentiation, thereby reducing over-reliance on labeled dataset for training.
\item It exploits the separability of spatial and temporal components, achieving approximately linear computational scaling even for high-dimensional systems, compared to the quadratic scaling of physics-informed Vanilla DeepONet models.
\end{itemize}

\SKsays{Our results demonstrate the effectiveness of PI-Latent-NO as a proof of concept in learning high-dimensional input–output  mappings for parametric PDEs}. The framework consistently captures complex system dynamics with high accuracy, while optimizing computational and memory efficiency. Moreover, it effectively reduces redundant features by leveraging the latent space, enabling faster convergence. In conclusion, the PI-Latent-NO framework represents a transformative step in physics-informed machine learning. By seamlessly integrating latent space representation with governing physics, it \SKsays{ tackles} high-dimensional  \SKsays{input–output
mappings in parametric} PDEs, offering scalable and accurate solutions for future advancements in scientific computing. 

\textcolor{black}{While the proposed framework demonstrates promising accuracy and computational efficiency across the presented examples, we acknowledge that the computational gains were modest - about 15\% for the 2D Burgers' example - compared to the 30–70\% gains observed in other examples. This highlights the need for further methodological developments to enhance scalability to more complex 2D and 3D systems. Future efforts will focus on improving the expressiveness of the latent representation, incorporating separability in the Reconstruction-DeepONet trunk network architecture, and enabling automatic enforcement of boundary conditions to improve both efficiency and accuracy in higher-dimensional settings.}

\textcolor{black}{Furthermore, to advance the robustness and generalization capabilities of the proposed framework, we aim to address certain architectural limitations.}
\textcolor{black}{In its current form, our framework assumes independence between latent representations at consecutive time steps (i.e., $z_t$ and $z_{t+1}$), which may limit generalization, particularly for long-time horizon predictions. In future work, we aim to develop architectures that enforce this latent temporal dependency in a purely physics-informed manner, i.e., the latent representation at time $t+1$ is explicitly conditioned on the latent representation at time $t$, thereby improving temporal coherence and extrapolation capabilities. Additionally, we plan to extend the current framework to handle multi-output fields, incorporate separability in the trunk network of the Reconstruction-DeepONet by designing separate sub-networks for each spatial dimension, and include uncertainty quantification by adopting Bayesian neural networks for each of the neural components in our framework. These enhancements will further solidify PI-Latent-NO as a scalable, accurate, and uncertainty-aware tool for data and physics-driven modeling of complex dynamical systems.}

\textbf{Acknowledgements:}
{\color{black} We express our sincere gratitude to Professor Yannis Kevrekidis and Professor Michael Shields for their insightful discussions and invaluable guidance, which have significantly contributed to the development of this work. This work has been made possible by the financial support provided by the U.S. Department of Energy, Office of Science, Office of Advanced Scientific Computing Research, under Award Number DE-SC0024162.} 

\bibliographystyle{elsarticle-num} 
\bibliography{bibliography}

\newpage
\appendix
\renewcommand{\thetable}{A\arabic{table}} %
\setcounter{table}{0} 
\renewcommand{\theequation}{A\arabic{equation}} 
\setcounter{equation}{0}
\section{Appendix}

\begin{table}[H]
\centering
\caption{Summary of network architectures and hyperparameters used for training PI-Vanilla-NO model for the 1D benchmarks considered. MLP refers to a multi-layer perceptron.}

\tiny
{
\begin{tabular}{>{\centering\arraybackslash}p{8.0cm} 
                >{\centering\arraybackslash}p{4.0cm} 
                >{\centering\arraybackslash}p{4.0cm}}
\toprule
\textbf{Case} & \textbf{1D Diffusion - reaction dynamics} & \textbf{1D Burgers’ \SKsays{ equation}}  \\ 
\midrule
\multicolumn{3}{c}{\textbf{Training and Test Configurations}} \\
\midrule
\textbf{No. of input functions } & 1000 & 1000  \\ 
\textbf{No. of training trajectories} ($n_{\text{train}}$) & \{0, 100, 200\} & \{0, 100, 200\} \\ 
\textbf{No. of testing trajectories} ($n_{\text{test}})$ & 500 & 500  \\ 
\textbf{Discretization of the solution field} ($(n_t + 1) \times n_{\bm{x}}$) & 101 $\times$ 100 & 101 $\times$ 101  \\ 
\midrule
\multicolumn{3}{c}{\textbf{PI-Vanilla-NO Architectures and Training Settings}} \\
\midrule
\textbf{DeepONet branch net} & 
MLP: [100, 64, SiLU, 64, SiLU, 64, SiLU, 128] & 
MLP: \SKsays{[101, 100, Tanh, 100, Tanh, 100, Tanh, 100, Tanh, 100, Tanh, 100, Tanh, 100, Tanh, 128]} \\ 
\textbf{DeepONet trunk net} & 
MLP: [2, 64, SiLU, 64, SiLU, 64, SiLU, 128] & 
MLP: \SKsays{[2, 100, Tanh, 100, Tanh, 100, Tanh, 100, Tanh, 100, Tanh, 100, Tanh, 100, Tanh, 128]}  \\ 
\textbf{No. of input functions used per iteration} ($n_i$) & 64 & 64 \\ 
\textbf{No. of collocation points within the domain} ($n_{t}^r, n_{\bm{x}}^r$) & 256, 256 & 512, 512  \\ 
\textbf{No. of collocation points on each boundary} ($n_{t}^{bc}, n_{\bm{x}}^{bc}$) & 256, 1 & 512, 1  \\ 
\textbf{No. of collocation points at} $t=0$ ($n_{\bm{x}}^{ic}$) & 256 & 512  \\ 
\textbf{Optimizer} & Adam  & Adam\\
\textbf{No. of iterations} & 50,000 & \SKsays{80,000}  \\ 
\textbf{Learning rate} & $3.5 \times 10^{-3}$ & \SKsays{$1.0 \times 10^{-3}$} \\
\textbf{Learning rate scheduler} & Step LR & \SKsays{Exponential LR} \\
\textbf{Scheduler step size} & 15,000 & \SKsays{2000} \\
\textbf{Scheduler decay factor ($\gamma$)} & 0.1  & \SKsays{0.9}\\
\textbf{Loss weighing factor ($\lambda$'s)} & 1  & \SKsays{$\lambda_{\text{ic}}=10$}\\
\bottomrule
\end{tabular}
}
\label{tab:architectures-hyperparameters-vanilla-1D}
\end{table}

\begin{table}[H]
\centering
\caption{Summary of network architectures and hyperparameters used for training PI-Vanilla-NO model for the 2D benchmarks considered. The Conv2D layers, representing 2D convolution layers, are defined by the number of output filters, kernel size, stride, padding, and activation function. The Average Pooling layers are specified by kernel size, stride, and padding. The ResNet layers, referring to residual networks, are configured with the number of ResNet blocks, the number of layers per block, the number of neurons in each layer, and the activation function. Additionally, MLP refers to multi-layer perceptron.}
\tiny
{
\begin{tabular}{>{\centering\arraybackslash}p{8.0cm} 
                >{\centering\arraybackslash}p{4.0cm} 
                >{\centering\arraybackslash}p{4.0cm}}
\toprule
\textbf{Case} & \textbf{2D Stove-Burner Simulation} & \textbf{2D Burgers’ \SKsays{ equation}}  \\ 
\midrule
\multicolumn{3}{c}{\textbf{Training and Test Configurations}} \\
\midrule
\textbf{No. of input functions  } & 2000 & 1000  \\ 
\textbf{No. of training trajectories} ($n_{\text{train}}$) & \{0, 150, 300\} & \{0, 150, 300\}  \\ 
\textbf{No. of testing trajectories} ($n_{\text{test}})$ & 100 & 50  \\ 
\textbf{Discretization of the solution field} ($(n_t + 1) \times n_{\bm{x}}$) & 21 $\times$ $64^2$ & 21 $\times$ $32^2$  \\ 
\midrule
\multicolumn{3}{c}{\textbf{PI-Vanilla-NO Architectures and Training Settings}} \\
\midrule
\textbf{DeepONet branch net} & 
\begin{tabular}[c]{@{}c@{}}
Input: (64, 64) \\
Conv2D: (40, (3, 3), 1, 0, ReLU) \\
Average pooling: ((2, 2), 2, 0) \\
Conv2D: (60, (3, 3), 1, 0, ReLU) \\
Average pooling: ((2, 2), 2, 0) \\
Conv2D: (80, (3, 3), 1, 0, ReLU) \\
Average pooling: ((2, 2), 2, 0) \\
Conv2D: (100, (3, 3), 1, 0, ReLU) \\
Average pooling: ((2, 2), 2, 0) \\
Flatten() \\
MLP: [150, ReLU, 150, ReLU, 128]
\end{tabular}  & 
\begin{tabular}[c]{@{}c@{}}
Input: (32, 32) \\
Conv2D: (20, (3, 3), 1, 0, SiLU) \\
Average pooling: ((2, 2), 2, 0) \\
Conv2D: (30, (3, 3), 1, 0, SiLU) \\
Average pooling: ((2, 2), 2, 0) \\
Conv2D: (40, (3, 3), 1, 0, SiLU) \\
Average pooling: ((2, 2), 2, 0) \\
Flatten() \\
MLP: [150, SiLU, 150, SiLU, 128]
\end{tabular} \\ 
\textbf{DeepONet trunk net} & 
MLP: [3, 128, SiLU, 128, SiLU, 128, SiLU, 128, SiLU, 128] & 
MLP: [43, 128, SiLU, 128, SiLU, 128, SiLU, 128, SiLU, 128]  \\ 
\textbf{No. of input functions used per iteration} ($n_i$) & 128 & 32 \\ 
\textbf{No. of collocation points within the domain} ($n_{t}^r, n_{\bm{x}}^r$) & 20, $64^2$ & 21, $64^2$  \\ 
\textbf{No. of collocation points on each boundary} ($n_{t}^{bc}, n_{\bm{x}}^{bc}$) & -, - & 21, 64  \\ 
\textbf{No. of collocation points at} $t=0$ ($n_{\bm{x}}^{ic}$) & - & $64^2$  \\ 
\textbf{Optimizer} & Adam  & Adam\\
\textbf{No. of iterations} & 50,000 & 80,000  \\ 
\textbf{Learning rate} & $10^{-3}$ & $10^{-3}$ \\
\textbf{Learning rate scheduler} & Step LR & Constant LR \\
\textbf{Scheduler step size} & 20,000 & - \\
\textbf{Scheduler decay factor ($\gamma$)} & 0.1  & -\\
\textbf{Loss weighing factor ($\lambda$'s)} & 1  & 1 \\
\bottomrule
\end{tabular}
}
\label{tab:architectures-hyperparameters-vanilla-2D}
\end{table}

\begin{table}[H]
\centering
\caption{Summary of network architectures and hyperparameters used in our PI-Latent-NO model for the 1D benchmarks considered. MLP refers to a multi-layer perceptron.}
\tiny
{
\begin{tabular}{>{\centering\arraybackslash}p{8.0cm} 
                >{\centering\arraybackslash}p{4.0cm} 
                >{\centering\arraybackslash}p{4.0cm} 
                 }
\toprule
\textbf{Case} & \textbf{1D Diffusion - reaction dynamics} & \textbf{1D Burgers’ \SKsays{  equation}}\\ 
\midrule
\multicolumn{3}{c}{\textbf{Training and Test Configurations}} \\
\midrule
\textbf{No. of input functions } & 1000 & 1000   \\ 
\textbf{No. of training trajectories} ($n_{\text{train}}$) & \{0, 100,  200\} & \{0, 100,  200\} \\ 
\textbf{No. of testing trajectories} ($n_{\text{test}})$ & 500 & 500   \\ 
\textbf{Discretization of the solution field} ($(n_t + 1) \times n_{\bm{x}}$) & 101 $\times$ 100 & 101 $\times$ 101 \\ 
\midrule
\multicolumn{3}{c}{\textbf{PI-Latent-NO Architectures and Training Settings}} \\
\midrule
\textbf{Latent dimension size} ($d_{\bm{z}}$) & 9 & 9  \\ 
\textbf{Latent deeponet branch net} & 
MLP: [100, 64, SiLU, 64, SiLU, 64, SiLU, 25$d_{\bm{z}}$] & 
MLP: \SKsays{[101, 100, Tanh, 100, Tanh, 100, Tanh, 100, Tanh, 100, Tanh, 100, Tanh, 100, Tanh, 16$d_{\bm{z}}$]}\\ 
\textbf{Latent deeponet trunk net} & 
MLP: [1, 64, SiLU, 64, SiLU, 64, SiLU, 25$d_{\bm{z}}$] & 
MLP: \SKsays{[1, 100, Tanh, 100, Tanh, 100, Tanh, 100, Tanh, 100, Tanh, 100, Tanh, 100, Tanh, 16$d_{\bm{z}}$]}  \\ 
\textbf{Reconstruction deeponet branch net} & 
MLP: [$d_{\bm{z}}$, 64, SiLU, 64, SiLU, 64, SiLU, 128] & 
MLP: \SKsays{[$d_{\bm{z}}$, 100, Tanh, 100, Tanh, 100, Tanh, 100, Tanh, 100, Tanh, 100, Tanh, 100, Tanh, 128]}  \\ 
\textbf{Reconstruction deeponet trunk net} & 
MLP: [1, 64, SiLU, 64, SiLU, 64, SiLU, 128]  & 
MLP: \SKsays{[1, 100, Tanh, 100, Tanh, 100, Tanh, 100, Tanh, 100, Tanh, 100, Tanh, 100, Tanh, 128]}  \\ 
\textbf{No. of input functions used per iteration} ($n_i$) & 64 & 64   \\ 
\textbf{No. of collocation points within the domain} ($n_{t}^r, n_{\bm{x}}^r$) & 256, 256 & 512, 512  \\ 
\textbf{No. of collocation points on each boundary} ($n_{t}^{bc}, n_{\bm{x}}^{bc}$) & 256, 1 & 512, 1   \\ 
\textbf{No. of collocation points at} $t=0$ ($n_{\bm{x}}^{ic}$) & 256 & 512  \\ 
\textbf{Optimizer} & Adam  & Adam\\
\textbf{No. of iterations} & 50,000 & \SKsays{80,000}   \\ 
\textbf{Learning rate} & $3.5 \times 10^{-3}$ & \SKsays{$1.0 \times 10^{-3}$} \\
\textbf{Learning rate scheduler} & Step LR & \SKsays{Exponential LR} \\
\textbf{Scheduler step size} & 15,000 & \SKsays{2,000} \\
\textbf{Scheduler decay factor ($\gamma$)} & 0.1  & \SKsays{0.9}\\
\textbf{Loss weighing factor ($\lambda$'s)} & 1  & \SKsays{$\lambda_{\text{ic}}=10$}\\
\bottomrule
\end{tabular}
}
\label{tab:architectures-hyperparameters-ours-1D}
\end{table}

\begin{table}[H]
\centering
\caption{Summary of network architectures and hyperparameters used in our PI-Latent-NO model for the 2D benchmarks considered. The Conv2D layers, representing 2D convolution layers, are defined by the number of output filters, kernel size, stride, padding, and activation function. The Average Pooling layers are specified by kernel size, stride, and padding. The ResNet layers, referring to residual networks, are configured with the number of ResNet blocks, the number of layers per block, the number of neurons in each layer, and the activation function. Additionally, MLP refers to multi-layer perceptron.}
\tiny
{
\begin{tabular}{>{\centering\arraybackslash}p{8.0cm} 
                >{\centering\arraybackslash}p{4.0cm} 
                >{\centering\arraybackslash}p{4.0cm} 
                  }
\toprule
\textbf{Case} & \textbf{2D Stove-Burner Simulation} & \textbf{2D Burgers’ \SKsays{ equation}}  \\ 
\midrule
\multicolumn{3}{c}{\textbf{Training and Test Configurations}} \\
\midrule
\textbf{No. of input functions  } & 2000 & 1000   \\ 
\textbf{No. of training trajectories} ($n_{\text{train}}$) & \{0, 150, 300\} & \{0, 150, 300\}  \\ 
\textbf{No. of testing trajectories} ($n_{\text{test}})$ & 100 & 50   \\ 
\textbf{Discretization of the solution field} ($(n_t + 1) \times n_{\bm{x}}$) & 21 $\times$ $64^2$ & 21 $\times$ $32^2$ \\ 
\midrule
\multicolumn{3}{c}{\textbf{PI-Latent-NO Architectures and Training Settings}} \\
\midrule
\textbf{Latent dimension size} ($d_{\bm{z}}$) & 16 & 60  \\ 
\textbf{Latent deeponet branch net} & 
\begin{tabular}[c]{@{}c@{}}
Input: (64, 64) \\
Conv2D: (40, (3, 3), 1, 0, ReLU) \\
Average pooling: ((2, 2), 2, 0) \\
Conv2D: (60, (3, 3), 1, 0, ReLU) \\
Average pooling: ((2, 2), 2, 0) \\
Conv2D: (80, (3, 3), 1, 0, ReLU) \\
Average pooling: ((2, 2), 2, 0) \\
Conv2D: (100, (3, 3), 1, 0, ReLU) \\
Average pooling: ((2, 2), 2, 0) \\
Flatten() \\
MLP: [150, ReLU, 150, ReLU, 16$d_{\bm{z}}$]
\end{tabular}& 
\begin{tabular}[c]{@{}c@{}}
Input: (32, 32) \\
Conv2D: (20, (3, 3), 1, 0, SiLU) \\
Average pooling: ((2, 2), 2, 0) \\
Conv2D: (30, (3, 3), 1, 0, SiLU) \\
Average pooling: ((2, 2), 2, 0) \\
Conv2D: (40, (3, 3), 1, 0, SiLU) \\
Average pooling: ((2, 2), 2, 0) \\
Flatten() \\
MLP: [150, SiLU, 150, SiLU, 10$d_{\bm{z}}$]
\end{tabular}  \\ 
\textbf{Latent deeponet trunk net} & 
MLP: [1, 128, SiLU, 128, SiLU, 128, SiLU, 128, SiLU, 16$d_{\bm{z}}$] & 
MLP: [1, 128, SiLU, 128, SiLU, 128, SiLU, 128, SiLU, 10$d_{\bm{z}}$] \\ 
\textbf{Reconstruction deeponet branch net} & 
MLP: [$d_{\bm{z}}$, 128, SiLU, 128, SiLU, 128, SiLU, 128, SiLU, 128] & 
MLP: [$d_{\bm{z}}$, 128, SiLU, 128, SiLU, 128, SiLU, 128, SiLU, 128]  \\ 
\textbf{Reconstruction deeponet trunk net} & 
MLP: [2, 128, SiLU, 128, SiLU, 128, SiLU, 128, SiLU, 128]  & 
MLP: [42, 128, SiLU, 128, SiLU, 128, SiLU, 128, SiLU, 128]  \\ 
\textbf{No. of input functions used per iteration} ($n_i$) & 128 & 32  \\ 
\textbf{No. of collocation points within the domain} ($n_{t}^r, n_{\bm{x}}^r$) & 20, $64^2$ & 21, $64^2$  \\ 
\textbf{No. of collocation points on each boundary} ($n_{t}^{bc}, n_{\bm{x}}^{bc}$) & -, - & 21, 64   \\ 
\textbf{No. of collocation points at} $t=0$ ($n_{\bm{x}}^{ic}$) & - & $64^2$  \\ 
\textbf{Optimizer} & Adam  & Adam\\
\textbf{No. of iterations} & 50,000 & 80,000  \\ 
\textbf{Learning rate} & $10^{-3}$ & $10^{-3}$ \\
\textbf{Learning rate scheduler} & Step LR & Constant LR \\
\textbf{Scheduler step size} & 20,000 & - \\
\textbf{Scheduler decay factor ($\gamma$)} & 0.1  & -\\
\textbf{Loss weighing factor ($\lambda$'s)} & 1  & 1\\
\bottomrule
\end{tabular}
}
\label{tab:architectures-hyperparameters-ours-2D}
\end{table}

\begin{table}[H]
\centering
\caption{Source Term Equations for Different Geometries}
\tiny
\renewcommand{\arraystretch}{2.3} 
\begin{tabular}{|l|l|}
\hline
\textbf{Shape} & \textbf{Equation} \\
\hline
\textbf{circle} & \( s(x_1, x_2, \text{circle}, a, r) = \exp \left( - \left| a \left( \sqrt{x_1^2 + x_2^2} - r \right) \right| \right) \) \\
\hline
\textbf{half-circle} & \( s(x_1, x_2, \text{half-circle}, a, r) = \exp \left( - \left| a \max \left( \sqrt{x_1^2 + (x_2 + \frac{r}{2})^2} - r, - \left( x_2 + \frac{r}{2} \right) \right) \right| \right) \) \\
\hline
\textbf{isosceles triangle} & \( s(x_1, x_2, \text{isosceles triangle}, a, r) = \exp \left( - \left| a \max \left( \max \left( x_2 - \frac{\sqrt{3}}{2}x_1 - \frac{r}{2}, x_2 + \frac{\sqrt{3}}{2}x_1 - \frac{r}{2} \right), -x_2 - \frac{r}{2} \right) \right| \right) \) \\
\hline
\textbf{right-angled triangle} & 
\( s(x_1, x_2, \text{right-angled triangle}, a, r) = \exp \left( - \left| a \cdot \max \left( \max \left( -\left( x_2 + \frac{r}{3} \right), - \left( x_1 + \frac{r}{3} \right) \right), \left( x_1 + \frac{r}{3} \right) + \left( x_2 + \frac{r}{3} \right) - r \right) \right| \right) \) \\
\hline
\textbf{rectangle} & \( s(x_1, x_2, \text{rectangle}, a, r) = \exp\left( - \left| a \left( \max\left( \frac{|x_1|}{r}, \frac{|x_2|}{r/2} \right) - 1 \right) \right| \right) \) \\
\hline
\textbf{square} & \( s(x_1, x_2, \text{square}, a, r) = \exp \left( - \left| a \left( \max \left( |x_1|, |x_2| \right) - \frac{r}{2} \right) \right| \right) \) \\
\hline
\textbf{rhombus} & \( s(x_1, x_2, \text{rhombus}, a, r) = \exp \left( - \left| a \left( |x_1| + |x_2| - r \right) \right| \right) \) \\
\hline
\textbf{\(n\)-sided polygon}& 
\(
\begin{aligned}
&s(x_1, x_2, n\text{-sided polygon}, a, r) = 
\exp \left( -a \cdot \displaystyle\min_{1 \leq i \leq n} d_{\text{signed}}(x_1, x_2, x_1^i, x_2^i, x_1^{i+1}, x_2^{i+1}, r, n) \right), \\
&\text{where } d_{\text{signed}}(x_1, x_2, x_1^i, x_2^i, x_1^{i+1}, x_2^{i+1}, r, n) = 
\sqrt{ \left( x_1^i + u \cdot (x_1^{i+1} - x_1^i) - x_1 \right)^2 + \left( x_2^i + u \cdot (x_2^{i+1} - x_2^i) - x_2 \right)^2 }, \\
&\text{with } u = \mathrm{clip} \left( \frac{(x_1 - x_1^i)(x_1^{i+1} - x_1^i) + (x_2 - x_2^i)(x_2^{i+1} - x_2^i)}{(x_1^{i+1} - x_1^i)^2 + (x_2^{i+1} - x_2^i)^2}, \; 0, \; 1 \right), \\
&\text{and } (x_1^i, x_2^i) = \left( r \cos\left( \tfrac{2\pi(i - 1)}{n} + \tfrac{\pi}{n} \right), \; r \sin\left( \tfrac{2\pi(i - 1)}{n} + \tfrac{\pi}{n} \right) \right)
\end{aligned}
\) \\
\hline
\end{tabular}
\label{tab:source_terms}
\end{table}

\end{document}